\documentclass{article}
\usepackage[T1]{fontenc}
\usepackage{iclr2027_conference,times}
\usepackage{graphicx,booktabs,amsmath,amssymb,float}
\usepackage{xcolor}
\colorlet{EARevisionBlue}{black}
\newcommand{\EARevision}[1]{{\color{EARevisionBlue}#1}}
\usepackage{hyperref,url,fontawesome5}
\setcitestyle{citesep={,}}
\hypersetup{
  hidelinks,
  pdfborder={0 0 0}
}
\usepackage{tikz,makecell,array,multirow}
\usetikzlibrary{positioning,arrows.meta}
\DeclareRobustCommand{\eaQuestion}[1]{\tikz[baseline=(q.base)]{\node[circle,fill=black,text=white,inner sep=0pt,minimum size=8pt,font=\bfseries\fontsize{6}{6}\selectfont] (q) {#1};}}
\usepackage{tcolorbox,listings,longtable,colortbl,placeins}

\newcommand{\eaIdentifier}[1]{\textcolor{magenta}{\texttt{#1}}}
\lstdefinestyle{eaActionCode}{
 basicstyle=\ttfamily,columns=fullflexible,keepspaces=true,
 showstringspaces=false,breaklines=true,
 literate={tool_calls}{{\eaIdentifier{tool\_calls}}}{10}
          {tool_call}{{\eaIdentifier{tool\_call}}}{9}
          {ask_user}{{\eaIdentifier{ask\_user}}}{8}
}

\tcbuselibrary{breakable,skins}
\definecolor{eaBoxBack}{RGB}{247,247,247}
\definecolor{eaKey}{RGB}{180,36,70}
\definecolor{eaValue}{RGB}{40,108,65}
\definecolor{eaInk}{RGB}{43,61,72}
\newtcolorbox{eaExampleBox}[2][]{
 enhanced,breakable,colback=eaBoxBack,colframe=black!65,
 boxrule=0.7pt,arc=0pt,left=7pt,right=7pt,top=6pt,bottom=6pt,
 fonttitle=\sffamily\bfseries\small,colbacktitle=eaInk,coltitle=white,
 title={#2},before skip=7pt,after skip=7pt,fontupper=\small,#1
}
\lstdefinelanguage{eaJSON}{
 sensitive=true,string=[s]{"}{"},stringstyle=\color{eaValue},
 morekeywords={true,false,null},keywordstyle=\color{eaValue}\bfseries
}
\lstdefinestyle{eaJSONStyle}{
 language=eaJSON,basicstyle=\ttfamily\footnotesize,
 columns=fullflexible,keepspaces=true,showstringspaces=false,
 breaklines=true,breakatwhitespace=false,numbers=none,
 frame=none,aboveskip=0pt,belowskip=0pt,tabsize=2
}

\definecolor{EAGrounded}{HTML}{78289B}
\definecolor{EAOtherCredit}{HTML}{FFB000}
\definecolor{EAOverallTint}{HTML}{CB9BCD}
\newtcolorbox{eaTakeaway}{colback=EARevisionBlue!3,colframe=EARevisionBlue!55!black,
 boxrule=.5pt,arc=1pt,left=5pt,right=5pt,top=3pt,bottom=3pt,
 before skip=5pt,after skip=5pt,fontupper=\small}
\newsavebox{\eaTableBox}
\newcommand{\EAfitTable}[1]{%
    \sbox{\eaTableBox}{#1}%
    \ifdim\wd\eaTableBox>\linewidth
        \resizebox{\linewidth}{!}{\usebox{\eaTableBox}}%
    \else
        \usebox{\eaTableBox}%
    \fi
}
\newsavebox{\tablebox}

\title{EA-Bench: Measuring Information and Action Efficiency under Imperfect Instructions}
\iclrfinalcopy
\renewcommand{\headrulewidth}{0pt}
\author{
Wenyue Xu$^{2,*}$ \quad Xu Wan$^{1,*}$ \quad Shenjie Zhao$^{2}$\\[5pt]
{\normalfont $^1$Independent Researcher \quad $^2$Tongji University}\\[3pt]
{\normalfont\small $^*$Equal contribution.}
}
\hypersetup{
  pdftitle={EA-Bench: Measuring Information and Action Efficiency under Imperfect Instructions},
  pdfauthor={Wenyue Xu; Xu Wan; Shenjie Zhao},
  pdfsubject={Preprint}
}
\begin{document}
\ifdefined\pdfminorversion
  \pdfminorversion=7
\else
  
\fi
\maketitle
\begin{abstract}
Personal agents help users complete everyday work across digital applications. Many standard benchmarks assume that users provide sufficient information to solve a task. However, agents often encounter incomplete, distracting, or inconsistent information in practice. To systematically study agent inefficiency under such imperfect information, we introduce Efficient Action (EA)-Bench, which evaluates both tool calls and user interactions in work-related tasks. To our knowledge, it is the first benchmark to jointly evaluate tool-call efficiency and user-interaction efficiency under controlled missing, redundant, and conflicting information. We construct 63 base-task instances adapted from existing work-related benchmarks. To guide perturbation design and quantify task requirements, we represent workflows as directed acyclic graphs (DAGs) and examine node-level dependencies to identify where information can be withheld, contradicted, or supplemented with distractors. We use a two-stage validation process combining AI-assisted checks with expert screening to select 232 conditions from 792 initial candidates. We organize these tasks into \texttt{Easy} and \texttt{Hard} tracks by task complexity. \EARevision{Experiments show that imperfect information can reduce accuracy and increase tool-use and user-interaction costs. More concerning, several models obtain apparently correct outcomes by guessing user-specific values without confirmation, revealing potential execution unreliability in current models under imperfect user information.}
\end{abstract}

{\centering\small\faCode\hspace{0.6em}\url{https://github.com/waunx/EA-Bench.git}\par}

\section{Introduction}

Language models are increasingly used for software engineering, document preparation, and spreadsheet analysis \citep{anthropic2026opus55,openai2026astra}. As agents, they combine tool operations with dialogue to obtain information and complete tasks \citep{trivedi2024appworld,yao2023react}. The user supplies goals and preferences, while the agent accesses records, makes decisions, and changes the environment. Reliable execution depends on how the agent handles the information available at each step, including when it needs to ask the user for clarification. However, user requests often leave essential details unstated. For example, a request such as ``Return the shirt I bought last week'' may refer to several purchases. Incorrect actions taken without confirming user intent can cause substantial financial loss. An effective personal agent must therefore efficiently recover missing facts, filter irrelevant material, and resolve conflicting information before acting.

\begin{figure}[t]\centering
\includegraphics[width=\linewidth]{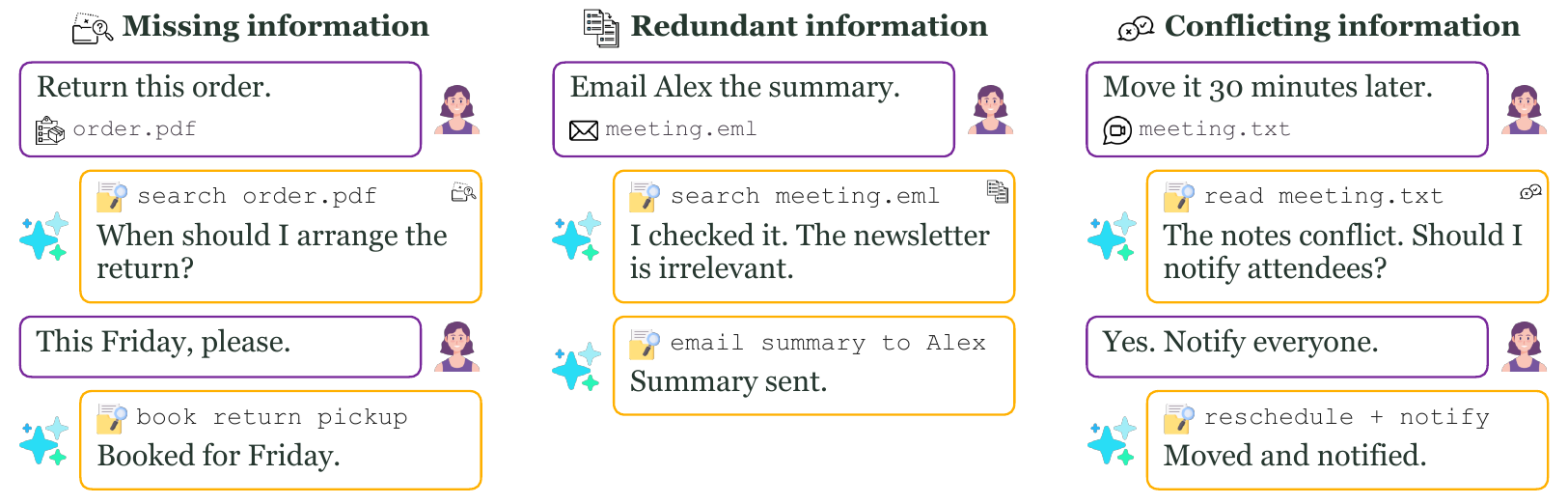}
\caption{\EARevision{\textbf{Illustrative interactions with hypothetical attachments.} Purple and orange outlines denote user and agent messages. Each agent block shows tool use followed by its message.}}
\label{fig:information-examples}
\end{figure}

Existing interactive benchmarks examine aspects of the efficiency and control challenges introduced by user involvement. $\tau$-bench measures task completion, policy adherence, and reliability across repeated trials with simulated users \citep{yao2024tau}. In $\tau^2$-bench, both agents and users can operate tools in a shared environment, making communication and coordination part of task execution \citep{tau-bench-2}. AppWorld-UL introduces controlled perturbations and user-held knowledge across multi-app tasks, and reports interaction recall, precision, and cost \citep{chen2026appworldul}. These benchmarks make evaluation more realistic, but a remaining challenge is to systematically quantify agents' ability to detect information imperfections and resolve them efficiently.

Our design considers three cases (Figure~\ref{fig:information-examples}):  \textbf{(1) Missing information.} An agent may skip an unstated constraint and still produce a result that happens to satisfy it. Although the task appears complete, this success does not show that the agent recovered the missing fact and may conceal a mismatch with the user's intended goal.  \textbf{(2) Redundant information.} An agent may repeatedly invoke tools to inspect files that appear relevant but contribute little to the task. Such unnecessary exploration increases tool calls and completion time without advancing the user's goal.  \textbf{(3) Conflicting information.} When available evidence cannot establish which statement is authoritative, the agent must ask the user for clarification before choosing between incompatible actions. Evaluation should measure whether it detects the conflict and resolves it with minimal interaction overhead.

EA-Bench evaluates these questions about user experience using quantifiable measures of tool-use and user-interaction efficiency. First, we adapt work-related scenarios from JobBench, AppWorld-UL, and WorkBench \citep{jobbench,chen2026appworldul,workbench}, drawing on tasks already curated by these benchmarks. Within each task, we represent its workflow as a directed acyclic graph (DAG) and examine node-level dependencies. We use DAG-guided counterfactual reasoning to construct candidate missing, redundant, conflicting, and combined conditions while preserving the business objective.

We use a two-stage validation process combining AI-assisted checks with expert screening to select 232 conditions from 792 initial candidates. The AI-assisted checker examines each candidate against its public inputs, DAG dependencies, and task requirements. It checks whether omitted facts remain inferable, whether conflicting statements have a resolvable authority ordering, and whether redundant additions preserve the authoritative facts and required outcome. Counterfactual checks test whether changing an unresolved fact affects task-relevant decisions. Executable replay and finite-world checks support consistency and solvability checks where the task protocol permits them.

Human experts then assess whether the introduced omissions, inconsistencies, and distractions plausibly reflect mistakes or communication patterns of real users. They further screen out contrived perturbations that do not represent realistic user behavior.

The final benchmark retains 63 base-task instances and 169 imperfect variants, together with one complete-information baseline per task. To separate efficiency from underlying task difficulty, we organize the instances into \texttt{Easy} and \texttt{Hard} tracks by task complexity. Grounded and ungrounded scores distinguish task or rubric credit supported by the necessary acquired information from credit obtained through unsupported assumptions. For efficiency, $E_t$ and $E_i$ measure signed differences between recorded tool calls or information requests and task-specific references within the declared protocols. We compute these gaps for successful and failed attempts where validated references exist, and retain actual resource measurements for open-ended \texttt{Hard} tasks. We evaluate nine mainstream open- and closed-weight model services across all 232 conditions.

Our contributions are:
\begin{enumerate}
\item \textbf{EA-Bench quantifies efficiency under imperfect instructions.} To our knowledge, EA-Bench is the first benchmark to jointly compare recorded tool use and information requests with explicit, protocol-scoped task references under missing, redundant, and conflicting instructions.
\item \textbf{EA-Bench provides a reusable construction procedure for imperfect information.} DAG dependencies and counterfactual checks identify task-specific action and information requirements. A two-stage checker combines AI-assisted validation with expert screening for perturbation validity and realism.
\item \textbf{EA-Bench reveals inefficient recovery from imperfect instructions.} Across nine evaluated services, imperfect variants reduce \texttt{Easy} accuracy by 5.3--24.5 percentage points relative to their task-matched Full baselines. Moreover, under imperfect variants, 22.7\% of successful \texttt{Easy} outputs receive ungrounded credit.
\end{enumerate}

\section{Related Work}

\textbf{Execution and efficiency evaluation.} Since early frameworks such as ReAct linked reasoning with external actions \citep{yao2023react}, agent evaluation has increasingly focused on effective tool use and task execution. BFCL evaluates function calling \citep{patil2025the}. WebArena, SWE-bench, and AppWorld check executed outcomes in realistic environments \citep{webarena,swebench,trivedi2024appworld}. Efficiency benchmarks additionally measure the resources consumed by successful solutions. EffiBench measures code runtime and memory against canonical programs \citep{huang2024effibench}, and OSWorld-Human compares computer-use trajectories with human references \citep{abhyankar2025osworldhuman}. AgentBoard tracks intermediate progress, supplies golden actions for tool subsets, and reports runtime estimates \citep{ma2024agentboard}. These references support different notions of necessary execution. They should not all be interpreted as theoretical optima. Joint accuracy and cost reporting is also motivated by AI Agents That Matter \citep{kapoor2024agents}.

\textbf{Information and user interaction.} User interaction is central to clarifying task requirements and aligning agent behavior with user intent \citep{subramonyam2024}. $\tau$-bench, $\tau^2$-bench, ToolSandbox, and UserBench make communication part of agent evaluation \citep{yao2024tau,tau-bench-2,toolsandbox,userbench}. AppWorld-UL additionally controls user-held knowledge and measures coverage of expected questions and interaction burden \citep{chen2026appworldul}. Context benchmarks examine complementary abilities. CL-bench studies applying supplied knowledge \citep{dou2026clbench}, RULER and HELMET test long-context use \citep{hsieh2024ruler,yen2024helmet}, and IFEval and FollowBench test verifiable instruction constraints \citep{zhou2023instruction,jiang2024followbench}. Simulated-user results remain conditional on the proxy's validity \citep{lost-in-simulation}. EA-Bench brings execution references and necessary-information accounting together with separate missing, redundant, and conflicting-policy variants. Table~\ref{tab:coverage-comparison} compares these selected dimensions.

\begin{table}[H]\centering\small
\caption{Benchmark comparison of agent interaction efficiency and outcome scores.}
\label{tab:coverage-comparison}
\begingroup
\newcommand{\benchyes}{\textcolor{EAGrounded}{\ensuremath{\boldsymbol{\checkmark}}}}
\newcommand{\benchno}{\textcolor{EAOtherCredit}{\ensuremath{\boldsymbol{\times}}}}
\setlength{\tabcolsep}{2pt}
\renewcommand{\arraystretch}{1.2}
\EAfitTable{\begin{tabular}{@{}lcccccc@{}}\toprule
Benchmark & DAG nodes & Tool Efficiency & Interaction Efficiency & User Simulator & Grounded Score & Ungrounded Score \\ \midrule
\texttt{WorkBench} & 7--11 & -- & -- & -- & \benchno & \benchno \\
\texttt{JobBench} & 13--30 & \benchyes & -- & -- & \benchno & \benchno \\
$\tau$\texttt{-bench} & $\approx$1--20 & -- & -- & \texttt{GPT-4} & \benchno & \benchno \\
$\tau^2$\texttt{-bench} & $\approx$1--12 & -- & -- & \texttt{GPT-4.1} & \benchno & \benchno \\
\texttt{AppWorld-UL} & 8--13 & -- & \benchyes & \texttt{GPT-5.5} & \benchyes & \benchno \\ \midrule
\rowcolor{EAOverallTint!25}\textbf{\texttt{EA-Bench}} & \textbf{4--30} & \benchyes & \benchyes & \texttt{GPT-5.5} & \benchyes & \benchyes \\
\bottomrule\end{tabular}}
\par\vspace{3pt}
\begin{minipage}{\linewidth}\footnotesize
\benchyes/\benchno: separate metric reported / not reported. --: not reported or not applicable.
Efficiency: tool-use / user-query cost. Simulators: cited defaults.
\end{minipage}
\endgroup
\end{table}

\section{Task and Interaction Formulation}
\label{sec:formulation}

\paragraph{Task specification and paired variants.}

Formally, we define a base task as
{\color{EARevisionBlue}
\begin{equation}
\mathcal T=(D,\theta,G,\mathcal A,\mathcal J),
\label{eq:task-definition}
\end{equation}}
Here, $D$ contains task data, such as supplier quotations or calendar records. The parameter $\theta$ specifies authoritative facts and user requirements, such as required supplier certification or attendee notification after rescheduling. The directed acyclic graph $G=(V,E)$ describes workflow dependencies. \EARevision{Figure~\ref{fig:dag-example} shows one example. Complete DAGs for all 63 base tasks are provided in Appendix~\ref{app:all_dags}.} The action space is $\mathcal A$, and $\mathcal J$ is the evaluator. Nodes denote operations or intermediate artifacts, and edges denote dependencies.

An information perturbation operator constructs paired public packets \EARevision{$\mathcal P_v=\Phi_v(D,\theta)$}, containing the initial instructions and agent-accessible materials. Variants hide facts, add distractors, introduce conflicting statements, or combine these changes while preserving the business objective and authoritative facts.

\begin{figure}[H]\centering
\includegraphics[width=\linewidth]{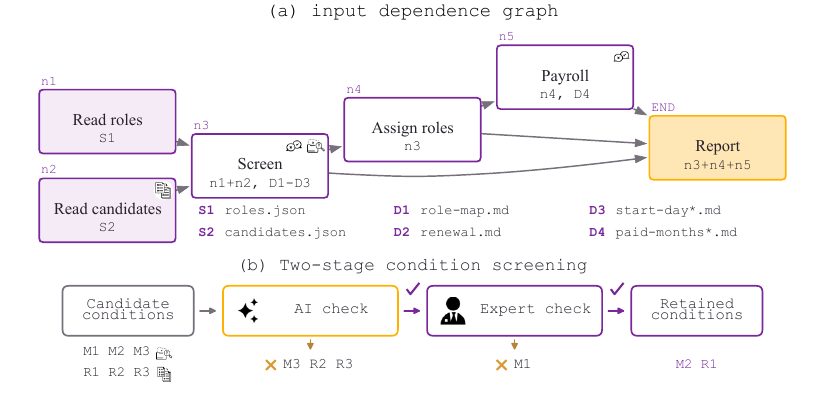}
\caption{\EARevision{\textbf{Input dependencies and candidate perturbation sites for EA-Bench.} (a) shows an example workflow DAG for \texttt{JB31}. (b) illustrates how the two-stage checker filters candidate conditions.}}
\label{fig:dag-example}
\end{figure}

\paragraph{Execution and terminal grading.}

\EARevision{In the standard agent loop, the agent selects $a_t$ from history $h_t$ at step $t$, after which the environment updates state $s_t$ and returns observation $o_{t+1}$:}
{\color{EARevisionBlue}
\begin{equation}
a_t\sim\pi(\cdot\mid h_t),\qquad
(s_{t+1},o_{t+1})\sim P(\cdot\mid s_t,a_t),
\label{eq:agent-loop}
\end{equation}}
Here, $\pi$ is the agent policy and $P$ the environment transition kernel. Actions include business-tool calls (\eaIdentifier{tool\_call}), user queries (\eaIdentifier{ask\_user}), and final submission. The history records instructions, actions, and observed feedback.

\EARevision{Once submission or budget exhaustion terminates the attempt, the evaluator $\mathcal J$ scores the terminal state, artifacts, and interaction records.} Final grades are not returned for repair within the same attempt.

\paragraph{Knowledge-bounded user simulation.}

To limit answer leakage and measure whether agents query key missing or conflicting facts, the user simulator answers from predefined task knowledge. It returns query-relevant facts without providing a complete solution.

Let $R_t$ contain requested information items and $F_t$ the facts actually returned at step $t$. Cumulative user clarification and total information requests satisfy
{\color{EARevisionBlue}
\begin{equation}
K_{t+1}=K_t\cup F_t,\qquad Q=\sum_t |R_t|.
\label{eq:user-information-accounting}
\end{equation}}
On steps without a user query, $R_t=F_t=\varnothing$. \EARevision{Because only returned facts enter $K_t$, a requested but unanswered item does not count as acquired user knowledge. When a query contains several atomic requests, $R_t$ records the distinct items, merging repetitions within that step.} We combine these records with public evidence actually obtained by the agent to assess fact confirmation, grounding, and information efficiency.

\section{Benchmark Construction and Efficiency Measures}
\label{sec:construction}

\subsection{Task Sources and Instruction Variants}
\label{sec:selected-tasks}

To preserve the solvability of the original work-related tasks, we adapt work-related scenarios from three mainstream work-related benchmarks, JobBench, AppWorld-UL, and WorkBench \citep{jobbench,chen2026appworldul,workbench}, and construct Missing, Redundant, Conflict, and combined variants through DAG-guided counterfactual analysis.

\EARevision{Furthermore, we use a two-stage checking process to ensure the validity and realism of the perturbations. First, AI-assisted checks assess missing-fact inferability, conflict authority, and preservation of task requirements and facts under redundancy. We use counterfactual and executable checks to test validity. Second, human experts screen for plausible user communication. We ultimately retain 232 of 792 candidate conditions across 63 base tasks, comprising 169 imperfect variants and 63 Full baselines. The task catalog appears in Appendix~\ref{sec:task-catalog}.}

\paragraph{Task difficulty and scoring.} To separate task complexity from the additional burden of imperfect information, we divide the base tasks into \texttt{Easy} and \texttt{Hard} tracks. \EARevision{\texttt{Easy} contains 33 tasks and 147 conditions, primarily structured application and office workflows. \texttt{Hard} contains 30 tasks and 85 conditions involving databases, multi-file analysis, and complex artifacts.} The division follows task structure and deliverables, not observed model accuracy.

\EARevision{\texttt{Easy} uses task pass rate over its evaluation conditions. \texttt{Hard} uses a weighted rubric score.} For \texttt{Hard} condition $i$,
{\color{EARevisionBlue}
\begin{equation}
s_i=100\frac{\sum_r w_{ir}z_{ir}}{\sum_r w_{ir}},
\label{eq:rubric-score}
\end{equation}}
where $w_{ir}$ is the weight of rubric $r$ and $z_{ir}\in\{0,1\}$ indicates whether that rubric receives credit. The overall \texttt{Hard} score is the equally weighted mean across conditions.

\begin{table}[!htbp]\centering\small
\caption{Task Specifications and Instruction Variants of EA-Bench.}
\label{tab:selected-inventory}
\begingroup\setlength{\tabcolsep}{2.5pt}
\EAfitTable{\begin{tabular}{@{}lcrrrrrrrrrrrrrrrrr@{}}\toprule
\multicolumn{1}{c}{\multirow{2}{*}[-0.5ex]{Track}} & \multirow{2}{*}[-0.5ex]{DAG nodes} & \multicolumn{1}{c}{\multirow{2}{*}[-0.5ex]{\texttt{Full}}} & \multicolumn{4}{c}{\texttt{M}} & \multicolumn{3}{c}{\texttt{R}} & \multicolumn{4}{c}{\texttt{C}} & \multicolumn{4}{c}{\texttt{Mixed}} & \multicolumn{1}{c}{\multirow{2}{*}[-0.5ex]{All}} \\
\cmidrule(lr){4-7}\cmidrule(lr){8-10}\cmidrule(lr){11-14}\cmidrule(lr){15-18}
 & & & \texttt{M1} & \texttt{M2} & \texttt{M3} & \texttt{M4} & \texttt{R1} & \texttt{R2} & \texttt{R3} & \texttt{C1} & \texttt{C2} & \texttt{C3} & \texttt{C4} & \texttt{M1C1} & \texttt{M2C2} & \texttt{M3C3} & \texttt{M1R1C1} &  \\ \midrule
\texttt{Easy} & 4--13 & 33 & 23 & 9 & 13 & 1 & 4 & 2 & 6 & 23 & 10 & 6 & 3 & 9 & 2 & 3 & 0 & 147 \\
\texttt{Hard} & 13--30 & 30 & 7 & 2 & 1 & 0 & 7 & 8 & 16 & 7 & 1 & 1 & 0 & 4 & 0 & 0 & 1 & 85 \\
\rowcolor{EAOverallTint!25}\texttt{Total} & 4--30 & 63 & 30 & 11 & 14 & 1 & 11 & 10 & 22 & 30 & 11 & 7 & 3 & 13 & 2 & 3 & 1 & 232 \\
\bottomrule\end{tabular}}
\endgroup
\end{table}

\subsection{Efficiency Measures and Lower Bounds}
\label{sec:efficiency-measures}

We measure signed tool-call and information-request gaps, $E_t$ and $E_i$, against task-specific references.

\paragraph{Information-request efficiency.} Let $Q$ be the number of requested atomic facts and $I^{\mathrm{ref}}=c(w)$ the minimum certificate identifying the required output in the finite checked world $w$. We define
\begin{equation}
E_i=Q-I^{\mathrm{ref}}.
\label{eq:information-excess}
\end{equation}
\EARevision{We determine $I^{\mathrm{ref}}$ from output differences across finite counterfactual worlds. One answer may make other facts irrelevant to the decision, so this reference need not equal the initial number of missing or conflicting facts.}

\EARevision{Because a single user interaction may request multiple information items, we measure information-request cost at the item level. To support this accounting, the simulator returns answered fact identifiers and values in a structured format, while request parsing separately records the queried items. This distinguishes the requested set $R_t$ from the returned set $F_t$, allowing us to count total information requests as $Q=\sum_t|R_t|$.}

\paragraph{Tool-call efficiency.} \EARevision{For the closed-state protocol, we determine $K$ when defining the DAG by mapping mandatory reading, artifact-generation, and submission operations to nodes $V_{\mathrm{req}}$, giving $K=|V_{\mathrm{req}}|$. Since dependency checks enforce every mandatory operation, completing the task in this protocol requires at least $K$ operations. Given the necessary facts, topological reference execution attains this bound.}

\EARevision{For each attempt, let $A$ count recorded tool calls, including user requests. We retain $A$ for both successful and failed attempts, so budget-exhausted runs also contribute their actual cost. To distinguish required work from extra execution cost, we compare $A$ with a reference for the same task condition. Since necessary facts can be requested together,}
\begin{equation}
A^{\mathrm{ref}}=K+\mathbf{1}[I^{\mathrm{ref}}>0],\qquad E_t=A-A^{\mathrm{ref}}.
\label{eq:tool-excess}
\end{equation}
We calculate each gap from the same episode counts used in the actual-cost figures, then average the paired differences. $E_t$ uses all closed-state attempts with an action reference; $E_i$ uses all \texttt{Easy} attempts with complete request accounting and a finite information reference. A negative gap means fewer calls or requests were recorded than the reference, which can occur when execution stops early or information remains unresolved. Since positive and negative gaps can cancel, Figure~\ref{fig:all-outcome-efficiency} shows their contributions separately. \texttt{Hard} lacks a validated global reference, so its gaps remain unavailable while its actual costs are retained.

\subsection{Models and Configuration}
\label{sec:protocol}
\EARevision{We use a custom runner to call model APIs and evaluate nine mainstream open- and closed-weight model services across all 232 conditions.} Instructions, files, and user-facing tool messages are in English. Each attempt permits at most 100 actions and 16,384 output tokens per model response. \EARevision{The knowledge-bounded user simulator follows the design of AppWorld-UL \citep{chen2026appworldul}. In our experiments, the user simulator and artifact judge both use \texttt{GPT-5.5} at medium effort.}

\setlength{\textfloatsep}{8pt plus 2pt minus 2pt}
\setlength{\floatsep}{6pt plus 2pt minus 2pt}
\setlength{\intextsep}{8pt plus 2pt minus 2pt}
\raggedbottom
\section{Experimental Results}
\label{sec:results}

This section addresses four complementary questions: \eaQuestion{1}~How does imperfect information affect agents' task completion? \eaQuestion{2}~How does imperfect information affect agents' execution efficiency? \eaQuestion{3}~How do task difficulty and perturbation severity affect model performance and efficiency? \eaQuestion{4}~What are the main reasons for model failures under imperfect information?

\paragraph{\eaQuestion{1} Task completion under imperfect information.}
To expose this distinction, we report total and grounded credits separately. On the one hand, imperfect variants in the \texttt{Easy} track reduce grounded credit by 5.9--40.4 percentage points and total credit by 5.3--24.5 percentage points relative to task-matched Full baselines, so information imperfections affect both completion and the reliability of successful outputs. More notably, \textbf{22.7\% of task completions in the \texttt{Easy} track exhibit ungrounded credit.} Specifically, models submit correct answers without fully reviewing all task-related dependency files. Although their outputs can pass the outcome checks, the agents still engage in guessing.

\begin{table}[!htbp]\centering\small
\setlength{\extrarowheight}{0pt}
\caption{\textbf{EA-Bench evaluation results, sorted by Overall total within each track.} Each cell shows total score and grounded score (G). Purple: grounded pass rate (\texttt{Easy}) or rubric score (\texttt{Hard}). Orange: ungrounded score.}
\label{tab:selected-performance}
\par\noindent\makebox[\linewidth][c]{\includegraphics{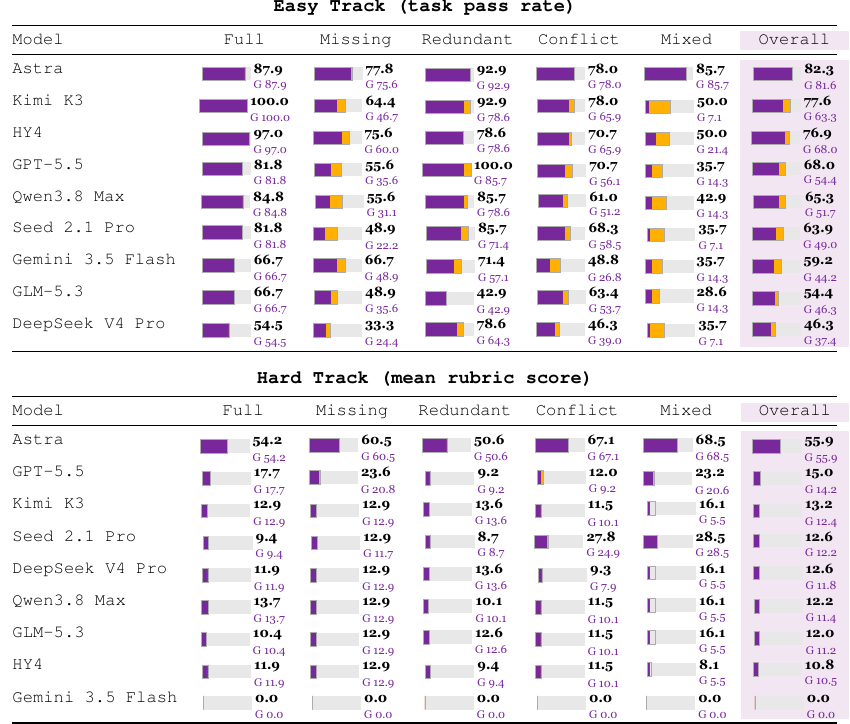}}
\end{table}

\begin{figure}[!htbp]\centering
\includegraphics[width=\linewidth]{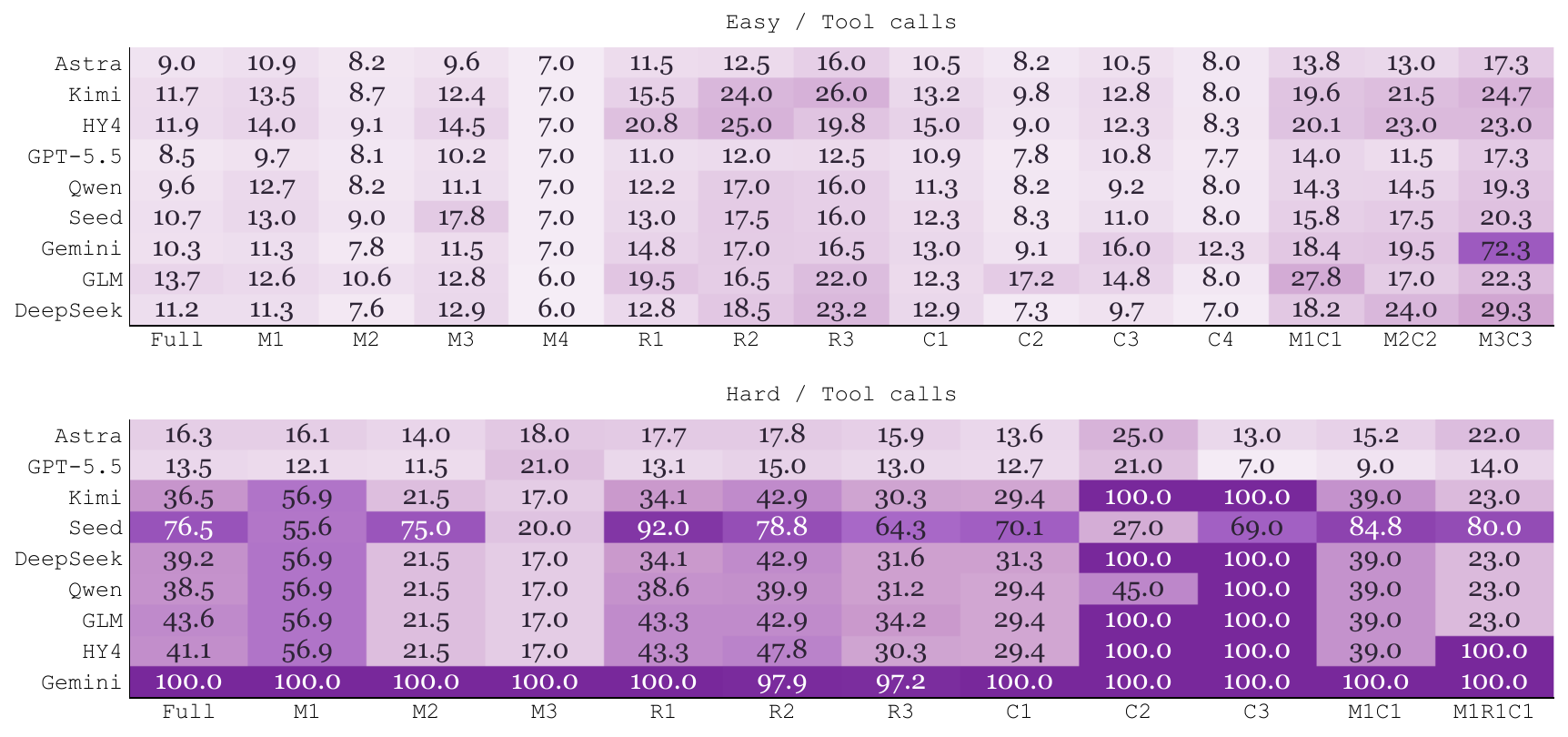}
\caption{\textbf{How many tool calls agents make.} Each cell shows the mean number of tool calls per attempt for the corresponding model and instruction condition, including successful and failed attempts. Full denotes the complete-information baseline.}
\label{fig:condition-actions}
\end{figure}

\begin{figure}[H]\centering
\includegraphics[width=\linewidth,height=.38\textheight,keepaspectratio]{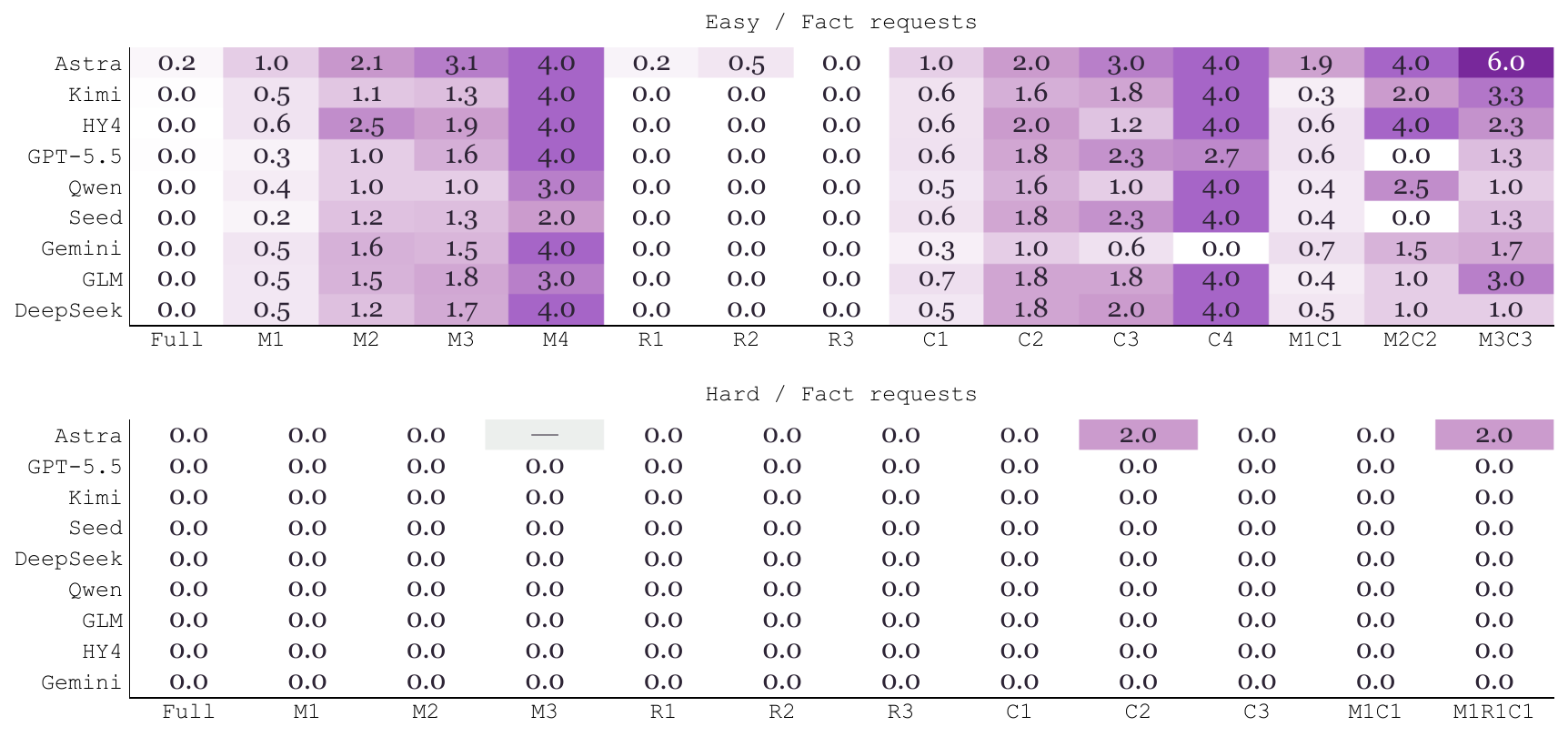}
\caption{\textbf{How much information agents request from users.} Cells show mean \textbf{requested information items} per attempt with complete records. \textbf{User-interaction calls} are shown in Figure~\ref{fig:condition-user-calls}.}
\label{fig:condition-facts}

\end{figure}

\paragraph{\eaQuestion{2} Execution efficiency under imperfect information.}
To measure requests introduced by imperfect conditions, we average the perturbation-column means in Figures~\ref{fig:condition-actions} and~\ref{fig:condition-facts} and compare them with each model's Full-column mean, including successes and failures. The results show that \textbf{tool calls increase by 14.4--71.5\% across models on \texttt{Easy}}. \textbf{Information requests under imperfect instructions increase from approximately 0.0 to 1.4 on the \texttt{Easy} track.}

\begin{figure}[H]\centering
\includegraphics[width=\linewidth]{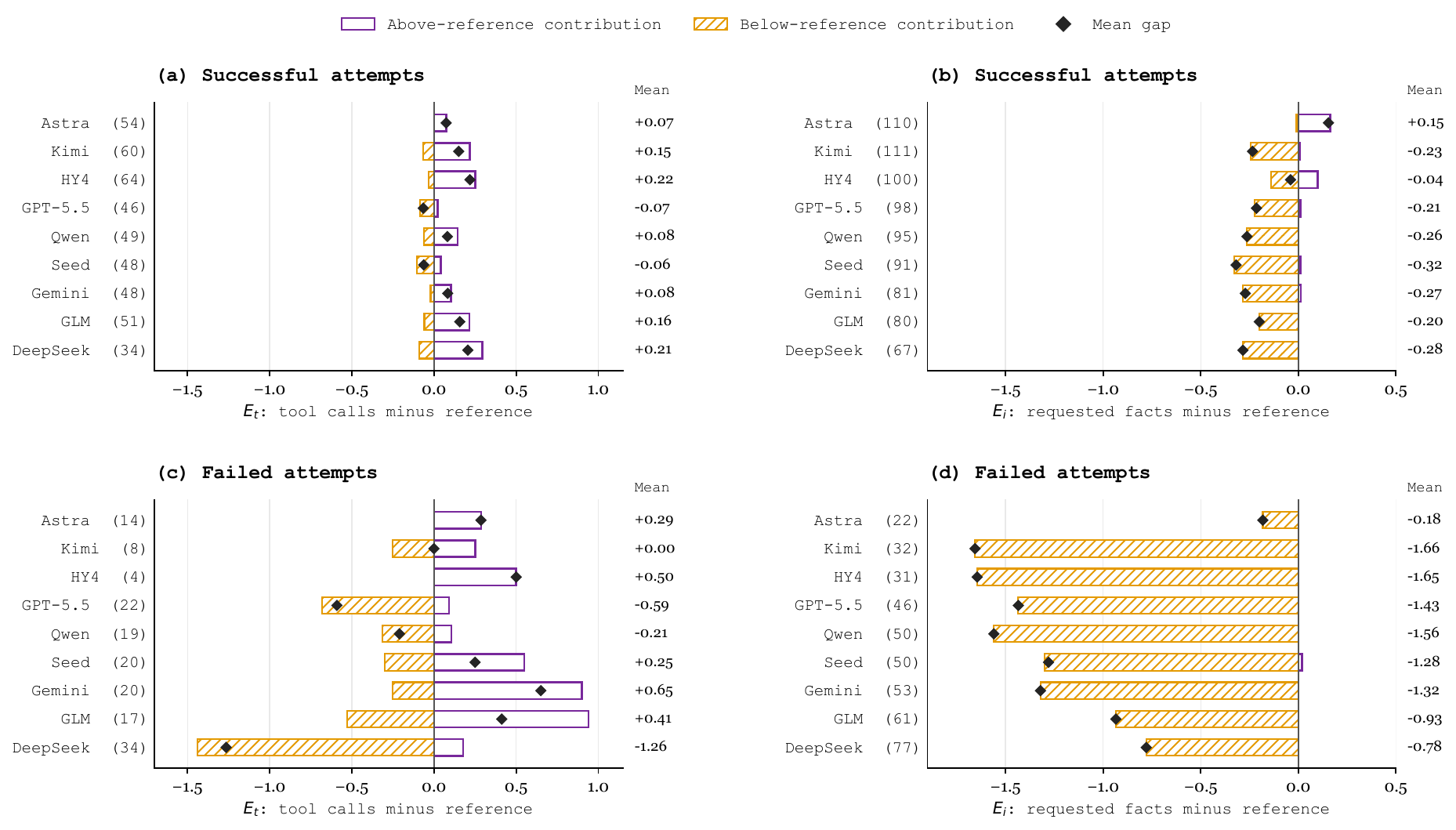}
\caption{\textbf{Signed execution gaps relative to references in \texttt{Easy}.} Successful (top) and failed (bottom) attempts are shown separately for $E_t$ (left) and $E_i$ (right). Diamonds mark means; parentheses give sample sizes.}
\label{fig:all-outcome-efficiency}
\end{figure}

Furthermore, Figure~\ref{fig:all-outcome-efficiency} separates $E_t$ and $E_i$ by task outcome. Failed attempts have negative mean $E_i$ for all nine model rows, consistently below their successful counterparts, indicating fewer information requests than the reference requires. Mean $E_t$ varies across models in both groups, showing that failure can occur with either additional tool use or fewer operations than the reference.

\paragraph{\eaQuestion{3} Task difficulty and perturbation severity.}
For harder tasks, limits in model capability can reduce the apparent sensitivity to imperfect instructions. Even under Full instructions, some models cannot complete any \texttt{Hard} task (Table~\ref{tab:selected-performance}). Because their scores are already near the lower bound, there is little room for further degradation. A small perturbation penalty can therefore reflect a \textbf{performance floor effect}. Tool-call cost is also constrained by the action budget. For example, Gemini 3.5 Flash averages 13.35 actions on the \texttt{Easy} track and 99.28 on the \texttt{Hard} track, close to the action limit. By contrast, imperfect information can markedly reduce completion on easier tasks that are otherwise solvable under Full instructions.

\begin{figure}[!ht]\centering
\includegraphics[width=\linewidth]{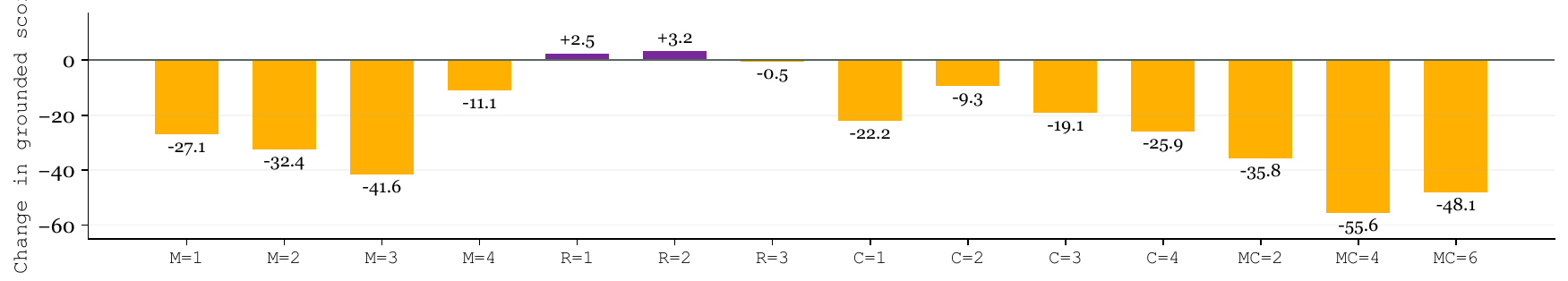}
\caption{\textbf{Detailed score changes under imperfect conditions relative to Full.} Bars pool normalized grounded-score changes across both tracks and all nine model rows, with equal task weights.}
\label{fig:analysis-severity}
\par\vspace{6pt}

\includegraphics[width=\linewidth]{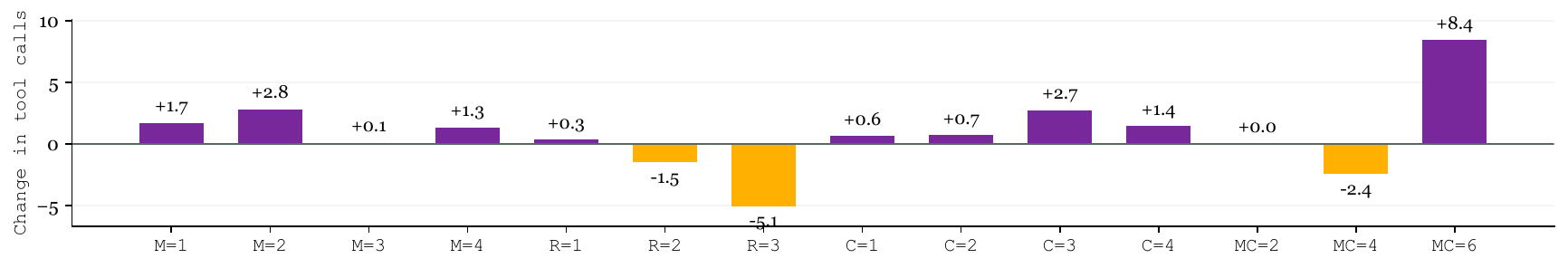}
\caption{\textbf{Detailed tool-call changes under imperfect conditions relative to Full.} Aggregation and condition labels follow Figure~\ref{fig:analysis-severity}.}
\label{fig:tool-call-changes}
\end{figure}

Regarding perturbation severity, Figure~\ref{fig:analysis-severity} shows instances of \textbf{lower grounded scores at higher perturbation levels for both missing and conflicting information}, while mixed conditions further amplify the score decline. In contrast, redundant information preserves all authoritative facts needed by the model to solve the task and therefore has little impact on task completion. However, it adds material that the model needs to inspect, primarily making execution less efficient.

\begin{figure}[!ht]\centering
\includegraphics[width=\linewidth]{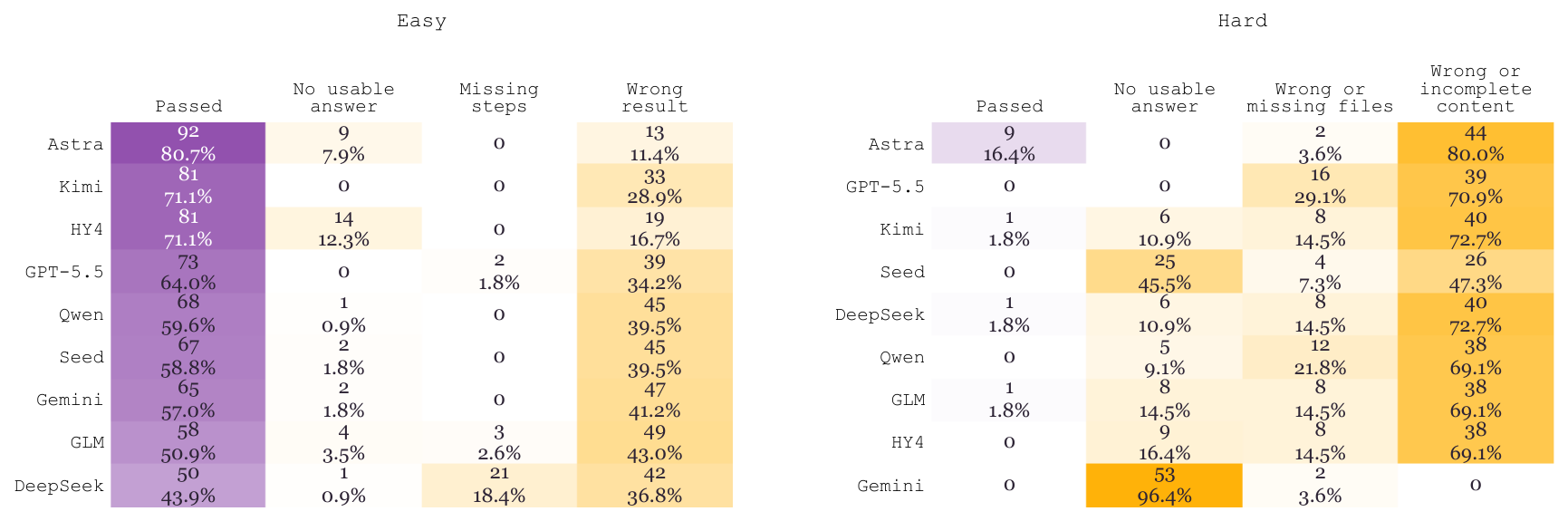}
\caption{\textbf{Model outcomes under imperfect instructions.} Cells show counts and percentages within each track, excluding Full.}
\label{fig:analysis-outcomes}
\end{figure}

\paragraph{\eaQuestion{4} Main failure reasons under imperfect information.}
To identify where execution breaks down, Figure~\ref{fig:analysis-outcomes} groups failed checks under imperfect conditions. Wrong results dominate \texttt{Easy} failures, while \texttt{Hard} failures concentrate on wrong or incomplete content and unusable submissions. To explain whether agents lack information or fail to use it correctly, we combine clarification and grading records to identify accumulated patterns of recurring failure behavior.

\FloatBarrier

One failure mode is shown in Box 1 below: three models make definitive decisions about unresolved information without asking the user. Interestingly, their corresponding runs under the Full condition all pass, whereas the quoted runs under imperfect conditions all fail task-relevant checks.

\begin{tcolorbox}[colback=white,colframe=black!45,colbacktitle=black!5,coltitle=black,
title={Typical failure mode 1: resolving uncertain inputs without asking},fonttitle=\bfseries\small,fontupper=\footnotesize,
boxrule=.5pt,arc=1pt,left=5pt,right=5pt,top=4pt,bottom=4pt,before skip=5pt,after skip=5pt]
\textbf{GPT-5.5 / \texttt{AW02/C1}.} A camping purchase depends on whether Sam's promised supplies should count.
\par\textit{Reply excerpt:} ``Sam's conditional pad pledge \textcolor{EAGrounded}{was not counted}.''
\par\textit{Observed result:} The agent \textcolor{EAGrounded}{orders four pads}; the order, linked-payment and inventory \textcolor{EAGrounded}{checks fail}.
\par\smallskip
\textbf{Astra / \texttt{WB07/C1}.} Creating a task depends on whether the latest metric revision is accepted.
\par\textit{Reply excerpt:} ``The pending November 29 revision \textcolor{EAGrounded}{was excluded}.''
\par\textit{Observed result:} \textcolor{EAGrounded}{No task is created}, and the conditional-task \textcolor{EAGrounded}{check fails}.
\par\smallskip
\textbf{HY4 / \texttt{AW06/C1}.} Checkout requires the user's designated payment card.
\par\textit{Reply excerpt:} ``using the designated Discover card \textcolor{EAGrounded}{DISC-B} in a single payment.''
\par\textit{Observed result:} The agent charges \textcolor{EAGrounded}{DISC-B} \textcolor{EAGrounded}{without confirmation}; the order/payment correctness \textcolor{EAGrounded}{check fails}.
\par\smallskip All three corresponding Full runs pass. In the quoted imperfect runs, \textcolor{EAGrounded}{no user query} is recorded.
\end{tcolorbox}

A second efficiency-related failure mode is asking the user for information already available in the task files. The following case succeeds, but incurs an avoidable information request.

\begin{tcolorbox}[colback=white,colframe=black!45,colbacktitle=black!5,coltitle=black,
title={Typical failure mode 2: unnecessary queries for available information},fonttitle=\bfseries\small,fontupper=\footnotesize,
boxrule=.5pt,arc=1pt,left=5pt,right=5pt,top=4pt,bottom=4pt,before skip=5pt,after skip=5pt]
\textbf{Astra / \texttt{AW14/R2}.} The task file states: ``Credit CR2 against bill B02 is \textcolor{EAGrounded}{currently effective}.''
\par\textit{Agent question:} ``\textcolor{EAGrounded}{Has this credit been confirmed and applied, or is it still pending?}''
\par\textit{Observed result:} The task passes, but it repeats a public fact: one request against a zero-fact reference ($E_i=+1$).
\par\smallskip
\textbf{HY4 / \texttt{WB16/M2}.} The supplied document states: ``\textcolor{EAGrounded}{Send an update notification for each rescheduled meeting}.''
\par\textit{Agent question excerpt:} ``\textcolor{EAGrounded}{should update notifications be sent for each rescheduled meeting?}''
\par\textit{Observed result:} A second user turn reconfirms the already supplied duration and notification policies; four facts are requested against a two-fact reference.
\par\smallskip
\textbf{Kimi K3 / \texttt{AW04/Full}.} The task file states: ``Grant's confirmed destination for this purchase is \textcolor{EAGrounded}{G\_ADDR2}.''
\par\textit{Agent question excerpt:} ``\textcolor{EAGrounded}{Which one is Grant's confirmed home address for this delivery?}''
\par\textit{Observed result:} The task passes, but the supplied address is requested again ($E_i=+1$).
\end{tcolorbox}

\section{Conclusion}

In this paper, we introduce EA-Bench, a benchmark that evaluates how personal agents complete work-related tasks when user instructions contain imperfect information in the form of missing, redundant, or conflicting facts. To construct effective and realistic perturbations, EA-Bench selects workflows from existing mainstream work-related benchmarks and abstracts their task dependencies into DAGs. This allows us to identify the information needed for task-relevant decisions and isolate facts to construct plausible imperfect variants. A two-stage process of AI-assisted checks and expert screening further selects 232 conditions from 792 candidates across 63 base-task instances. We divide EA-Bench into \texttt{Easy} and \texttt{Hard} tracks according to the tasks' inherent difficulty. For evaluation, we go beyond task success rates by introducing grounded credit to assess whether agents have reviewed all task-relevant facts, while also measuring the efficiency of tool calls and user interactions.

Our experiments show that imperfect information can substantially reduce task completion and increase action costs by 14.4--71.5\% on the \texttt{Easy} track, while increasing the mean number of requested information items from approximately 0.0 to 1.4. More concerning, 22.7\% of successful \texttt{Easy} task completions receive ungrounded credit, indicating that models guess facts to complete tasks. Overall, EA-Bench provides a useful evaluation setting for improving the efficiency of future personal agents.

\label{mainend}\clearpage\flushbottom\label{statementsstart}
\section*{AI Use Statement}

We used generative AI for language editing, coding and methodological discussion. The authors are responsible for reviewing the text, analyses and artifacts.

\section*{Reproducibility Statement}

Sections~\ref{sec:formulation} and~\ref{sec:construction} define construction and execution. Anonymous analysis inputs reproduce the inventory and reported tables. Task packets, execution identities, per-attempt records and grading configurations are retained. Hosted-service reruns require matching configurations and access. We will release the dataset and evaluation code upon acceptance.
\clearpage\label{referencesstart}
% Pre-generated bibliography for portable arXiv compilation.

\clearpage\appendix\label{appendixstart}\setcounter{table}{0}\renewcommand{\thetable}{A\arabic{table}}\renewcommand{\theHtable}{appendix.\arabic{table}}
\section{Detailed Performance and Structural Analyses}
\label{app:detailed-analysis}

\begin{figure}[!htbp]\centering
\includegraphics[width=\linewidth]{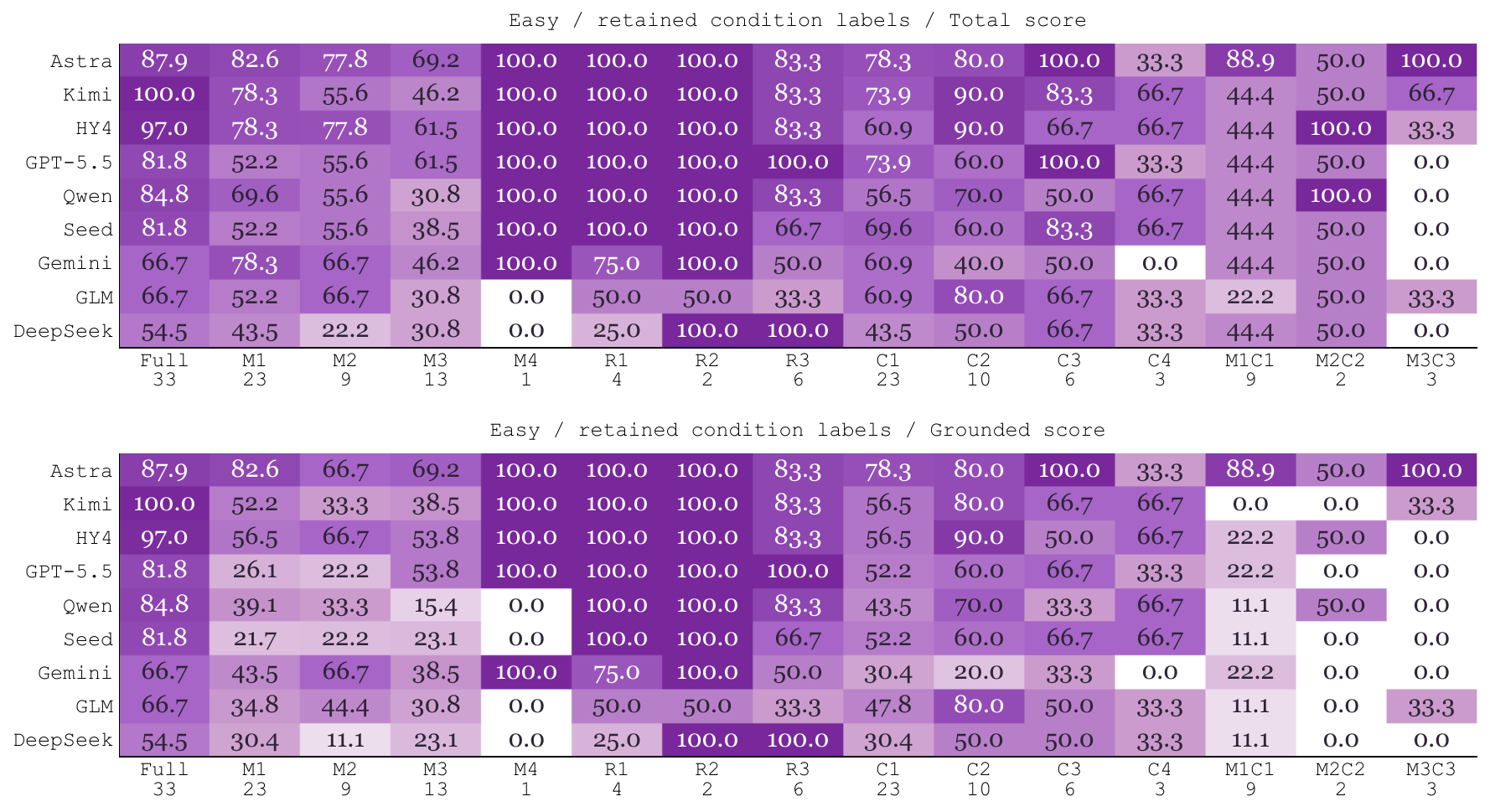}
\caption{\textbf{Models' total and grounded scores in the EA-Bench \texttt{Easy} track.} Row notation follows Table~\ref{tab:selected-performance}.}
\label{fig:conditions-easy}
\end{figure}

\begin{figure}[!htbp]\centering
\includegraphics[width=\linewidth]{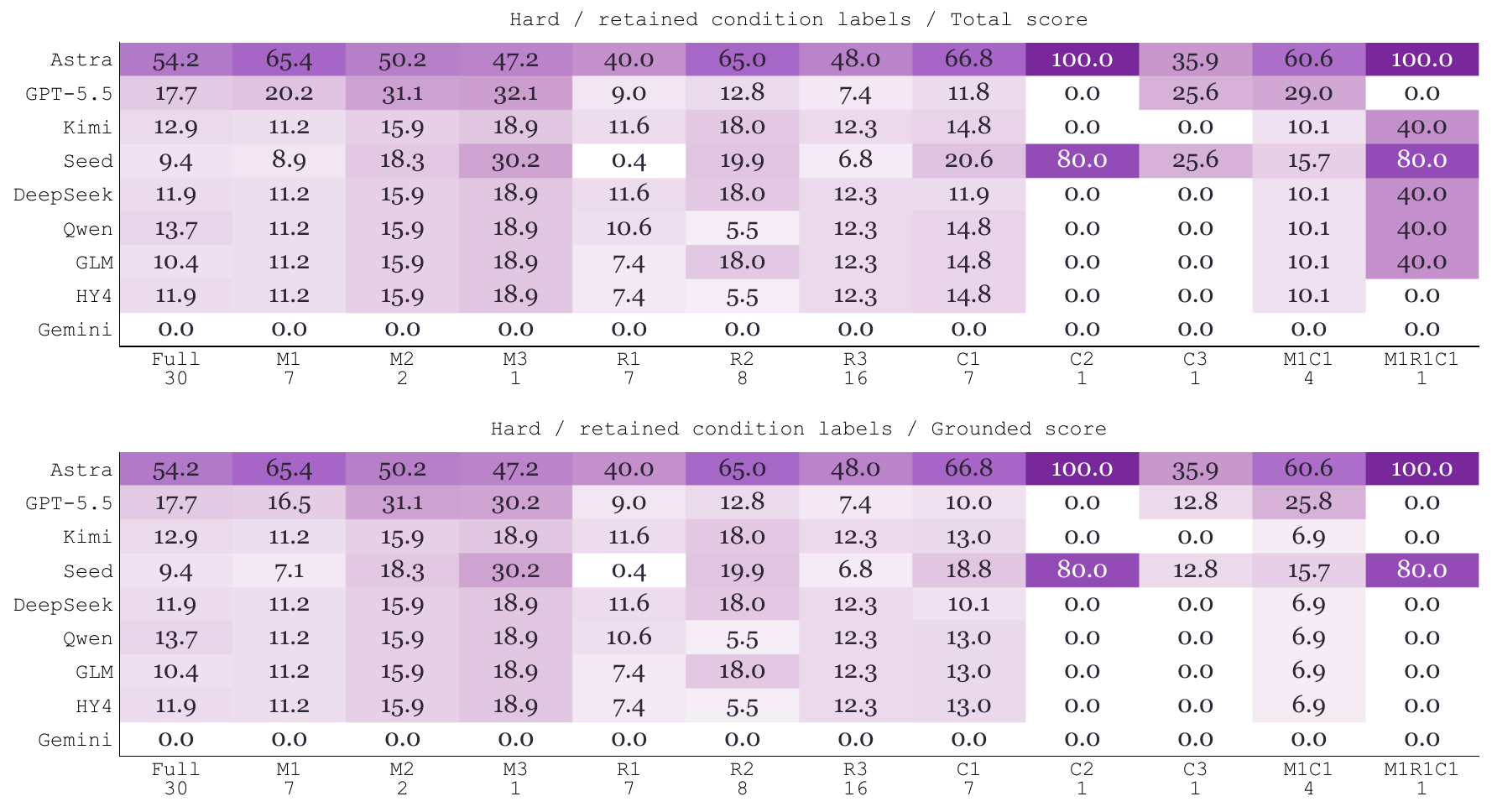}
\caption{\textbf{Models' total and grounded scores in the EA-Bench \texttt{Hard} track.} Row notation follows Table~\ref{tab:selected-performance}.}
\label{fig:conditions-hard}
\end{figure}

\begin{figure}[!htbp]\centering
\includegraphics[width=\linewidth,height=.38\textheight,keepaspectratio]{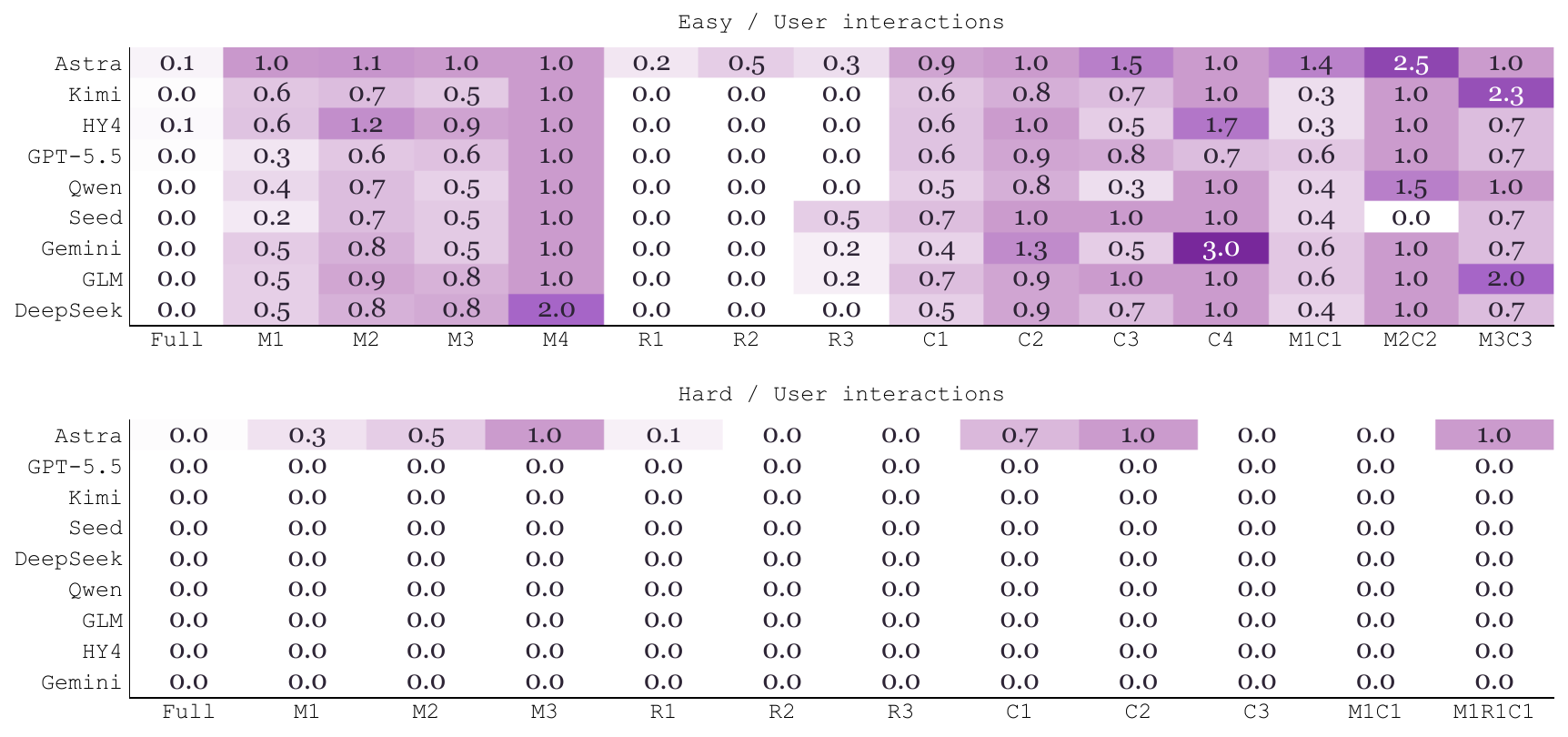}
\caption{\textbf{How often agents ask users.} Cells show mean \textbf{user-interaction calls} per attempt, including successes and failures.}
\label{fig:condition-user-calls}
\end{figure}

\begin{figure}[!htbp]\centering
\includegraphics[width=\linewidth]{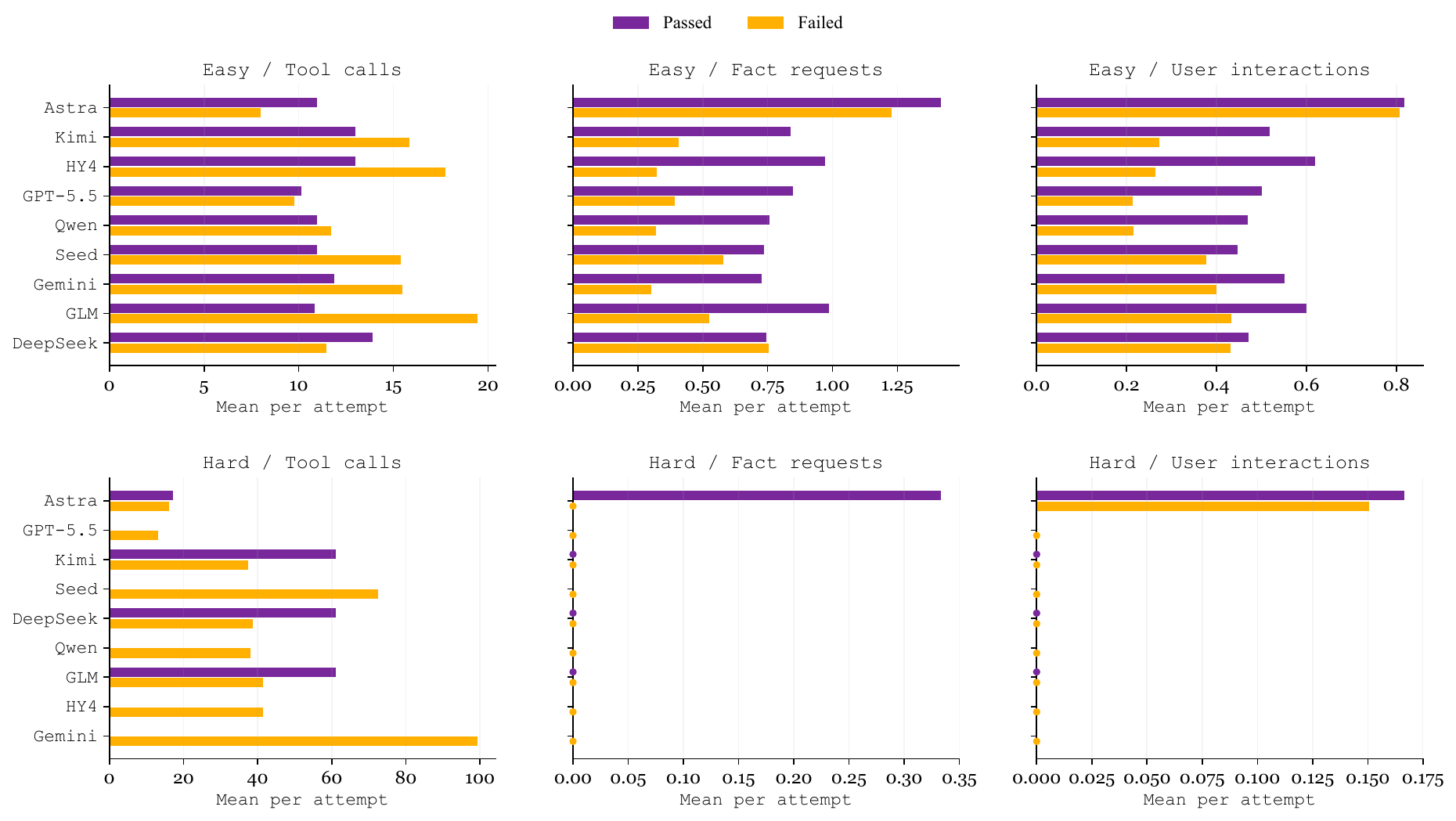}
\caption{\textbf{Detailed cost comparisons separated by successful and failed attempts.}}
\label{fig:costs-by-outcome}
\end{figure}

\EARevision{We organize the analyses of all nine evaluated services into four parts. First, Figures~\ref{fig:conditions-easy} and~\ref{fig:conditions-hard} compare total and grounded scores for each retained condition label in \texttt{Easy} and \texttt{Hard}. To explain the associated costs, Figures~\ref{fig:condition-actions}--\ref{fig:costs-by-outcome} report tool calls, fact requests and user interactions, including both successful and failed attempts. Finally, we report paired performance estimates.}

\subsection{Condition-Level Performance}
\clearpage
\subsection{Tool Use, Fact Requests and User Interactions}

\EARevision{These costs measure different parts of execution. A tool call is an action, while a fact request counts an information item requested from the user. One user interaction can request several items. We therefore report recorded tool calls, requested facts and \texttt{ask\_user} invocations separately. Successful and failed attempts are both included. Tool and user-call counts cover all 2,088 records, while fact-request means use the 2,010 records with complete request ledgers. Missing counts are not replaced with zero.}

\clearpage
\begin{figure}[!htbp]\centering
\includegraphics[width=\linewidth]{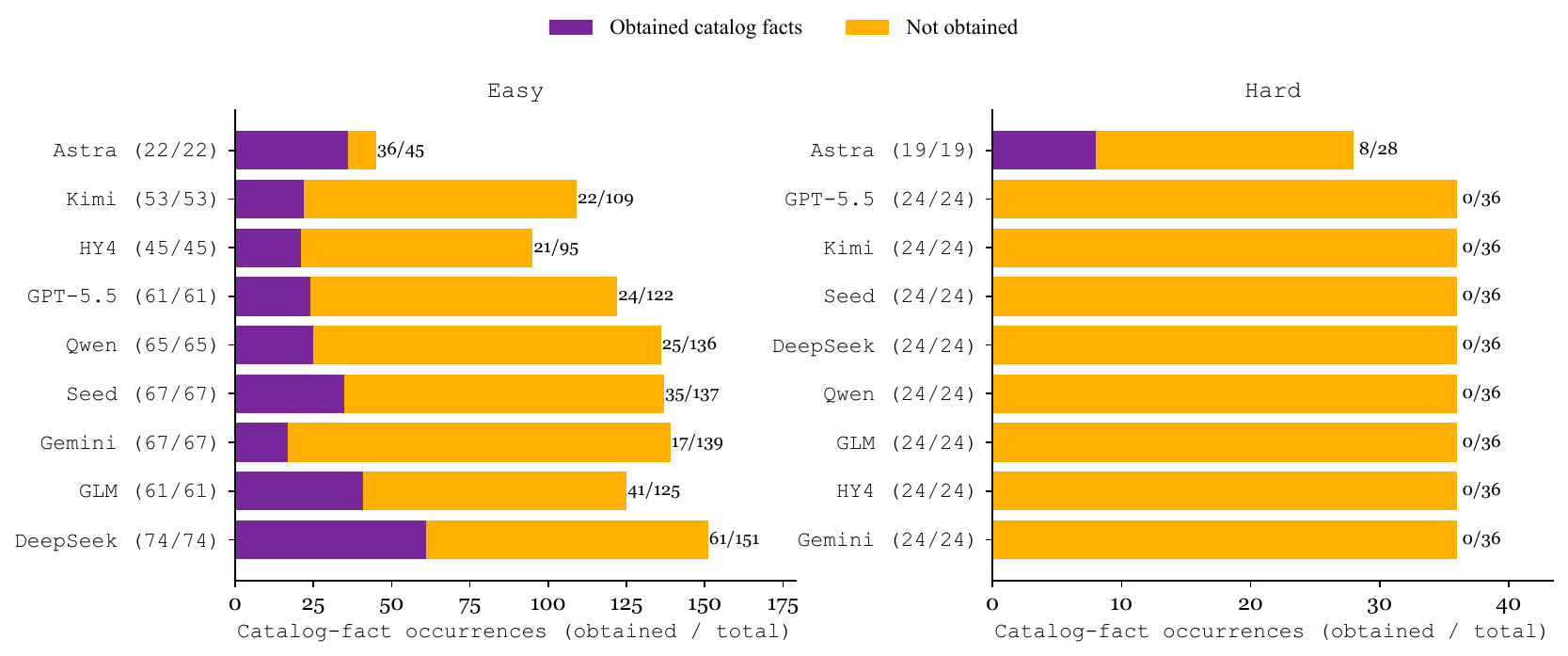}
\caption{Catalog facts obtained and not obtained in unsuccessful or non-fully-grounded outputs with at least one M/C fact.}
\label{fig:analysis-failure-info}
\end{figure}

\subsection{Paired Performance Estimates}
% Small circles: one observed task-level paired difference per base task.
% Their x coordinates are reproduced from selected-release-rows.json.
% Shared scale includes every observation; only vertical position is jittered.
\newcommand{\eaPairInterval}[4]{%
  \begin{tikzpicture}[x=1.65pt,y=1pt,baseline=0pt]
    \path[use as bounding box] (-85,-14) rectangle (45,14);
    \draw[black!20,densely dashed,line width=.4pt] (0,-13) -- (0,13);
    \foreach \x/\y in {#4}{
      \fill[EAGrounded!42] (\x,\y) circle[radius=1.05pt];
    }
    \draw[EAGrounded,line width=1.1pt] (#2,0) -- (#3,0);
    \draw[EAGrounded,line width=.8pt] (#2,-2.5) -- (#2,2.5);
    \draw[EAGrounded,line width=.8pt] (#3,-2.5) -- (#3,2.5);
    \filldraw[fill=EAGrounded,draw=white,line width=.3pt] (#1,2.8) -- ({#1+1.7},0) -- (#1,-2.8) -- ({#1-1.7},0) -- cycle;
    \node[font=\scriptsize\bfseries,anchor=south,inner sep=1pt,fill=white] at (#1,7.5) {#1};
    \node[font=\scriptsize,text=black!65,anchor=north,inner sep=1pt,fill=white] at (#1,-7.5) {[#2, #3]};
  \end{tikzpicture}%
}
\begin{table}[!ht]\centering\small
\caption{\textbf{Matched comparisons across all nine reported evaluation rows.} Small circles: task-level differences from Full. Diamonds: means; bars: 95\% confidence intervals.}
\label{tab:selected-pairs}
\setlength{\tabcolsep}{7pt}
\begin{tabular}{@{}lrrc@{}}\toprule
\texttt{Perturbation} & $n_t$ & $n_c$ & \texttt{Task differences and 95\% CI} \\
 & & & \begin{tikzpicture}[x=1.65pt,y=1pt,baseline=0pt]
  \path[use as bounding box] (-85,-12) rectangle (45,5);
  \draw[black!35,line width=.4pt] (-80,2) -- (40,2);
  \foreach \x/\v in {-80/-80,-60/-60,-40/-40,-20/-20,0/0,20/+20,40/+40}{
    \draw[black!45,line width=.4pt] (\x,0) -- (\x,4);
    \node[font=\scriptsize,text=black!65,anchor=north,inner sep=2pt] at (\x,0) {\v};
  }
\end{tikzpicture} \\ \midrule
\rowcolor{EAOverallTint!18}\multicolumn{4}{@{}l}{\strut\textbf{\texttt{Easy}}} \\
\texttt{Missing} & 29 & 45 & \eaPairInterval{-19.16}{-27.78}{-10.92}{-77.7777777778/0,-66.6666666667/0,-55.5555555556/0,-44.4444444444/0,-44.4444444444/2.4,-44.4444444444/-2.4,-38.8888888889/0,-33.3333333333/0,-33.3333333333/2.4,-22.2222222222/0,-22.2222222222/2.4,-22.2222222222/-2.4,-22.2222222222/4.8,-16.6666666667/0,-11.1111111111/0,-11.1111111111/2.4,-11.1111111111/-2.4,-11.1111111111/4.8,-11.1111111111/-4.8,0.0000000000/0,0.0000000000/2.4,0.0000000000/-2.4,0.0000000000/4.8,0.0000000000/-4.8,0.0000000000/7.2,0.0000000000/-7.2,11.1111111111/0,11.1111111111/2.4,22.2222222222/0} \\
\texttt{Redundant} & 12 & 14 & \eaPairInterval{+1.85}{-5.56}{+10.19}{-22.2222222222/0,-11.1111111111/0,-11.1111111111/2.4,-11.1111111111/-2.4,0.0000000000/0,0.0000000000/2.4,0.0000000000/-2.4,11.1111111111/0,11.1111111111/2.4,11.1111111111/-2.4,11.1111111111/4.8,33.3333333333/0} \\
\texttt{Conflict} & 32 & 41 & \eaPairInterval{-14.93}{-22.74}{-6.94}{-66.6666666667/0,-55.5555555556/0,-55.5555555556/2.4,-55.5555555556/-2.4,-44.4444444444/0,-33.3333333333/0,-22.2222222222/0,-22.2222222222/2.4,-22.2222222222/-2.4,-22.2222222222/4.8,-22.2222222222/-4.8,-22.2222222222/7.2,-22.2222222222/-7.2,-16.6666666667/0,-16.6666666667/2.4,-11.1111111111/0,-11.1111111111/2.4,-11.1111111111/-2.4,-11.1111111111/4.8,-5.5555555556/0,0.0000000000/0,0.0000000000/2.4,0.0000000000/-2.4,0.0000000000/4.8,0.0000000000/-4.8,0.0000000000/7.2,0.0000000000/-7.2,0.0000000000/9.6,11.1111111111/0,11.1111111111/2.4,16.6666666667/0,33.3333333333/0} \\
\texttt{Mixed} & 9 & 14 & \eaPairInterval{-27.57}{-41.36}{-14.20}{-55.5555555556/0,-55.5555555556/2.4,-50.0000000000/0,-37.0370370370/0,-22.2222222222/0,-11.1111111111/0,-11.1111111111/2.4,-5.5555555556/0,0.0000000000/0} \\
\addlinespace[3pt]\midrule
\rowcolor{EAOverallTint!18}\multicolumn{4}{@{}l}{\strut\textbf{\texttt{Hard}}} \\
\texttt{Missing} & 7 & 10 & \eaPairInterval{+3.16}{-3.28}{+10.24}{-11.2179487179/0,-2.2222222222/0,0.2096436059/0,1.7094017094/0,3.0612244898/2.4,11.1111111111/0,19.4444444444/0} \\
\texttt{Redundant} & 25 & 31 & \eaPairInterval{+0.93}{-3.42}{+5.27}{-23.8095238095/0,-17.7777777778/0,-12.0481927711/0,-6.1728395062/0,-6.0606060606/2.4,-6.0606060606/-2.4,-5.4814814815/4.8,-2.7777777778/0,-2.4853801170/2.4,-0.6905210295/0,0.0000000000/2.4,0.0000000000/-2.4,0.0000000000/4.8,0.2710027100/-4.8,1.1251758087/0,1.6516516517/2.4,1.8300653595/-2.4,5.5555555556/0,7.4074074074/0,8.2621082621/2.4,10.4575163399/0,10.8225108225/2.4,11.1111111111/-2.4,14.8148148148/0,33.3333333333/0} \\
\texttt{Conflict} & 8 & 9 & \eaPairInterval{+3.44}{-2.38}{+9.74}{-7.0512820513/0,-6.6666666667/2.4,-5.4507337526/0,1.8140589569/0,5.4131054131/0,8.8888888889/0,11.1111111111/0,19.4444444444/0} \\
\texttt{Mixed} & 5 & 5 & \eaPairInterval{+7.35}{-3.45}{+17.93}{-10.6837606838/0,2.2675736961/0,7.4074074074/0,11.1111111111/0,26.6666666667/0} \\
\bottomrule
\end{tabular}
\par\vspace{4pt}
\begin{minipage}{\linewidth}\footnotesize
\textit{Notes.} All rows share one scale; the dashed line marks zero. Vertical offsets separate overlapping tasks. Bold labels give means; brackets give 95\% limits. Each variant is paired with Full in the same row. Variants are averaged within row--task, then rows within task. $n_t$ counts base tasks; $n_c$ counts conditions per row. The 5,000 source-cluster resamples retain all related observations, including shared executions. Row-based matching does not isolate changes in actual execution configuration.
\end{minipage}
\end{table}

\section{Task Catalog}
\label{sec:task-catalog}

The catalog contains 63 constructed base-task instances, including 33 \texttt{Easy} and 30 \texttt{Hard} tasks. Full is the complete-information baseline. Other labels denote the retained imperfect conditions. Source inspiration does not imply an unchanged native task or an independent source scenario. Known source-linked pairs are disclosed in Section~\ref{sec:selected-tasks}.

Four confirmed source-linked pairs are \texttt{JB31}/\texttt{JB11}, \texttt{JB32}/\texttt{JB30}, \texttt{JB34}/\texttt{JB23} and \texttt{WB20}/\texttt{WB02}. Their authored goals and data differ, and these pairs remain grouped in source-cluster analyses.

\texttt{Easy} includes sixteen AppWorld-UL adaptations, eleven WorkBench adaptations, and six JobBench-inspired closed-state workflows. \texttt{Hard} consists of professional JobBench tasks. For example, JB17 combines 3,930 ITSM records, asset inventory, and project budgets to produce a four-sheet workbook and a gap-register CSV. JB11 combines 20 applicants and 1,470 employee records with Word/Excel outputs.

\begingroup\footnotesize\setlength{\tabcolsep}{3pt}
\begin{longtable}{@{}>{\ttfamily\raggedright\arraybackslash}p{.10\linewidth}>{\ttfamily\raggedright\arraybackslash}p{.075\linewidth}>{\ttfamily\raggedright\arraybackslash}p{.18\linewidth}r >{\ttfamily\raggedright\arraybackslash}p{.49\linewidth}@{}}
\caption{\textbf{Complete instances and conditions in EA-Bench.}}\label{tab:revised-catalog}\\
\toprule
Task & Track & Source & n & Retained conditions \\ \midrule\endfirsthead
\toprule
Task & Track & Source & n & Retained conditions \\ \midrule\endhead
\bottomrule\endfoot
\texttt{AW01} & \texttt{Easy} & AppWorld-UL & 7 & Full, M1, R1, C2, M1C1, M2C2, M3C3 \\
\texttt{AW02} & \texttt{Easy} & AppWorld-UL & 4 & Full, C1, C3, R3 \\
\texttt{AW03} & \texttt{Easy} & AppWorld-UL & 4 & Full, C1, C3, R3 \\
\texttt{AW04} & \texttt{Easy} & AppWorld-UL & 6 & Full, M1, R3, C1, M1C1, M2C2 \\
\texttt{AW06} & \texttt{Easy} & AppWorld-UL & 4 & Full, M1, C1, R3 \\
\texttt{AW08} & \texttt{Easy} & AppWorld-UL & 7 & Full, M1, M3, R3, C1, M1C1, M3C3 \\
\texttt{AW10} & \texttt{Easy} & AppWorld-UL & 5 & Full, M1, M3, C1, R1 \\
\texttt{AW11} & \texttt{Easy} & AppWorld-UL & 6 & Full, M1, M2, C1, R2, M1C1 \\
\texttt{AW12} & \texttt{Easy} & AppWorld-UL & 6 & Full, M1, M3, C1, R3, M1C1 \\
\texttt{AW14} & \texttt{Easy} & AppWorld-UL & 5 & Full, M1, M3, C1, R2 \\
\texttt{AW15} & \texttt{Easy} & AppWorld-UL & 5 & Full, M1, M2, C3, C4 \\
\texttt{AW16} & \texttt{Easy} & AppWorld-UL & 5 & Full, M1, M2, C1, C2 \\
\texttt{AW17} & \texttt{Easy} & AppWorld-UL & 5 & Full, M1, M2, C2, C4 \\
\texttt{AW18} & \texttt{Easy} & AppWorld-UL & 6 & Full, M1, M2, M3, C2, C3 \\
\texttt{AW19} & \texttt{Easy} & AppWorld-UL & 4 & Full, M1, M2, C4 \\
\texttt{AW20} & \texttt{Easy} & AppWorld-UL & 5 & Full, M1, M3, C1, C2 \\
\texttt{JB31} & \texttt{Easy} & JobBench & 3 & Full, M2, C2 \\
\texttt{JB32} & \texttt{Easy} & JobBench & 3 & Full, M2, C1 \\
\texttt{JB33} & \texttt{Easy} & JobBench & 3 & Full, M1, C1 \\
\texttt{JB34} & \texttt{Easy} & JobBench & 3 & Full, M1, C2 \\
\texttt{JB35} & \texttt{Easy} & JobBench & 3 & Full, M1, C2 \\
\texttt{JB36} & \texttt{Easy} & JobBench & 4 & Full, M1, C1, C3 \\
\texttt{WB01} & \texttt{Easy} & WorkBench & 5 & Full, M1, M3, C1, M1C1 \\
\texttt{WB02} & \texttt{Easy} & WorkBench & 5 & Full, M1, C1, M3, M1C1 \\
\texttt{WB07} & \texttt{Easy} & WorkBench & 4 & Full, C1, C3, R1 \\
\texttt{WB12} & \texttt{Easy} & WorkBench & 6 & Full, M1, M3, C1, R1, M1C1 \\
\texttt{WB14} & \texttt{Easy} & WorkBench & 5 & Full, M1, C1, M1C1, M3C3 \\
\texttt{WB15} & \texttt{Easy} & WorkBench & 4 & Full, M1, M3, C1 \\
\texttt{WB16} & \texttt{Easy} & WorkBench & 3 & Full, M2, C2 \\
\texttt{WB17} & \texttt{Easy} & WorkBench & 3 & Full, M3, C1 \\
\texttt{WB18} & \texttt{Easy} & WorkBench & 3 & Full, M3, C1 \\
\texttt{WB19} & \texttt{Easy} & WorkBench & 3 & Full, M4, C2 \\
\texttt{WB20} & \texttt{Easy} & WorkBench & 3 & Full, M3, C1 \\
\texttt{JB01} & \texttt{Hard} & JobBench & 5 & Full, M1, C2, R2, M1R1C1 \\
\texttt{JB02} & \texttt{Hard} & JobBench & 4 & Full, R1, R3, R2 \\
\texttt{JB03} & \texttt{Hard} & JobBench & 2 & Full, R3 \\
\texttt{JB04} & \texttt{Hard} & JobBench & 3 & Full, R1, R3 \\
\texttt{JB05} & \texttt{Hard} & JobBench & 2 & Full, R2 \\
\texttt{JB06} & \texttt{Hard} & JobBench & 2 & Full, R3 \\
\texttt{JB07} & \texttt{Hard} & JobBench & 2 & Full, R1 \\
\texttt{JB08} & \texttt{Hard} & JobBench & 2 & Full, R2 \\
\texttt{JB09} & \texttt{Hard} & JobBench & 2 & Full, R3 \\
\texttt{JB10} & \texttt{Hard} & JobBench & 2 & Full, R3 \\
\texttt{JB11} & \texttt{Hard} & JobBench & 3 & C1, R3, Full \\
\texttt{JB12} & \texttt{Hard} & JobBench & 3 & R1, R3, Full \\
\texttt{JB13} & \texttt{Hard} & JobBench & 5 & C1, M1, M1C1, R1, Full \\
\texttt{JB14} & \texttt{Hard} & JobBench & 1 & Full \\
\texttt{JB15} & \texttt{Hard} & JobBench & 4 & C1, M1, M3, Full \\
\texttt{JB16} & \texttt{Hard} & JobBench & 2 & R2, Full \\
\texttt{JB17} & \texttt{Hard} & JobBench & 4 & C1, M1, R3, Full \\
\texttt{JB18} & \texttt{Hard} & JobBench & 2 & R3, Full \\
\texttt{JB19} & \texttt{Hard} & JobBench & 6 & C1, C3, M1, M1C1, R3, Full \\
\texttt{JB20} & \texttt{Hard} & JobBench & 1 & Full \\
\texttt{JB21} & \texttt{Hard} & JobBench & 2 & R3, Full \\
\texttt{JB22} & \texttt{Hard} & JobBench & 2 & R3, Full \\
\texttt{JB23} & \texttt{Hard} & JobBench & 5 & C1, M1, M1C1, M2, Full \\
\texttt{JB24} & \texttt{Hard} & JobBench & 3 & R1, R2, Full \\
\texttt{JB25} & \texttt{Hard} & JobBench & 3 & R1, R3, Full \\
\texttt{JB26} & \texttt{Hard} & JobBench & 2 & R2, Full \\
\texttt{JB27} & \texttt{Hard} & JobBench & 2 & R3, Full \\
\texttt{JB28} & \texttt{Hard} & JobBench & 2 & R2, Full \\
\texttt{JB29} & \texttt{Hard} & JobBench & 5 & C1, M1, M1C1, M2, Full \\
\texttt{JB30} & \texttt{Hard} & JobBench & 2 & R3, Full \\
\end{longtable}\endgroup

\clearpage\section{Workflow Graphs for All Base Tasks}
\label{app:all_dags}
\EARevision{This appendix provides the complete Full-condition DAGs for all 63 base tasks, covering 18 closed-state, 15 application, and 30 professional workflows.} Node IDs refer to the accompanying descriptions. END denotes final submission. Reference scope is defined in Section~\ref{sec:efficiency-measures}.

\begin{figure}[!ht]\centering

\includegraphics[width=\linewidth,height=.90\textheight,keepaspectratio]{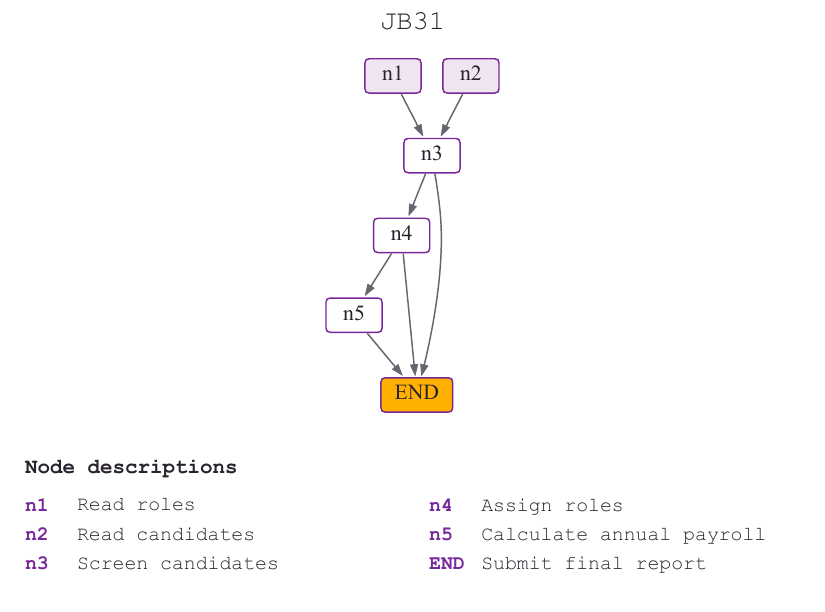}
\caption{DAG for \texttt{JB31}.}
\label{fig:dag-workforce}

\end{figure}

\clearpage

\begin{figure}[p]\centering

\includegraphics[width=.95\linewidth]{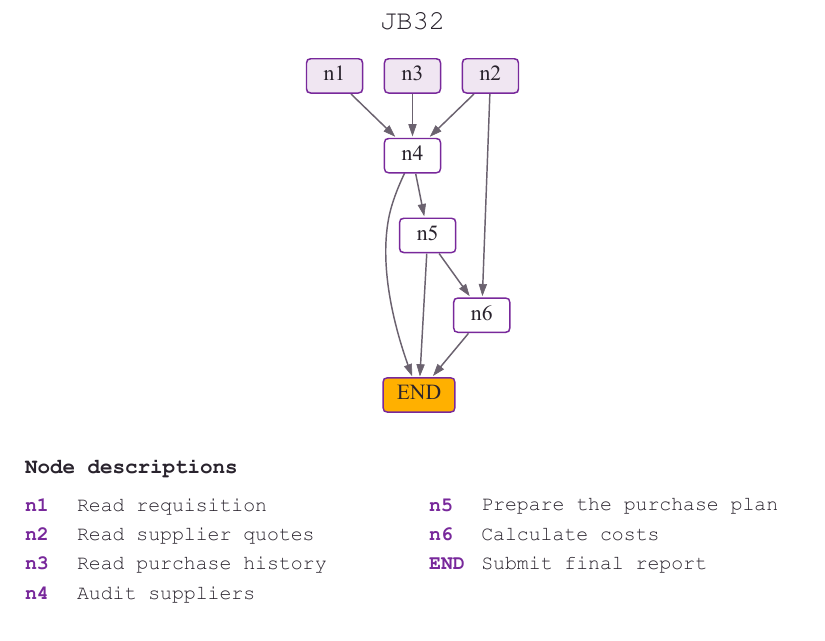}
\caption{DAG for \texttt{JB32}.}
\label{fig:dag-procurement}

\par\vspace{8pt}

\includegraphics[width=.95\linewidth]{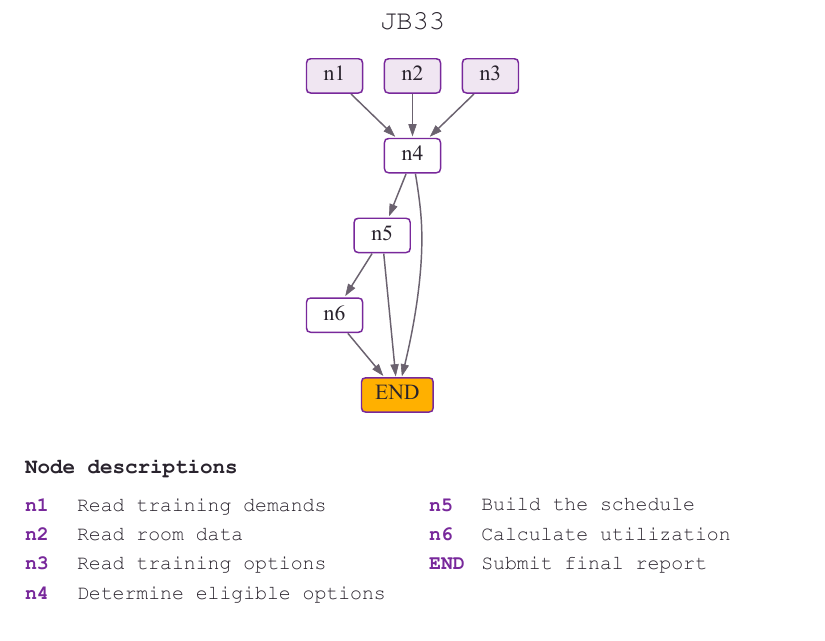}
\caption{DAG for \texttt{JB33}.}
\label{fig:dag-training}

\end{figure}

\clearpage

\begin{figure}[!ht]\centering

\includegraphics[width=\linewidth,height=.90\textheight,keepaspectratio]{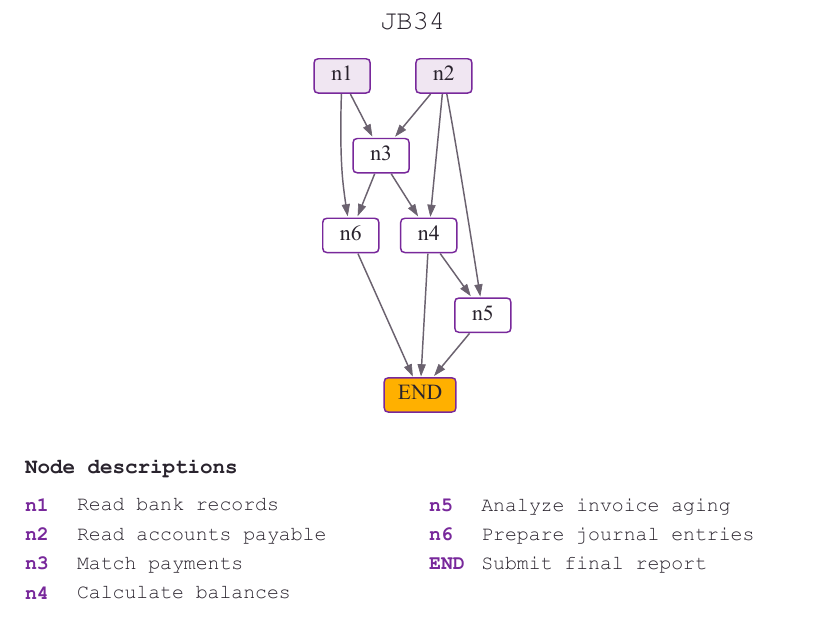}
\caption{DAG for \texttt{JB34}.}
\label{fig:dag-reconciliation}

\end{figure}

\clearpage

\begin{figure}[!ht]\centering

\includegraphics[width=\linewidth,height=.90\textheight,keepaspectratio]{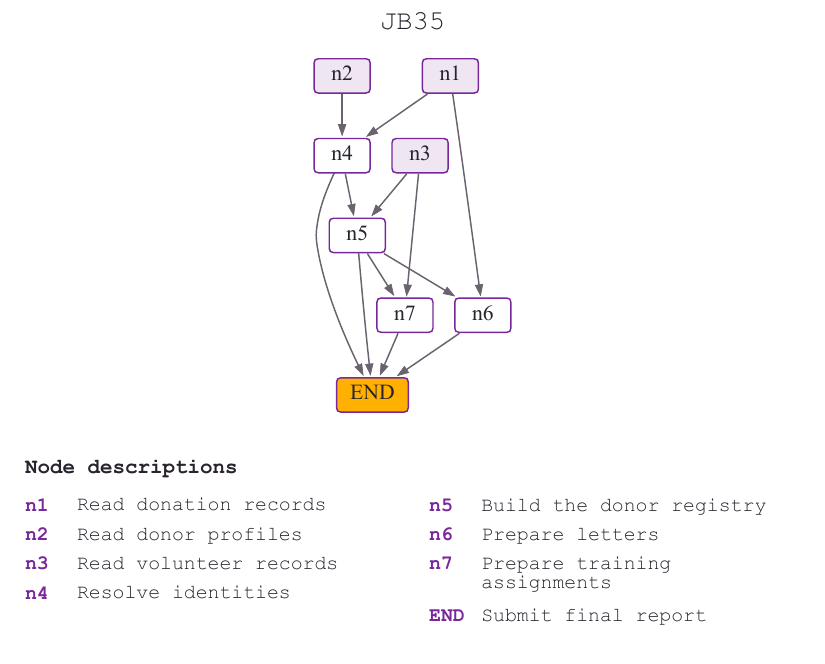}
\caption{DAG for \texttt{JB35}.}
\label{fig:dag-donor_registry}

\end{figure}

\clearpage

\begin{figure}[p]\centering

\includegraphics[width=.95\linewidth]{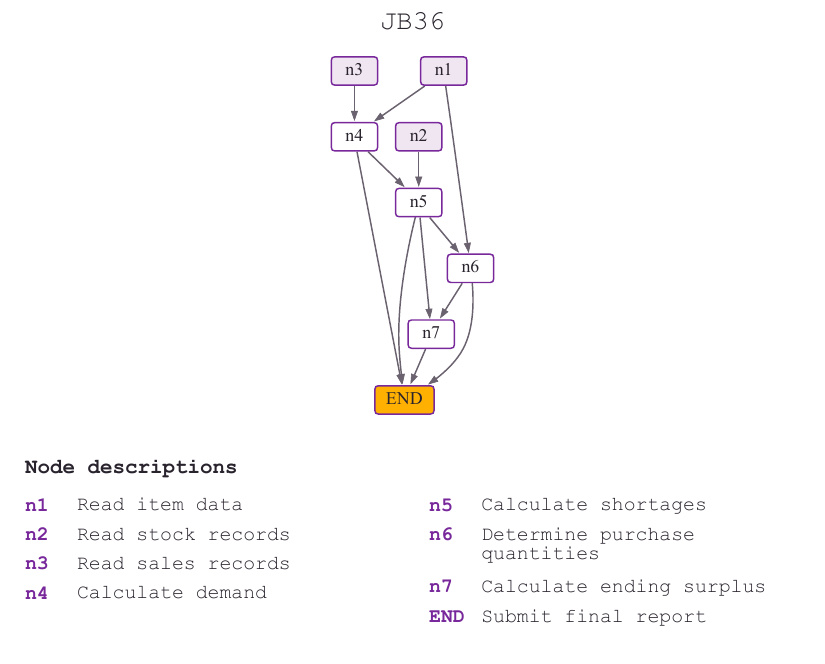}
\caption{DAG for \texttt{JB36}.}
\label{fig:dag-inventory_replenishment}

\par\vspace{8pt}

\includegraphics[width=.95\linewidth]{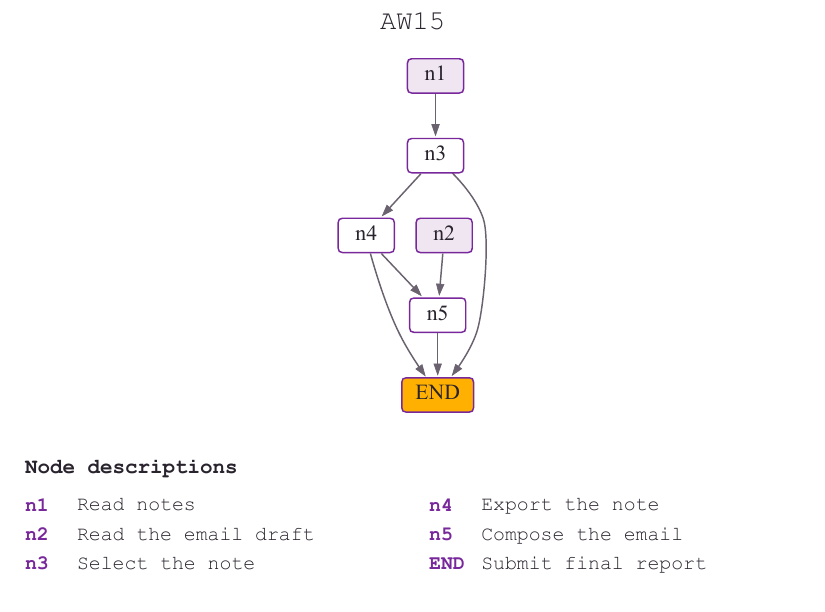}
\caption{DAG for \texttt{AW15}.}
\label{fig:dag-aw_note_mail}

\end{figure}

\clearpage

\begin{figure}[!ht]\centering

\includegraphics[width=\linewidth,height=.90\textheight,keepaspectratio]{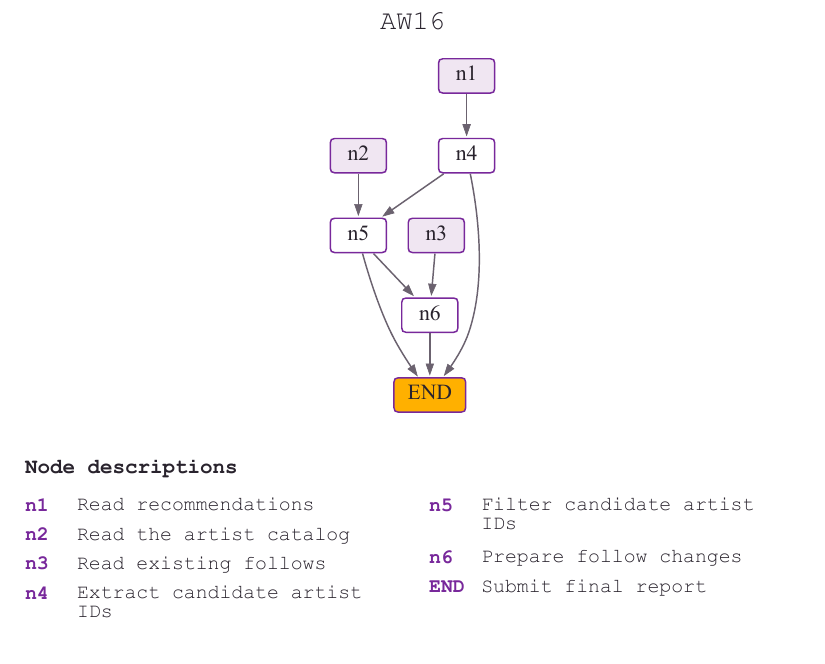}
\caption{DAG for \texttt{AW16}.}
\label{fig:dag-aw_artist_follow}

\end{figure}

\clearpage

\begin{figure}[p]\centering

\includegraphics[width=.95\linewidth]{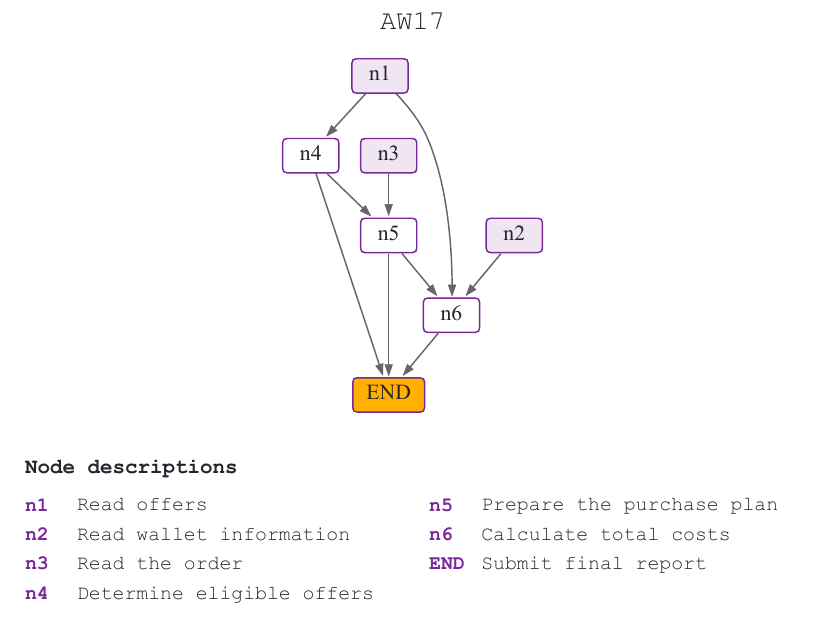}
\caption{DAG for \texttt{AW17}.}
\label{fig:dag-aw_trip_checkout}

\par\vspace{8pt}

\includegraphics[width=.95\linewidth]{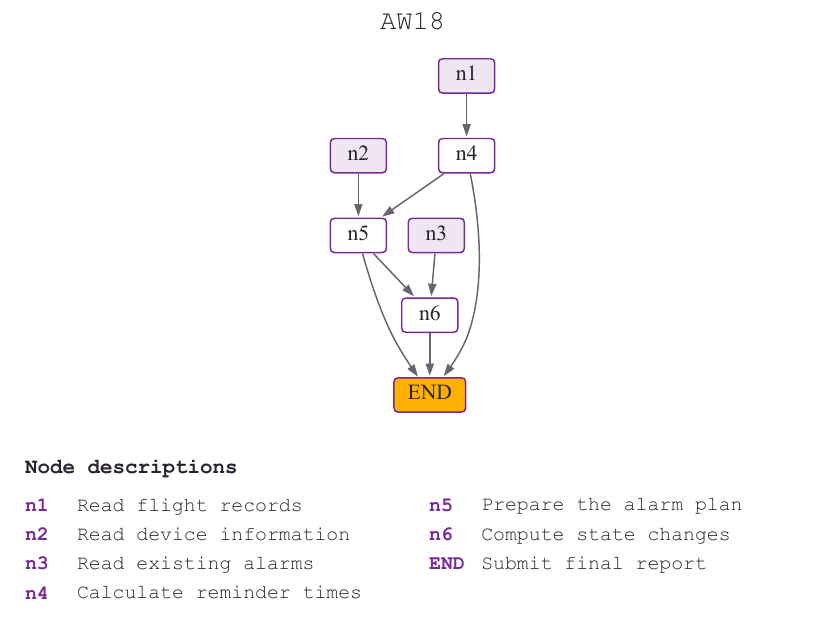}
\caption{DAG for \texttt{AW18}.}
\label{fig:dag-aw_flight_alarms}

\end{figure}

\clearpage

\begin{figure}[p]\centering

\includegraphics[width=.95\linewidth]{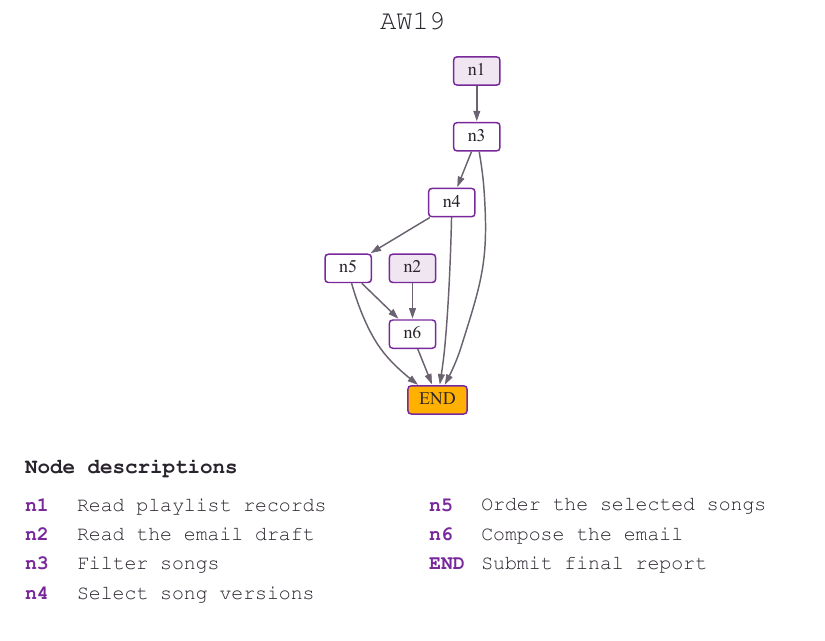}
\caption{DAG for \texttt{AW19}.}
\label{fig:dag-aw_music_mail}

\par\vspace{8pt}

\includegraphics[width=.95\linewidth]{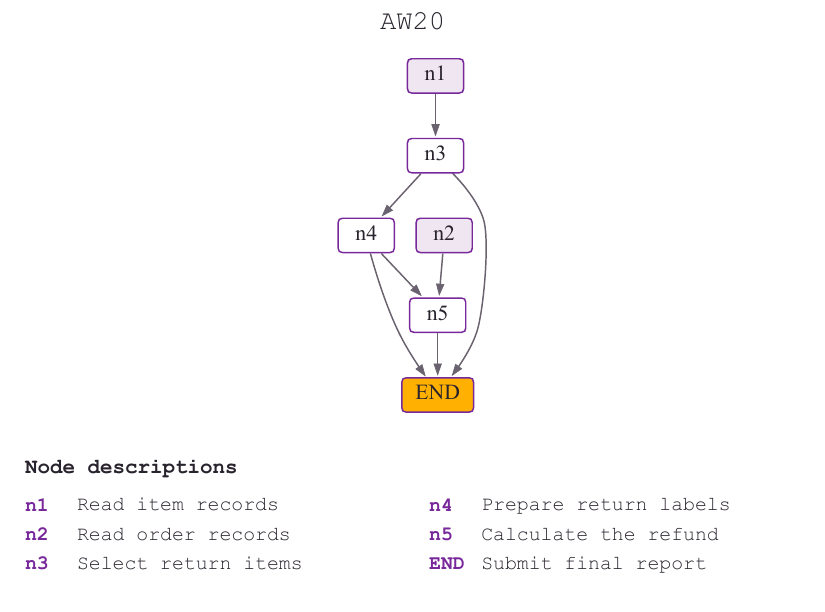}
\caption{DAG for \texttt{AW20}.}
\label{fig:dag-aw_order_returns}

\end{figure}

\clearpage

\begin{figure}[p]\centering

\includegraphics[width=.95\linewidth]{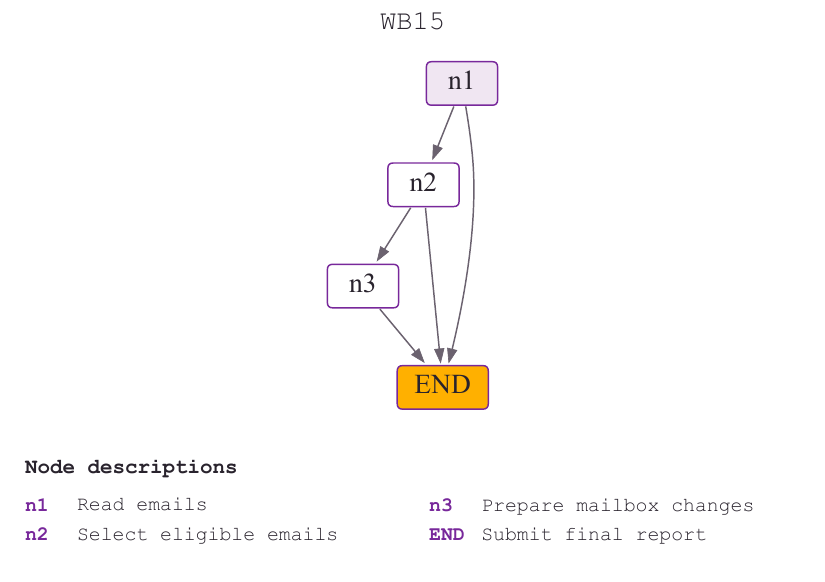}
\caption{DAG for \texttt{WB15}.}
\label{fig:dag-wb_mail_cleanup}

\par\vspace{8pt}

\includegraphics[width=.95\linewidth]{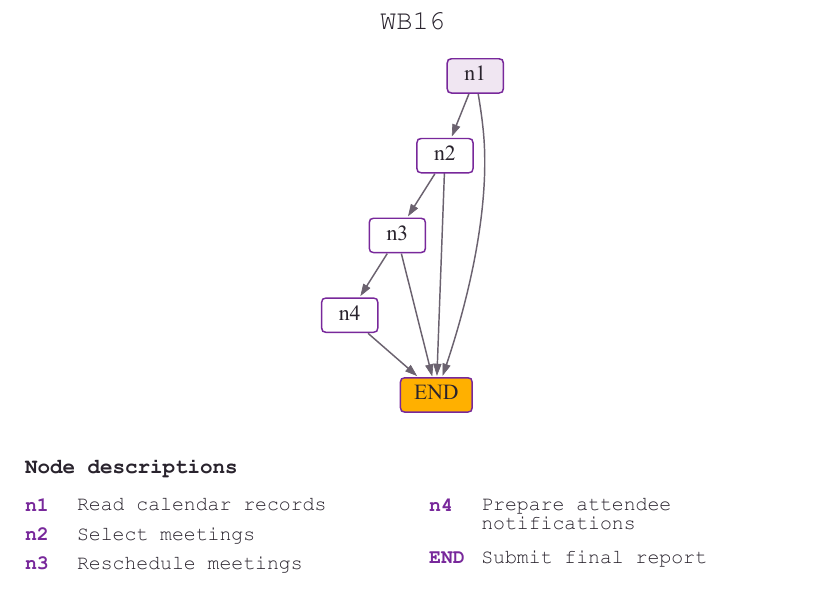}
\caption{DAG for \texttt{WB16}.}
\label{fig:dag-wb_calendar_shift}

\end{figure}

\clearpage

\begin{figure}[p]\centering

\includegraphics[width=.95\linewidth]{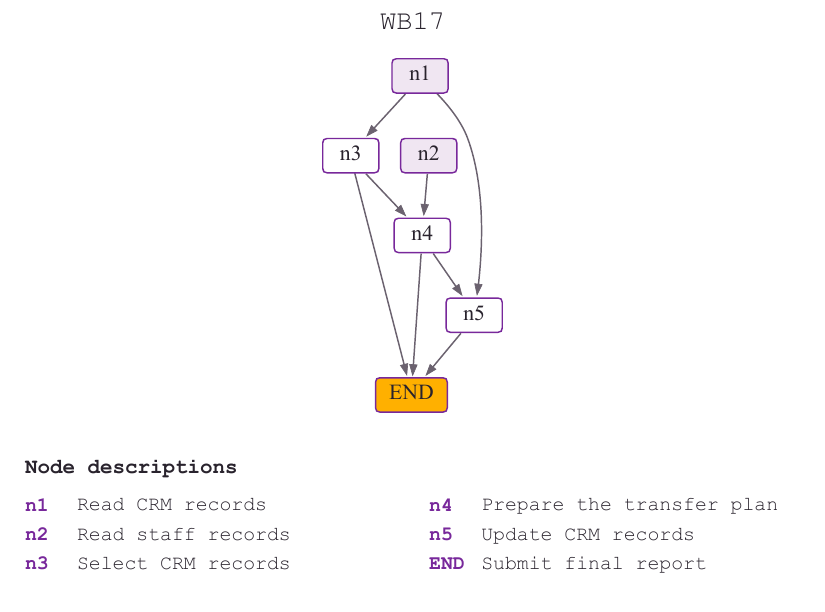}
\caption{DAG for \texttt{WB17}.}
\label{fig:dag-wb_crm_transfer}

\par\vspace{8pt}

\includegraphics[width=.95\linewidth]{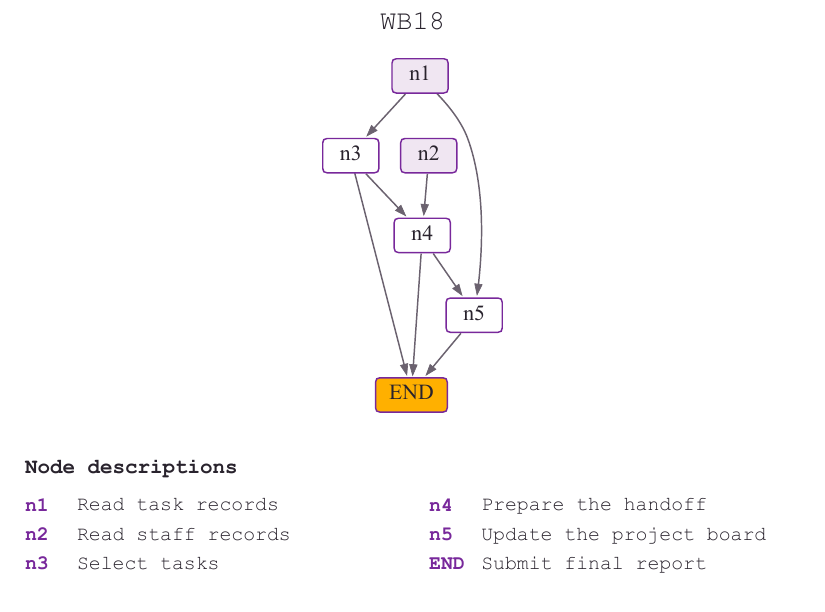}
\caption{DAG for \texttt{WB18}.}
\label{fig:dag-wb_project_handoff}

\end{figure}

\clearpage

\begin{figure}[!ht]\centering

\includegraphics[width=\linewidth,height=.90\textheight,keepaspectratio]{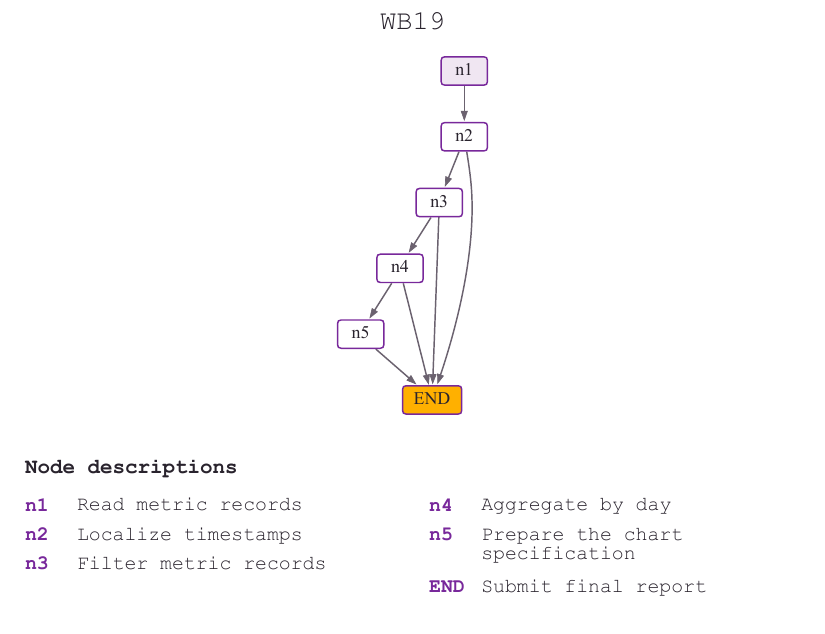}
\caption{DAG for \texttt{WB19}.}
\label{fig:dag-wb_analytics_plot}

\end{figure}

\clearpage

\begin{figure}[!ht]\centering

\includegraphics[width=\linewidth,height=.90\textheight,keepaspectratio]{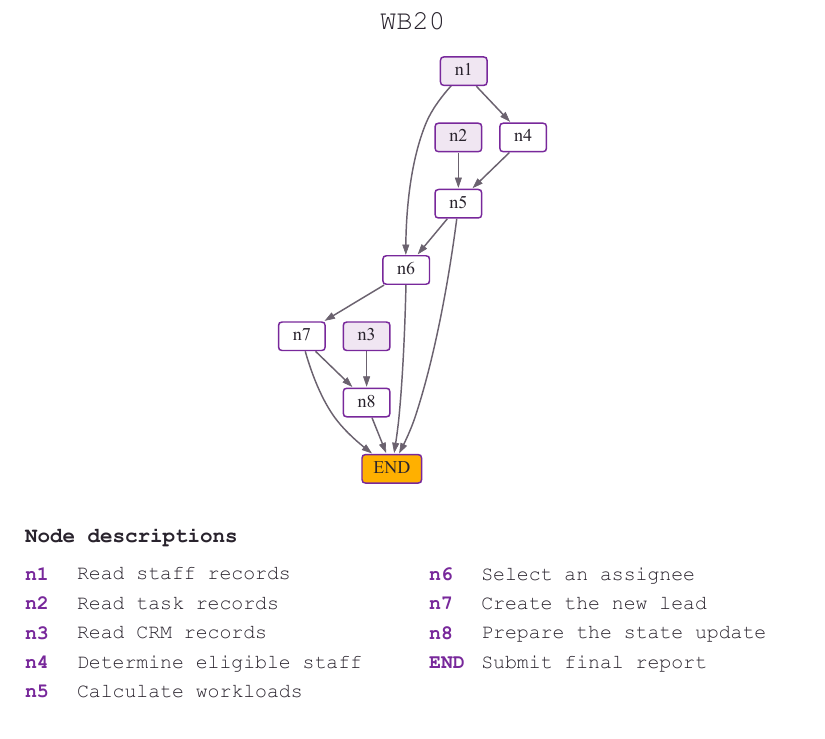}
\caption{DAG for \texttt{WB20}.}
\label{fig:dag-wb_capacity_lead}

\end{figure}

\clearpage

\begin{figure}[!ht]\centering

\includegraphics[width=\linewidth,height=.90\textheight,keepaspectratio]{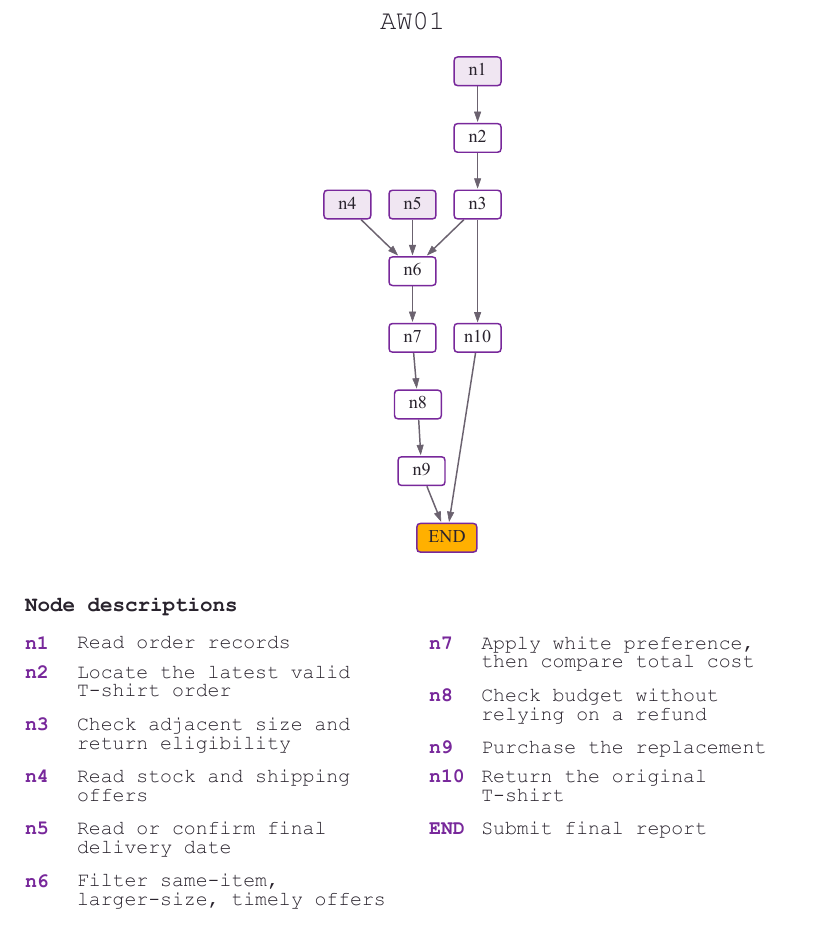}
\caption{DAG for \texttt{AW01}.}
\label{fig:dag-AW01}

\end{figure}

\clearpage

\begin{figure}[!ht]\centering

\includegraphics[width=\linewidth,height=.90\textheight,keepaspectratio]{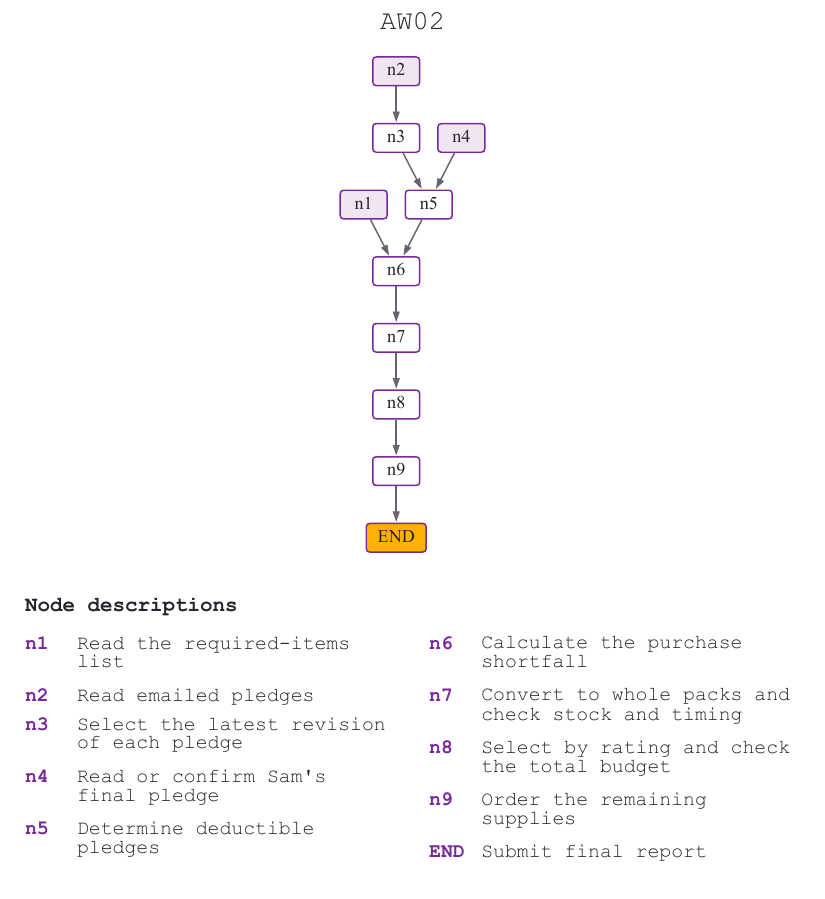}
\caption{DAG for \texttt{AW02}.}
\label{fig:dag-AW02}

\end{figure}

\clearpage

\begin{figure}[!ht]\centering

\includegraphics[width=\linewidth,height=.90\textheight,keepaspectratio]{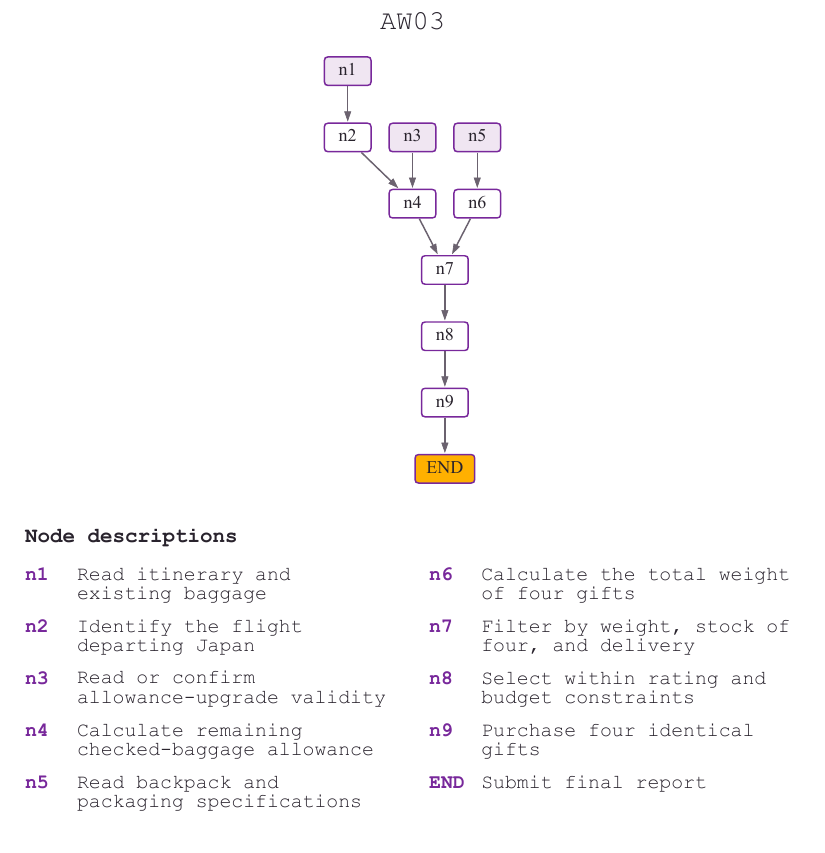}
\caption{DAG for \texttt{AW03}.}
\label{fig:dag-AW03}

\end{figure}

\clearpage

\begin{figure}[!ht]\centering

\includegraphics[width=\linewidth,height=.90\textheight,keepaspectratio]{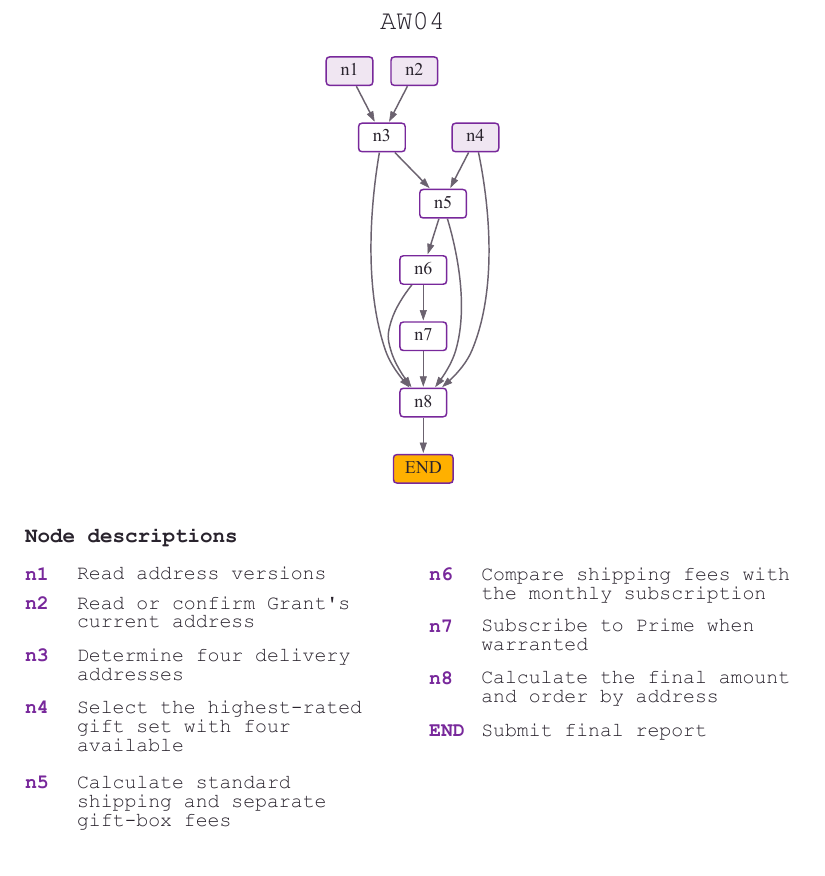}
\caption{DAG for \texttt{AW04}.}
\label{fig:dag-AW04}

\end{figure}

\clearpage

\begin{figure}[!ht]\centering

\includegraphics[width=\linewidth,height=.90\textheight,keepaspectratio]{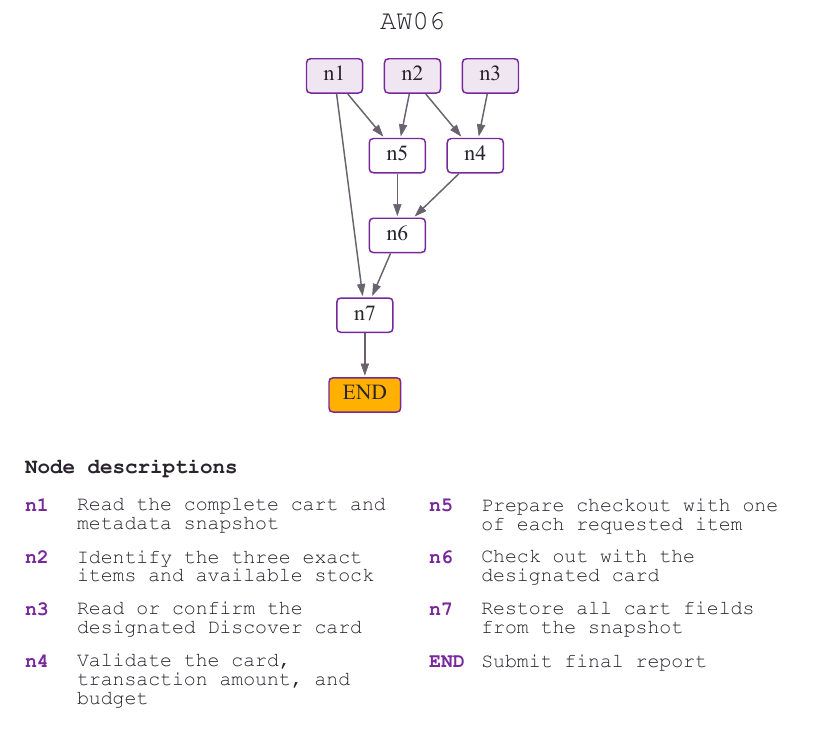}
\caption{DAG for \texttt{AW06}.}
\label{fig:dag-AW06}

\end{figure}

\clearpage

\begin{figure}[!ht]\centering

\includegraphics[width=\linewidth,height=.90\textheight,keepaspectratio]{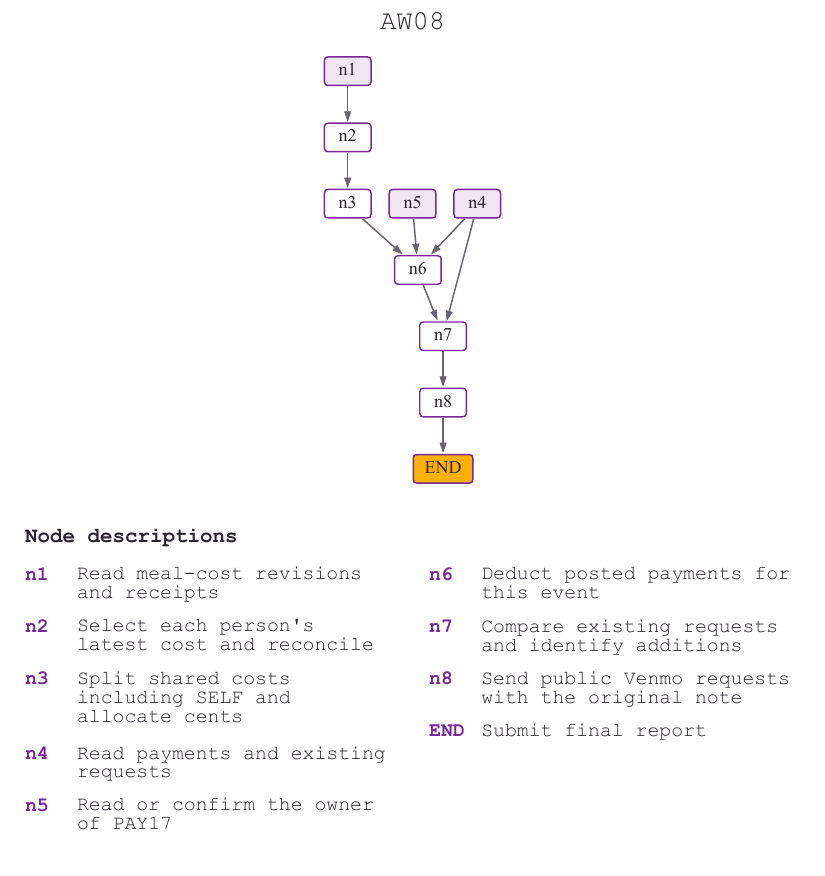}
\caption{DAG for \texttt{AW08}.}
\label{fig:dag-AW08}

\end{figure}

\clearpage

\begin{figure}[!ht]\centering

\includegraphics[width=\linewidth,height=.90\textheight,keepaspectratio]{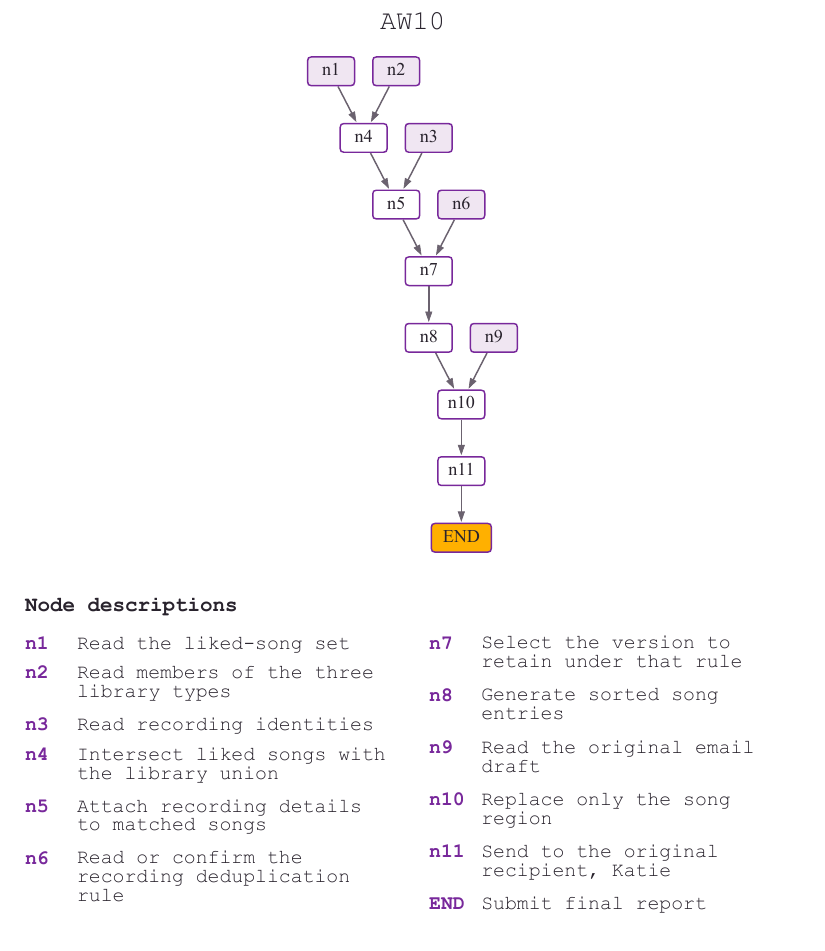}
\caption{DAG for \texttt{AW10}.}
\label{fig:dag-AW10}

\end{figure}

\clearpage

\begin{figure}[!ht]\centering

\includegraphics[width=\linewidth,height=.90\textheight,keepaspectratio]{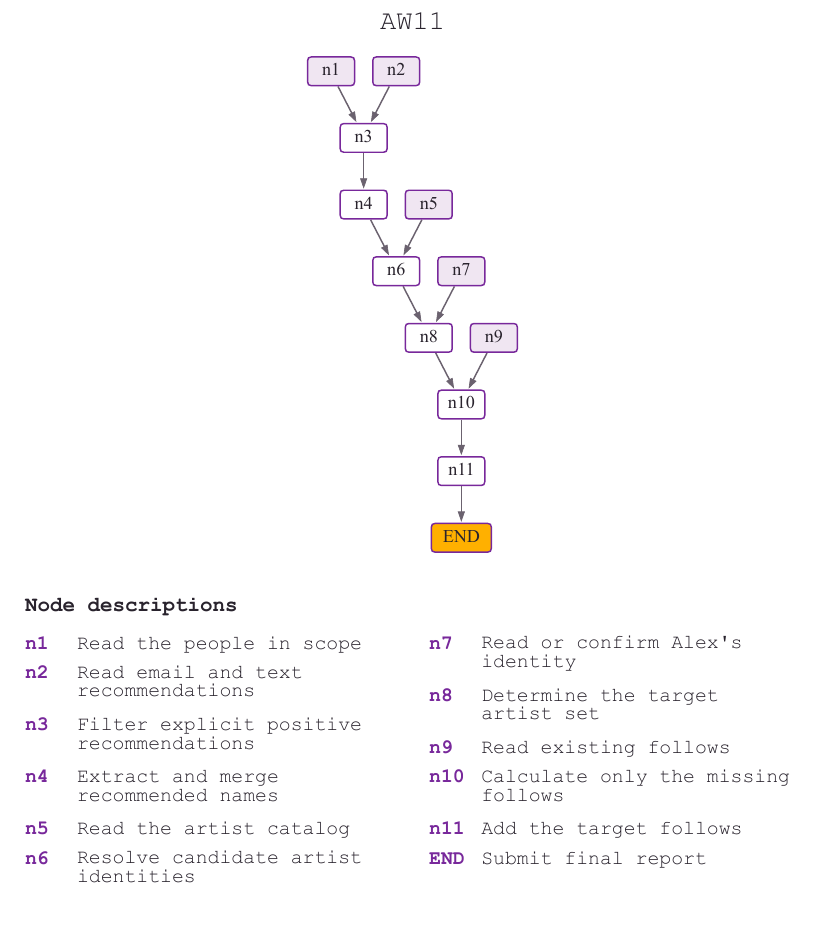}
\caption{DAG for \texttt{AW11}.}
\label{fig:dag-AW11}

\end{figure}

\clearpage

\begin{figure}[!ht]\centering

\includegraphics[width=\linewidth,height=.90\textheight,keepaspectratio]{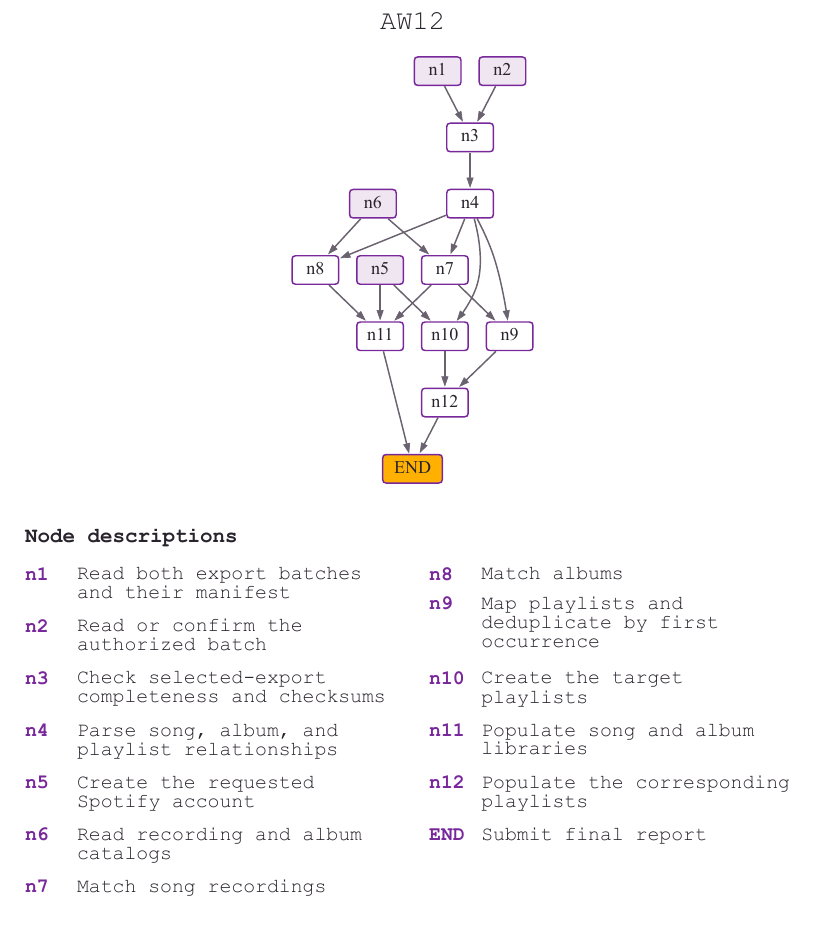}
\caption{DAG for \texttt{AW12}.}
\label{fig:dag-AW12}

\end{figure}

\clearpage

\begin{figure}[!ht]\centering

\includegraphics[width=\linewidth,height=.90\textheight,keepaspectratio]{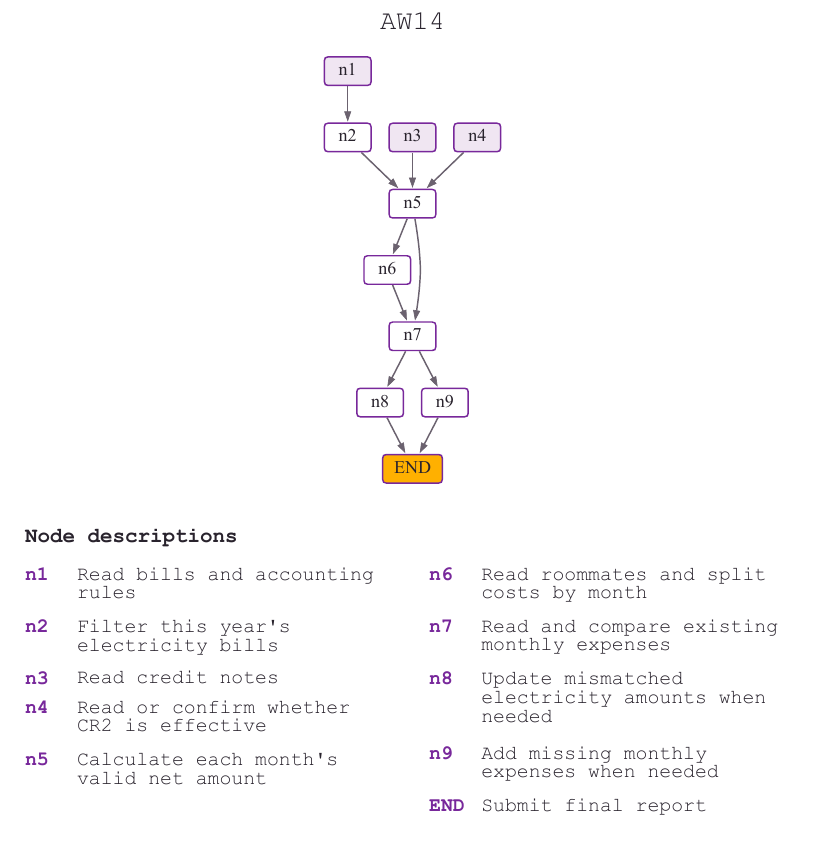}
\caption{DAG for \texttt{AW14}.}
\label{fig:dag-AW14}

\end{figure}

\clearpage

\begin{figure}[!ht]\centering

\includegraphics[width=\linewidth,height=.90\textheight,keepaspectratio]{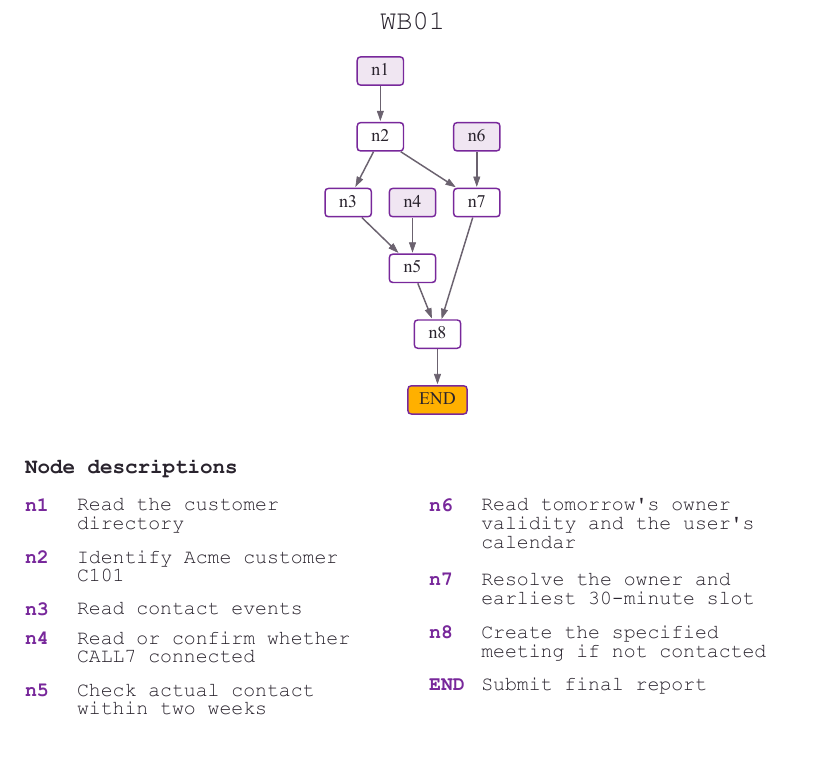}
\caption{DAG for \texttt{WB01}.}
\label{fig:dag-WB01}

\end{figure}

\clearpage

\begin{figure}[!ht]\centering

\includegraphics[width=\linewidth,height=.90\textheight,keepaspectratio]{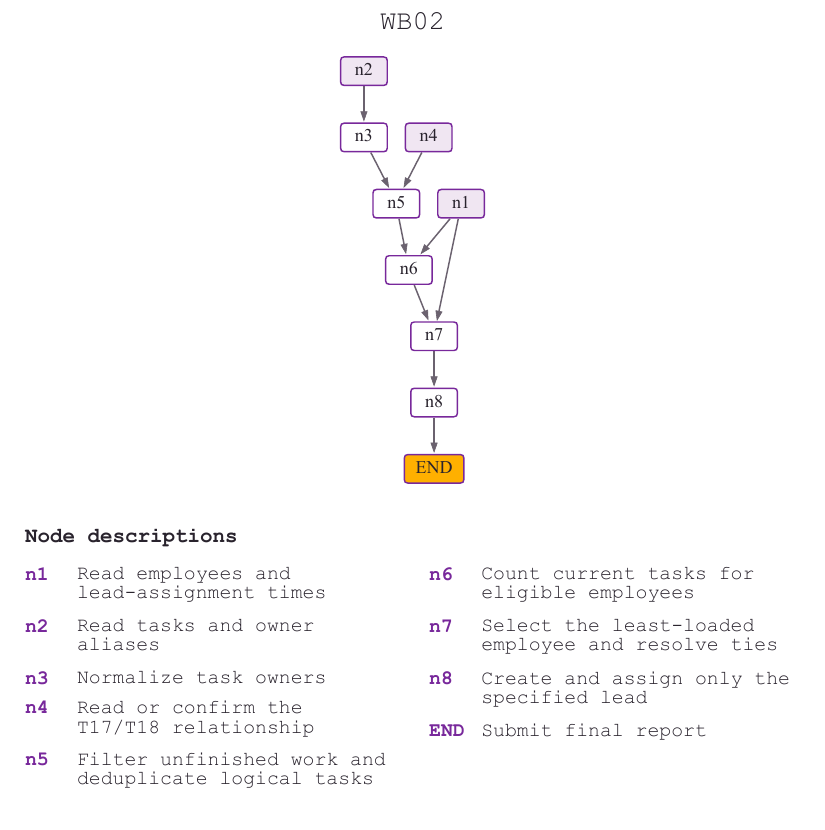}
\caption{DAG for \texttt{WB02}.}
\label{fig:dag-WB02}

\end{figure}

\clearpage

\begin{figure}[!ht]\centering

\includegraphics[width=\linewidth,height=.90\textheight,keepaspectratio]{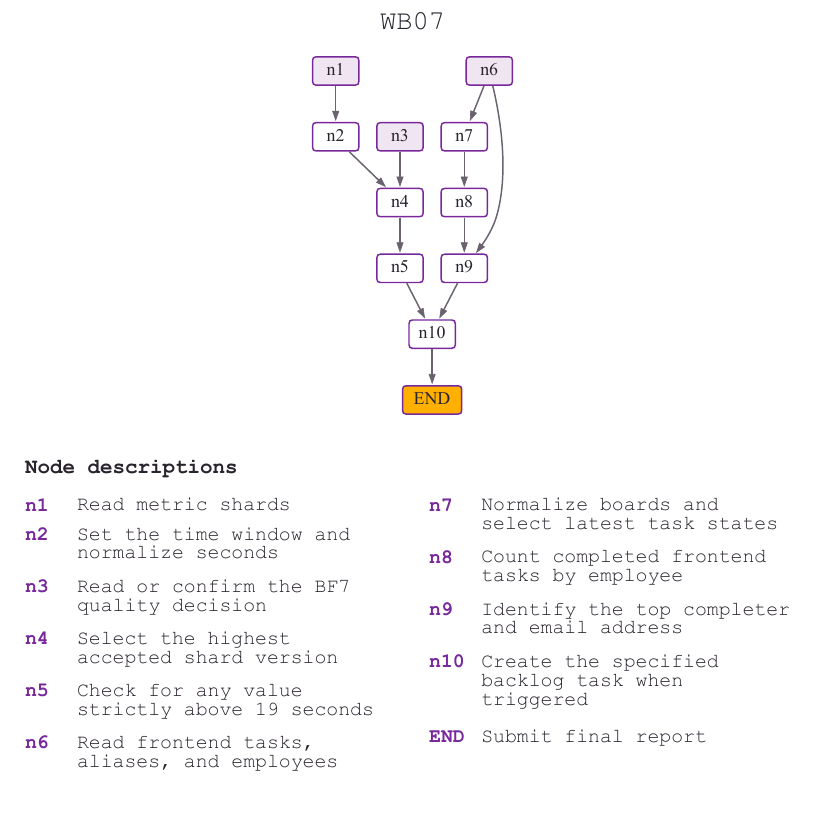}
\caption{DAG for \texttt{WB07}.}
\label{fig:dag-WB07}

\end{figure}

\clearpage

\begin{figure}[!ht]\centering

\includegraphics[width=\linewidth,height=.90\textheight,keepaspectratio]{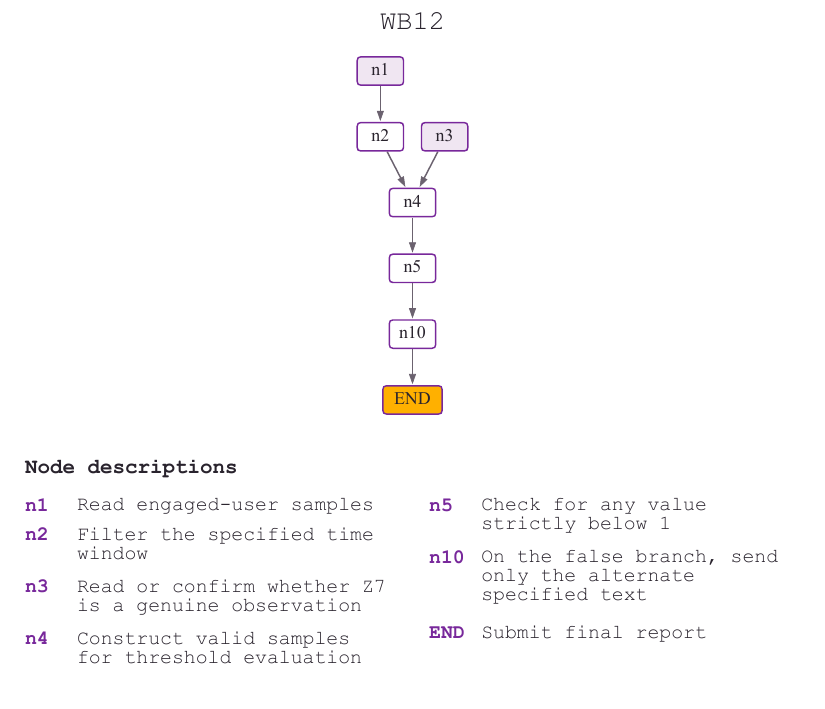}
\caption{DAG for \texttt{WB12}.}
\label{fig:dag-WB12}

\end{figure}

\clearpage

\begin{figure}[!ht]\centering

\includegraphics[width=\linewidth,height=.90\textheight,keepaspectratio]{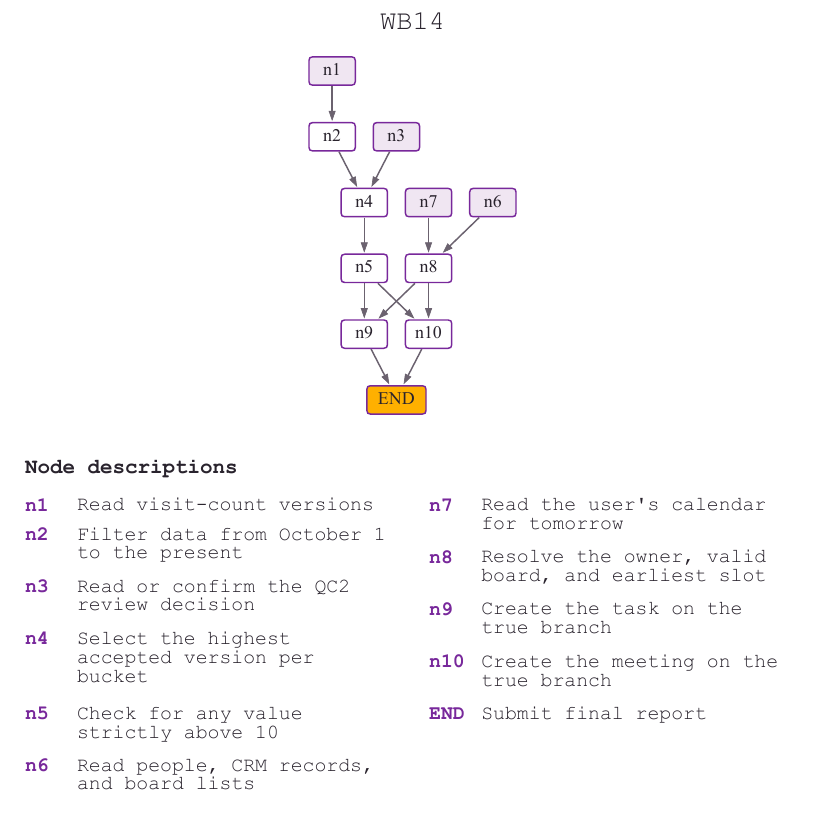}
\caption{DAG for \texttt{WB14}.}
\label{fig:dag-WB14}

\end{figure}

\clearpage

\begin{figure}[!ht]\centering

\includegraphics[width=\linewidth,height=.90\textheight,keepaspectratio]{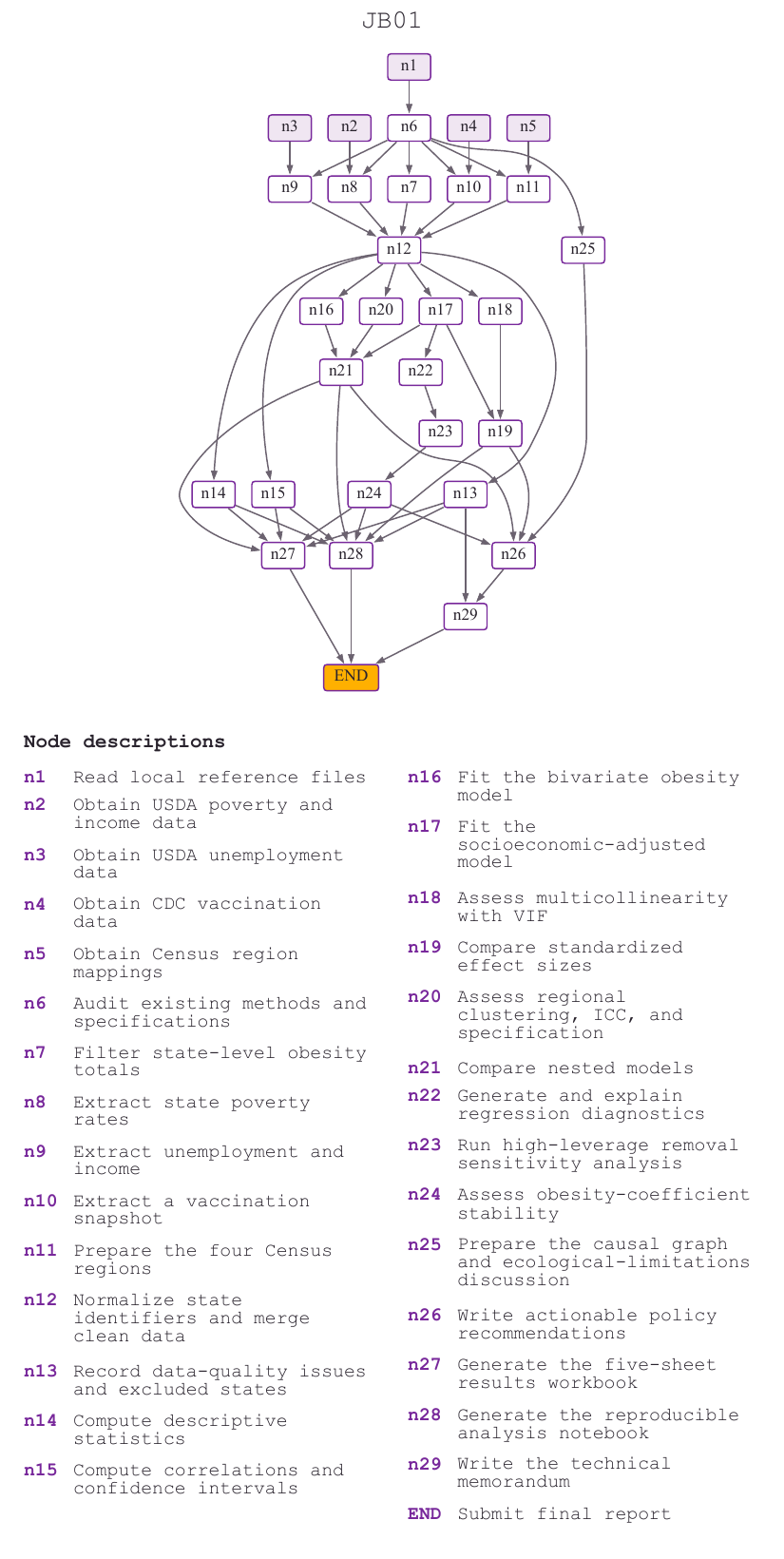}
\caption{DAG for \texttt{JB01}.}
\label{fig:dag-JB01}

\end{figure}

\clearpage

\begin{figure}[!ht]\centering

\includegraphics[width=\linewidth,height=.90\textheight,keepaspectratio]{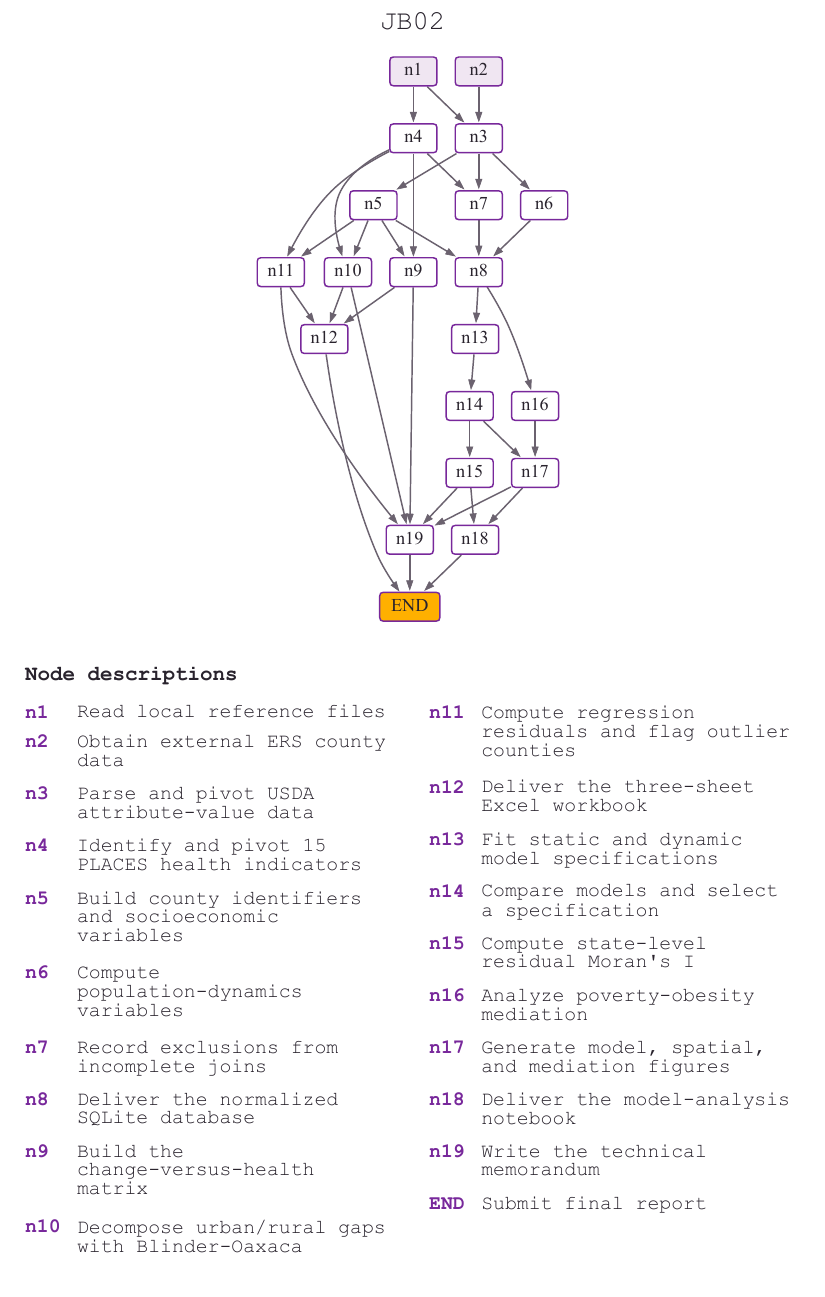}
\caption{DAG for \texttt{JB02}.}
\label{fig:dag-JB02}

\end{figure}

\clearpage

\begin{figure}[!ht]\centering

\includegraphics[width=\linewidth,height=.90\textheight,keepaspectratio]{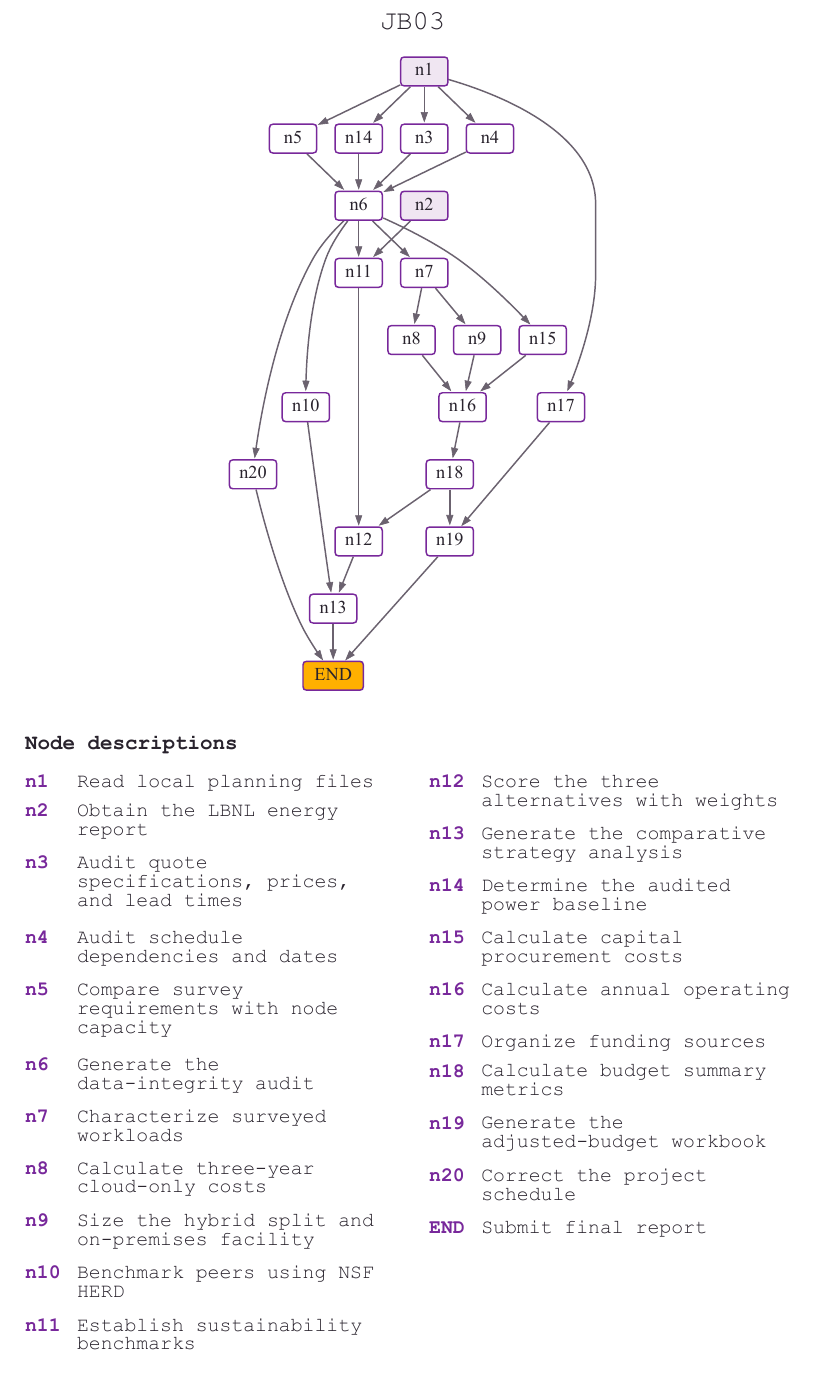}
\caption{DAG for \texttt{JB03}.}
\label{fig:dag-JB03}

\end{figure}

\clearpage

\begin{figure}[!ht]\centering

\includegraphics[width=\linewidth,height=.90\textheight,keepaspectratio]{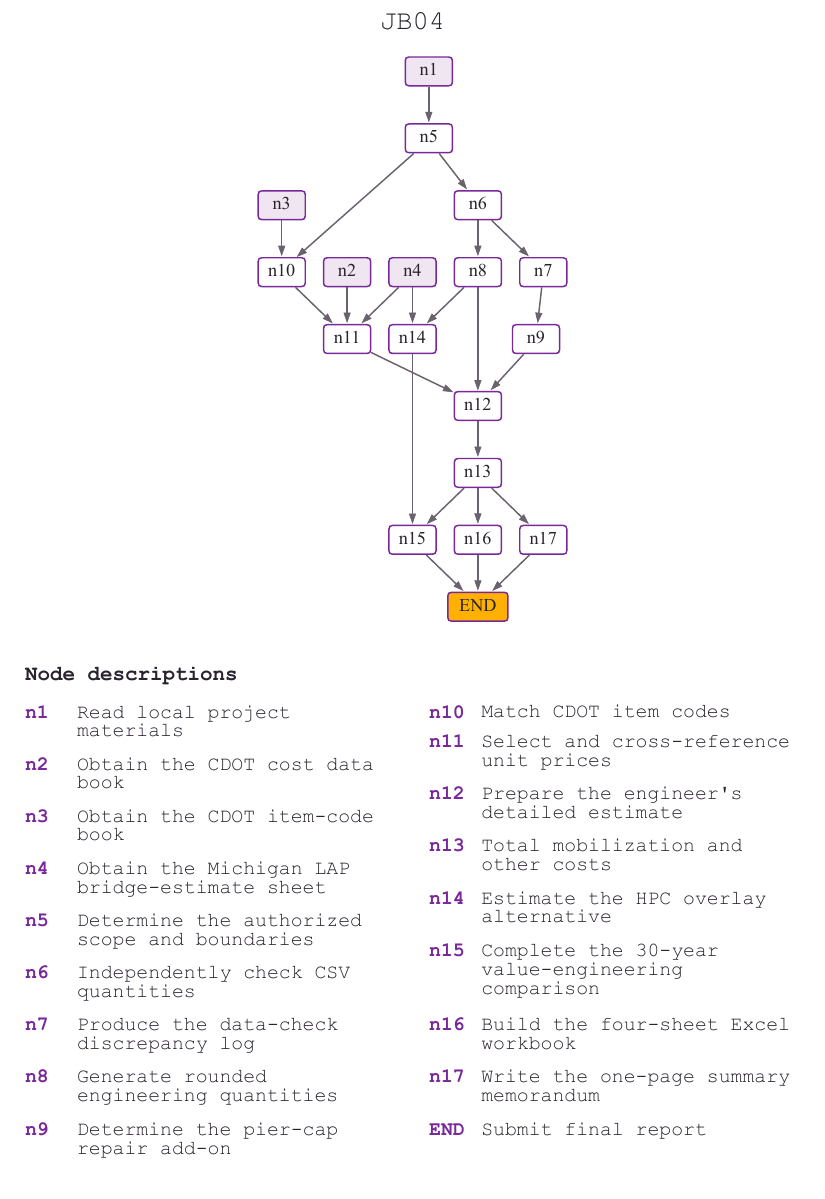}
\caption{DAG for \texttt{JB04}.}
\label{fig:dag-JB04}

\end{figure}

\clearpage

\begin{figure}[!ht]\centering

\includegraphics[width=\linewidth,height=.90\textheight,keepaspectratio]{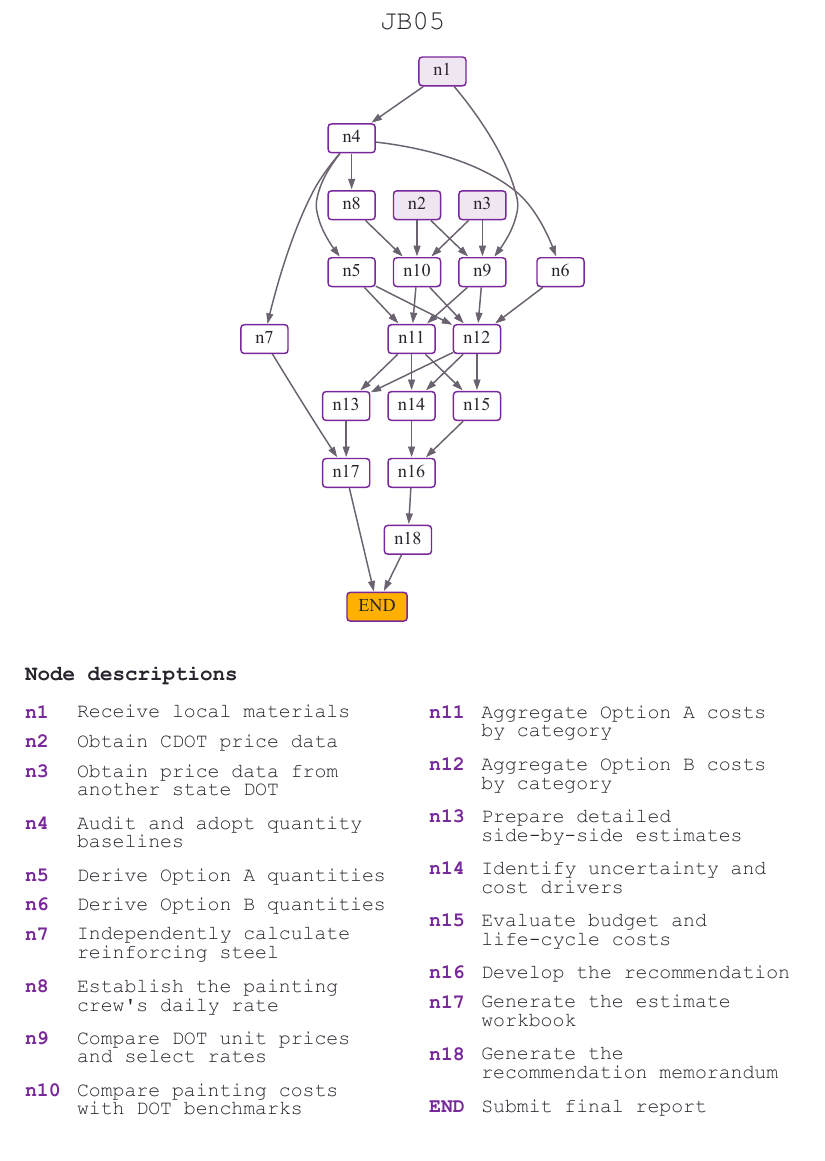}
\caption{DAG for \texttt{JB05}.}
\label{fig:dag-JB05}

\end{figure}

\clearpage

\begin{figure}[!ht]\centering

\includegraphics[width=\linewidth,height=.90\textheight,keepaspectratio]{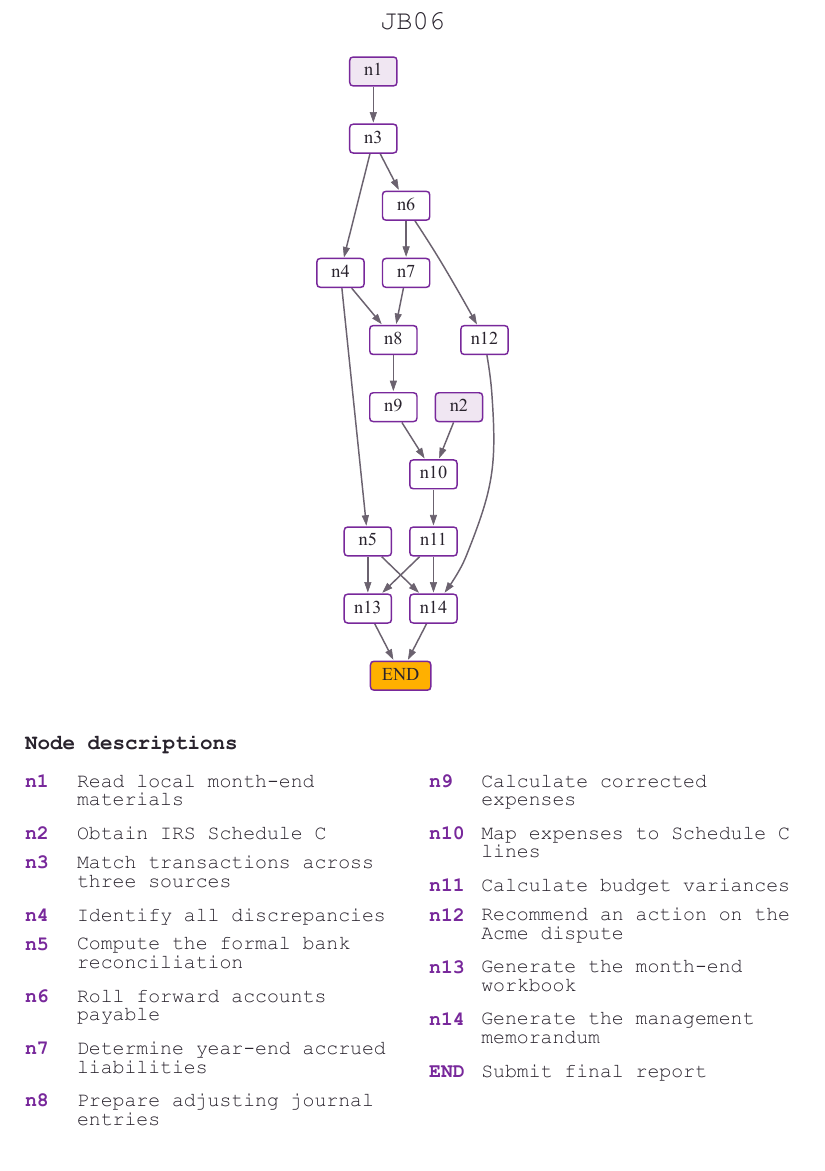}
\caption{DAG for \texttt{JB06}.}
\label{fig:dag-JB06}

\end{figure}

\clearpage

\begin{figure}[!ht]\centering

\includegraphics[width=\linewidth,height=.90\textheight,keepaspectratio]{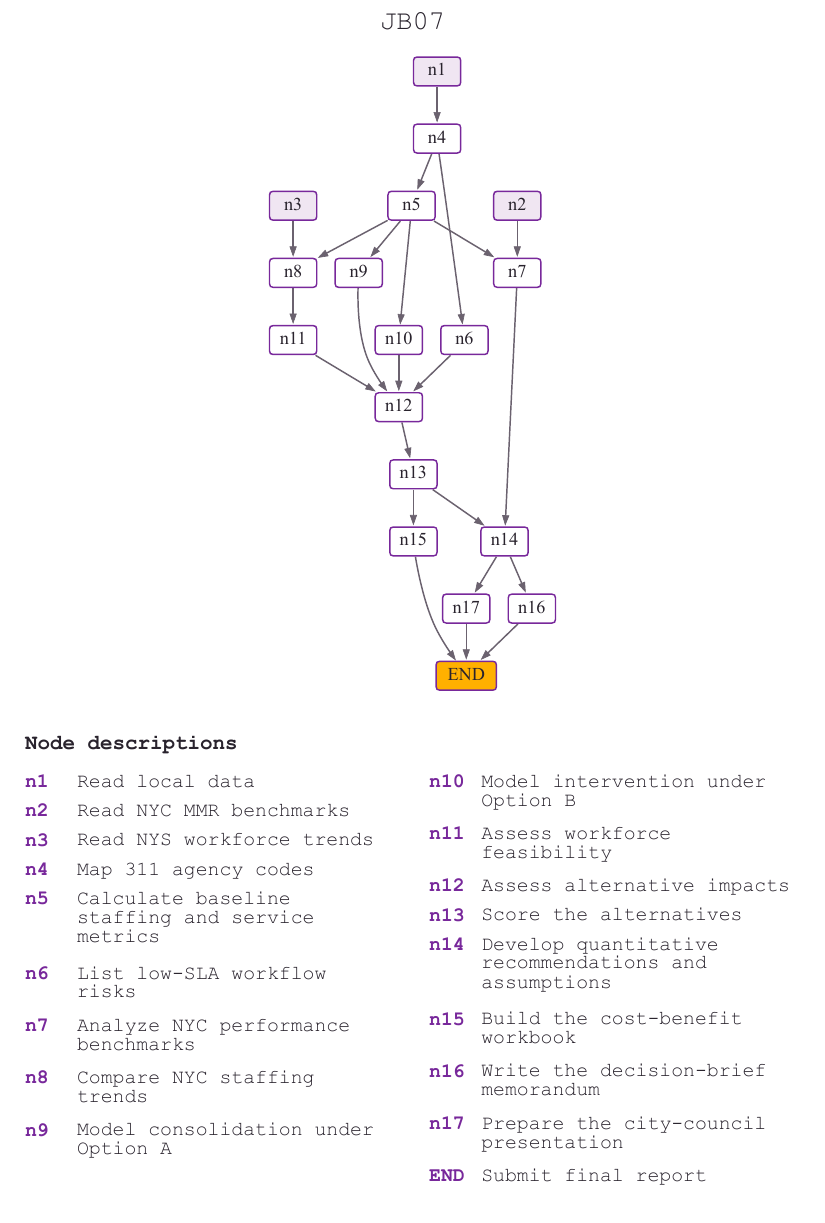}
\caption{DAG for \texttt{JB07}.}
\label{fig:dag-JB07}

\end{figure}

\clearpage

\begin{figure}[!ht]\centering

\includegraphics[width=\linewidth,height=.90\textheight,keepaspectratio]{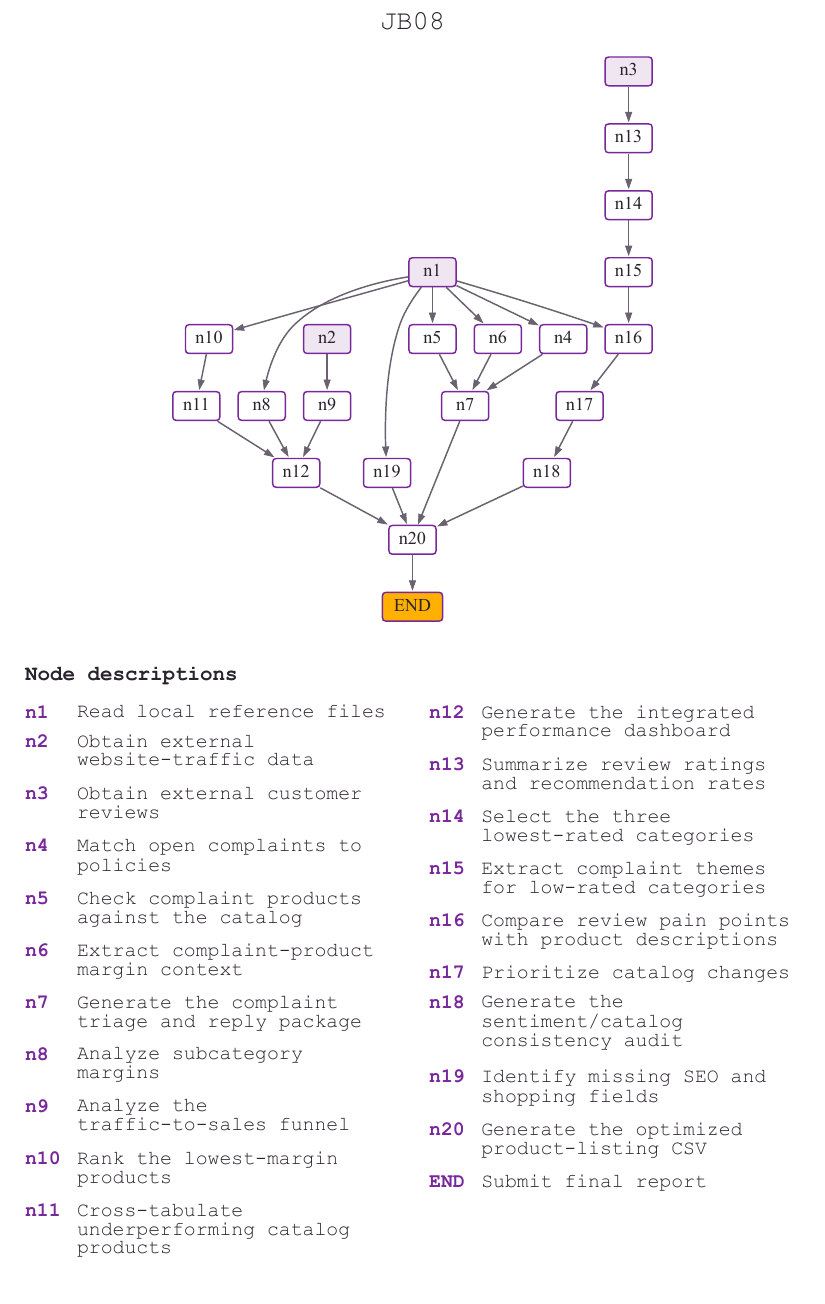}
\caption{DAG for \texttt{JB08}.}
\label{fig:dag-JB08}

\end{figure}

\clearpage

\begin{figure}[!ht]\centering

\includegraphics[width=\linewidth,height=.90\textheight,keepaspectratio]{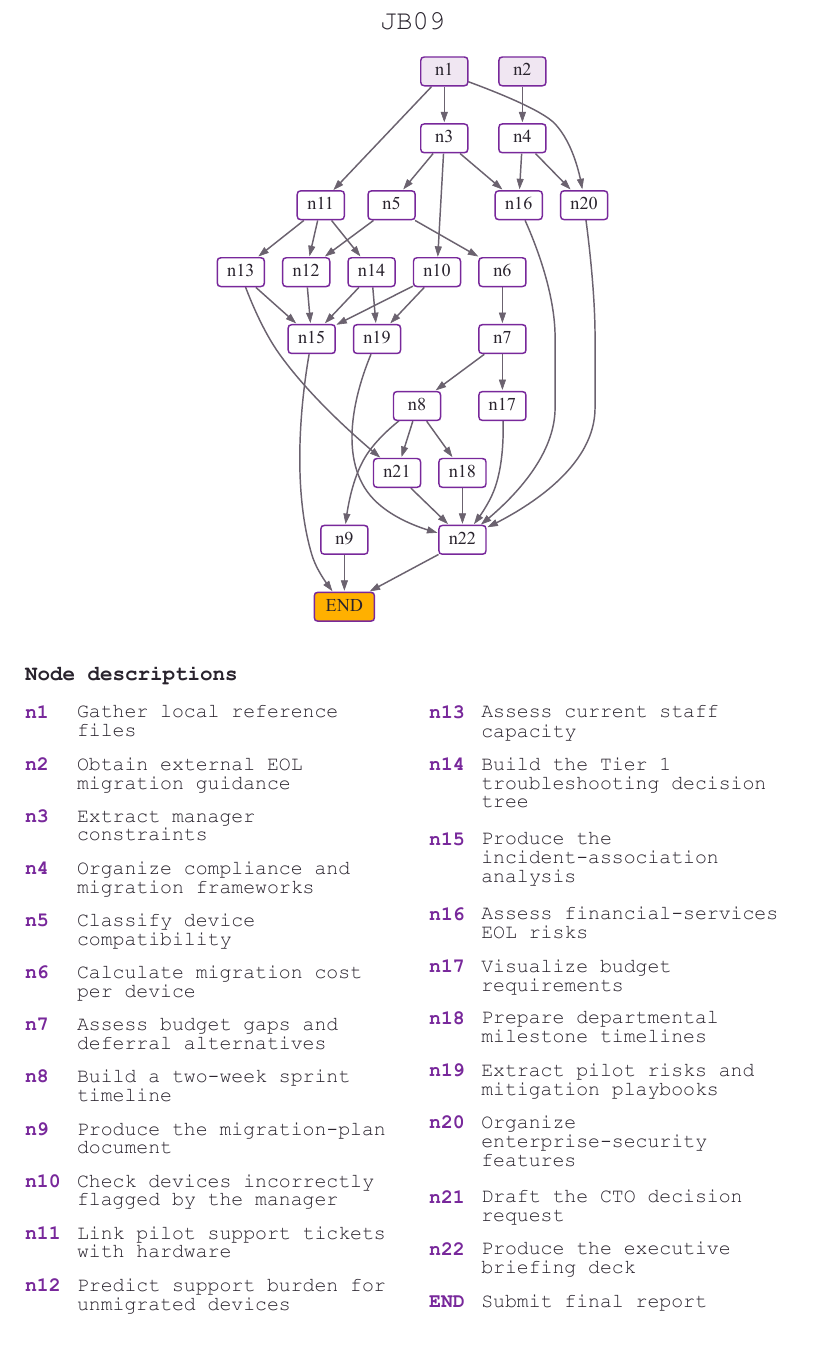}
\caption{DAG for \texttt{JB09}.}
\label{fig:dag-JB09}

\end{figure}

\clearpage

\begin{figure}[!ht]\centering

\includegraphics[width=\linewidth,height=.90\textheight,keepaspectratio]{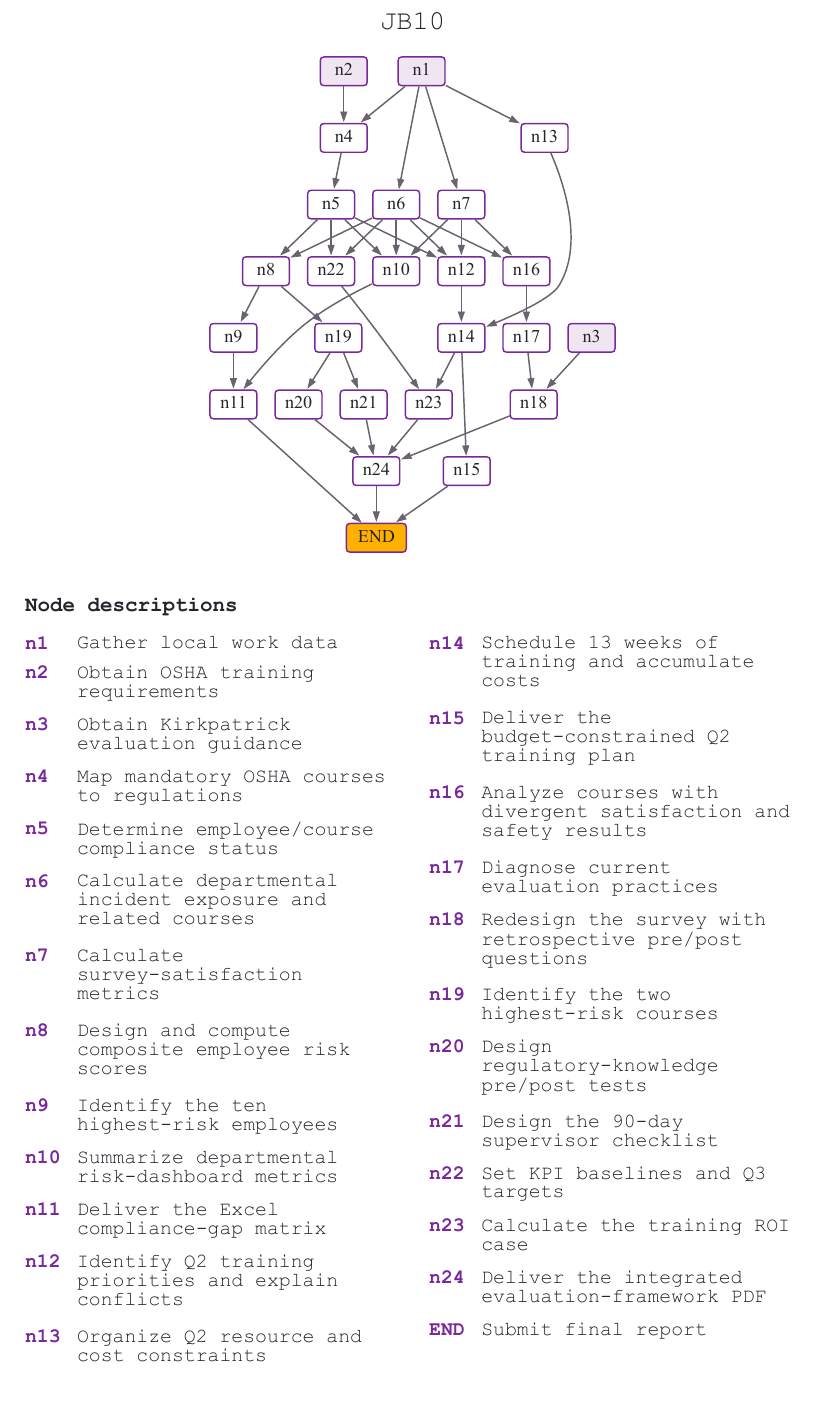}
\caption{DAG for \texttt{JB10}.}
\label{fig:dag-JB10}

\end{figure}

\clearpage

\begin{figure}[!ht]\centering

\includegraphics[width=\linewidth,height=.90\textheight,keepaspectratio]{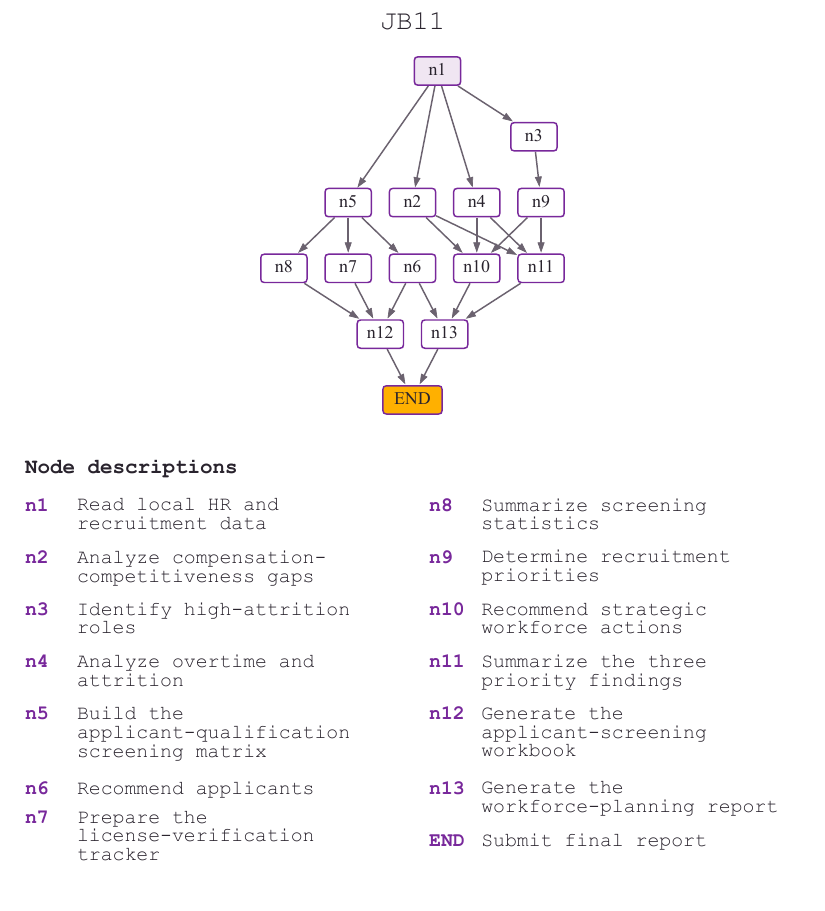}
\caption{DAG for \texttt{JB11}.}
\label{fig:dag-JB11}

\end{figure}

\clearpage

\begin{figure}[!ht]\centering

\includegraphics[width=\linewidth,height=.90\textheight,keepaspectratio]{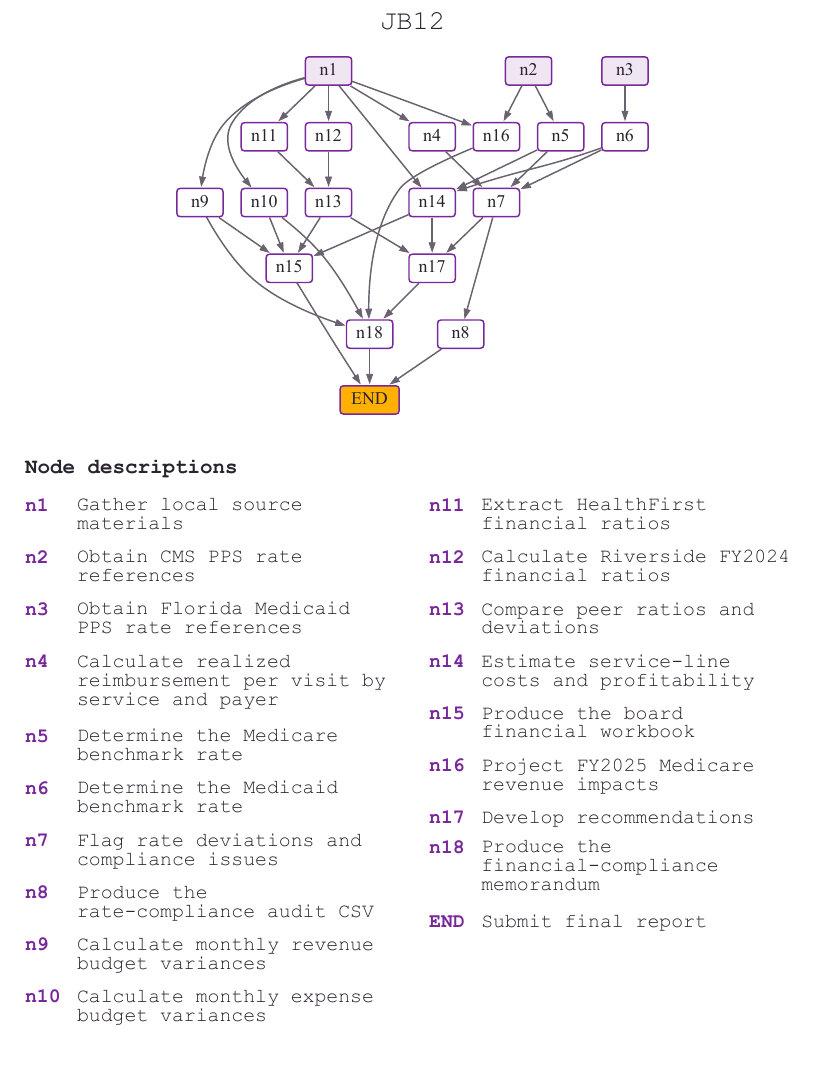}
\caption{DAG for \texttt{JB12}.}
\label{fig:dag-JB12}

\end{figure}

\clearpage

\begin{figure}[!ht]\centering

\includegraphics[width=\linewidth,height=.90\textheight,keepaspectratio]{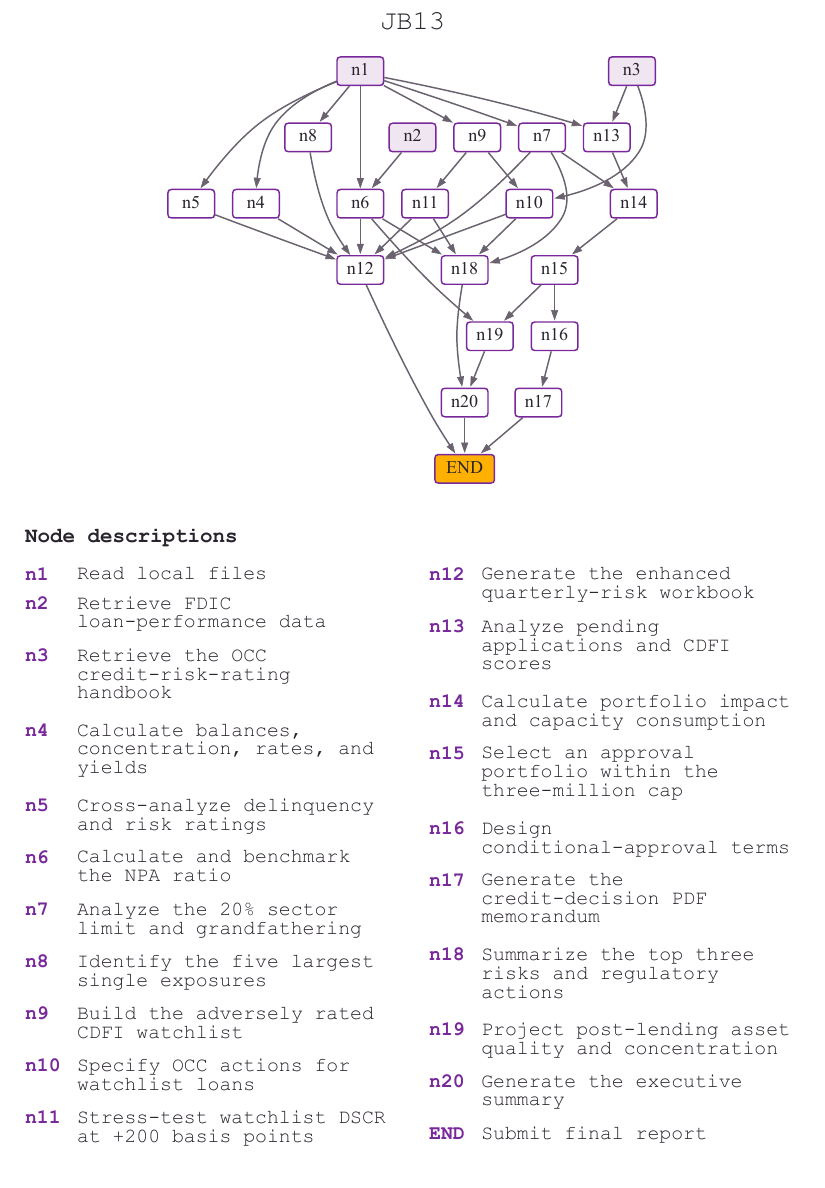}
\caption{DAG for \texttt{JB13}.}
\label{fig:dag-JB13}

\end{figure}

\clearpage

\begin{figure}[!ht]\centering

\includegraphics[width=\linewidth,height=.90\textheight,keepaspectratio]{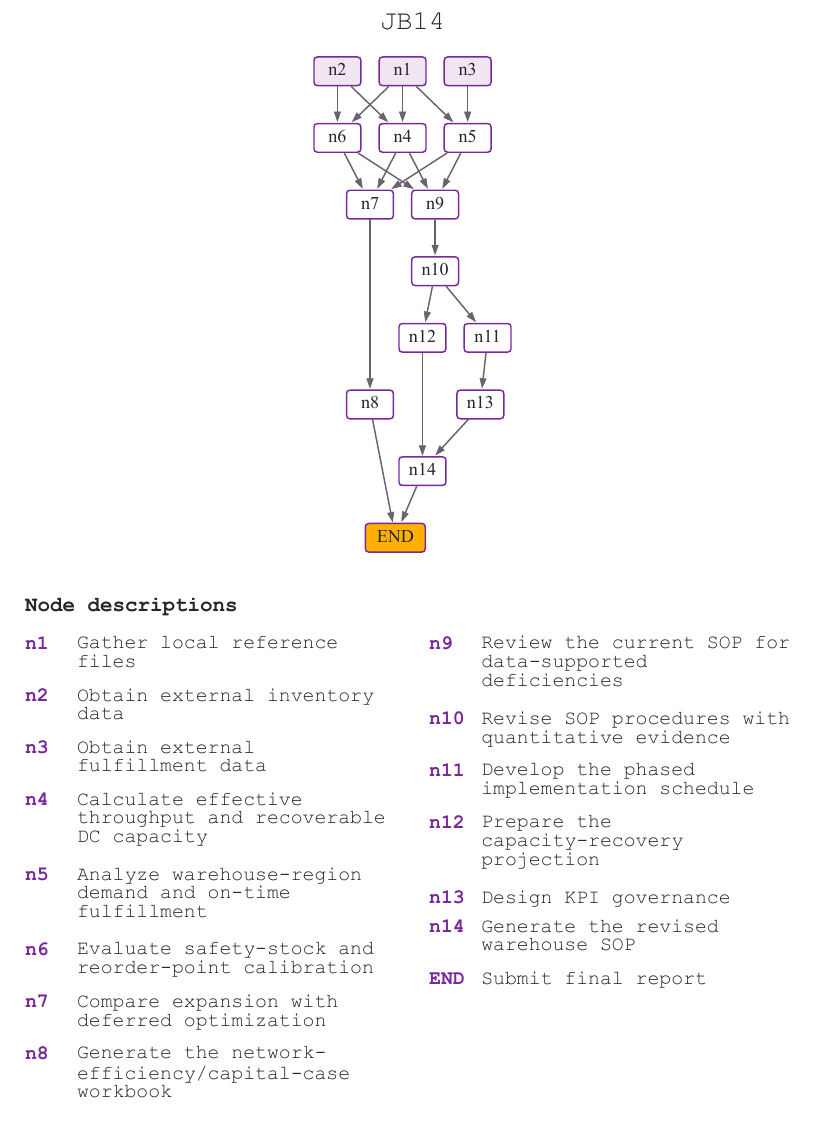}
\caption{DAG for \texttt{JB14}.}
\label{fig:dag-JB14}

\end{figure}

\clearpage

\begin{figure}[!ht]\centering

\includegraphics[width=\linewidth,height=.90\textheight,keepaspectratio]{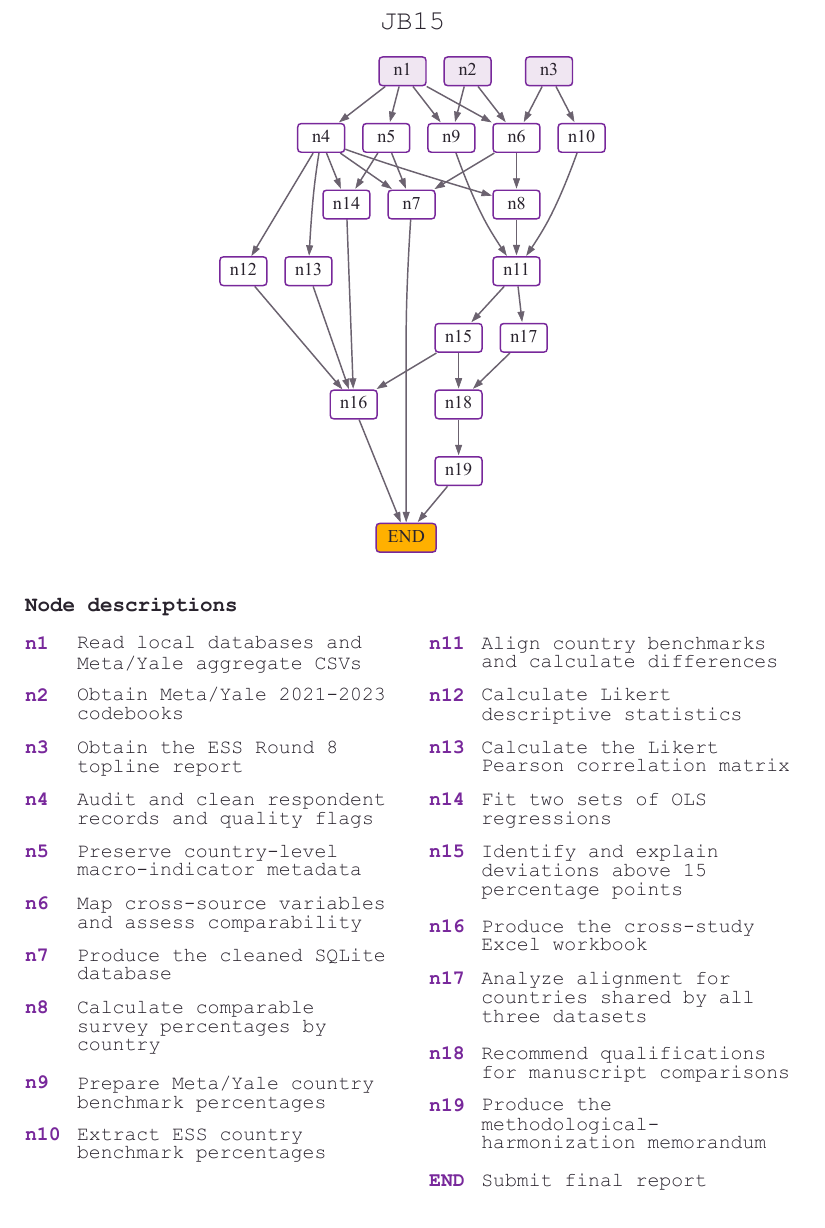}
\caption{DAG for \texttt{JB15}.}
\label{fig:dag-JB15}

\end{figure}

\clearpage

\begin{figure}[!ht]\centering

\includegraphics[width=\linewidth,height=.90\textheight,keepaspectratio]{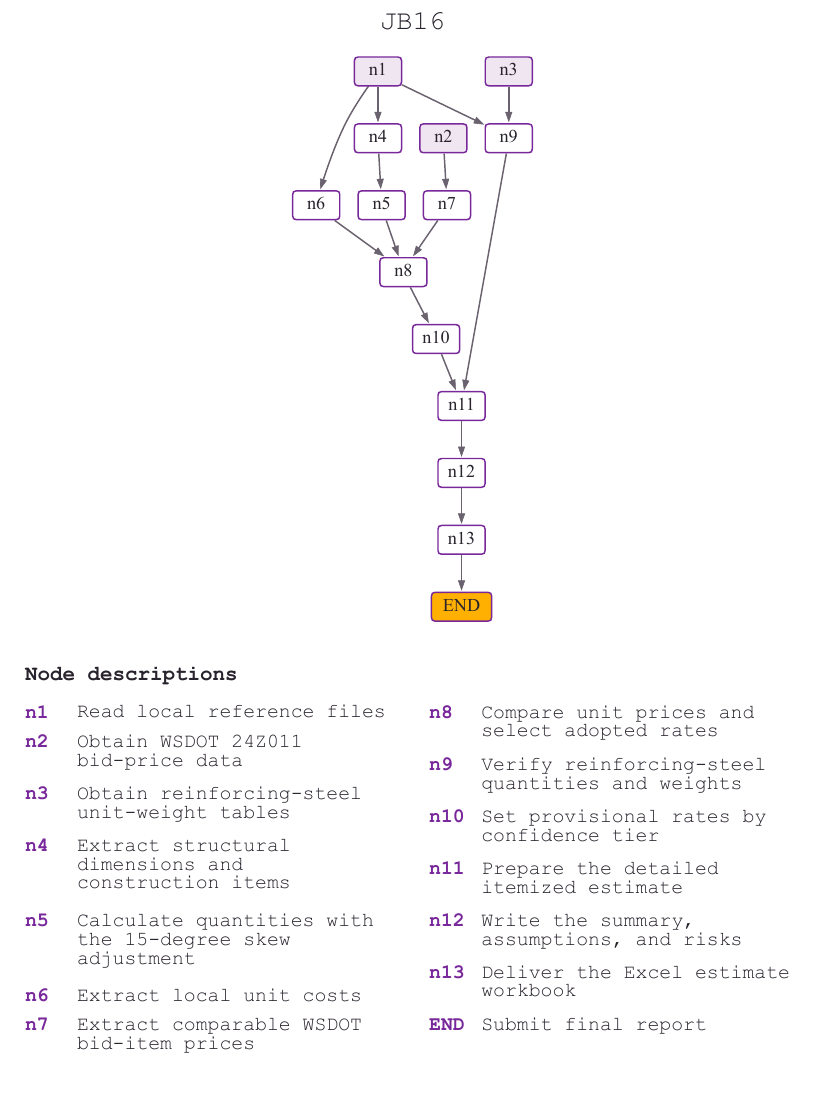}
\caption{DAG for \texttt{JB16}.}
\label{fig:dag-JB16}

\end{figure}

\clearpage

\begin{figure}[!ht]\centering

\includegraphics[width=\linewidth,height=.90\textheight,keepaspectratio]{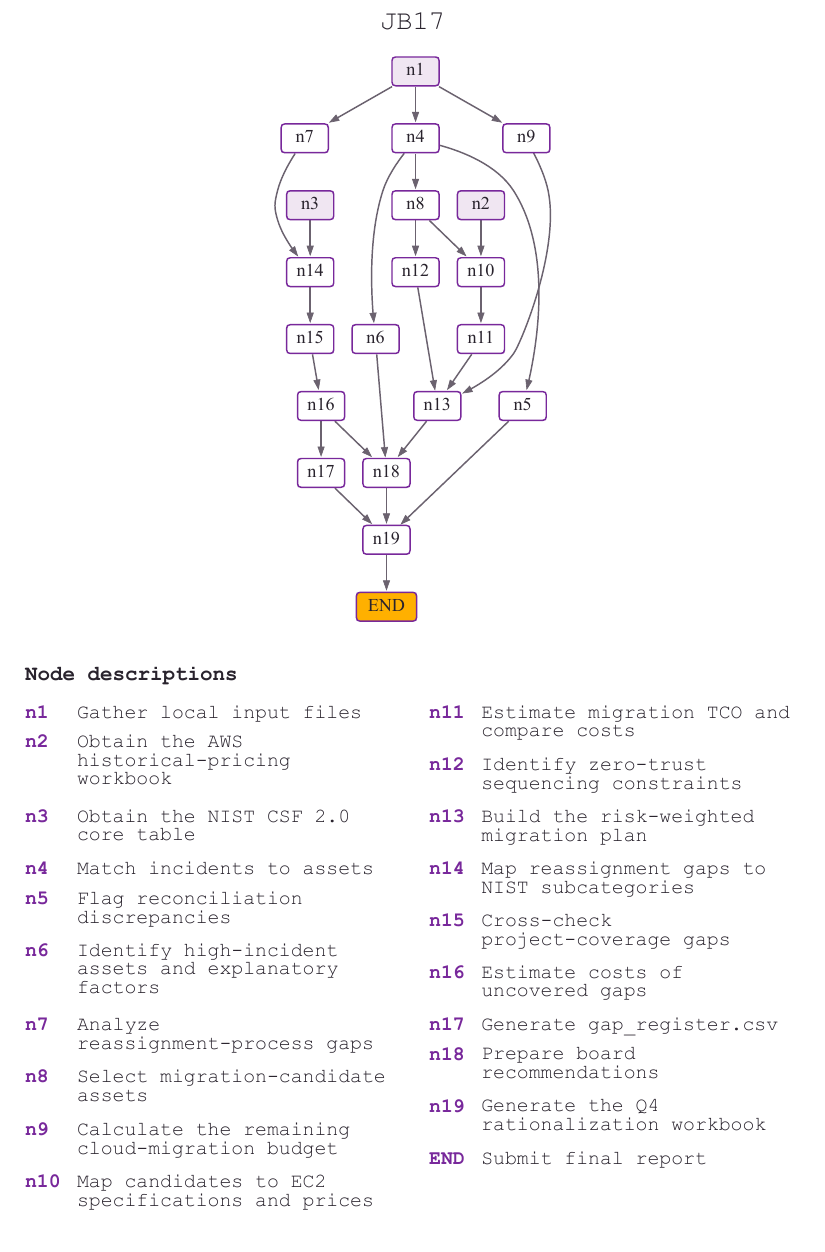}
\caption{DAG for \texttt{JB17}.}
\label{fig:dag-JB17}

\end{figure}

\clearpage

\begin{figure}[!ht]\centering

\includegraphics[width=\linewidth,height=.90\textheight,keepaspectratio]{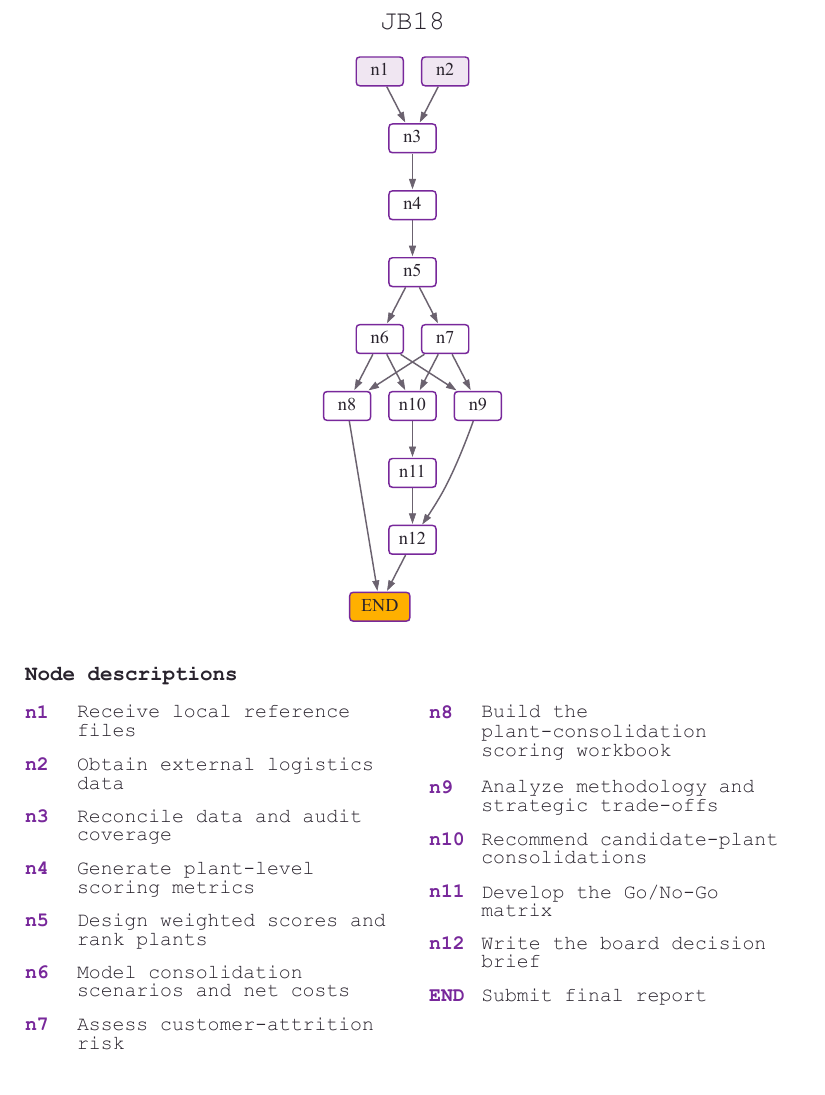}
\caption{DAG for \texttt{JB18}.}
\label{fig:dag-JB18}

\end{figure}

\clearpage

\begin{figure}[!ht]\centering

\includegraphics[width=\linewidth,height=.90\textheight,keepaspectratio]{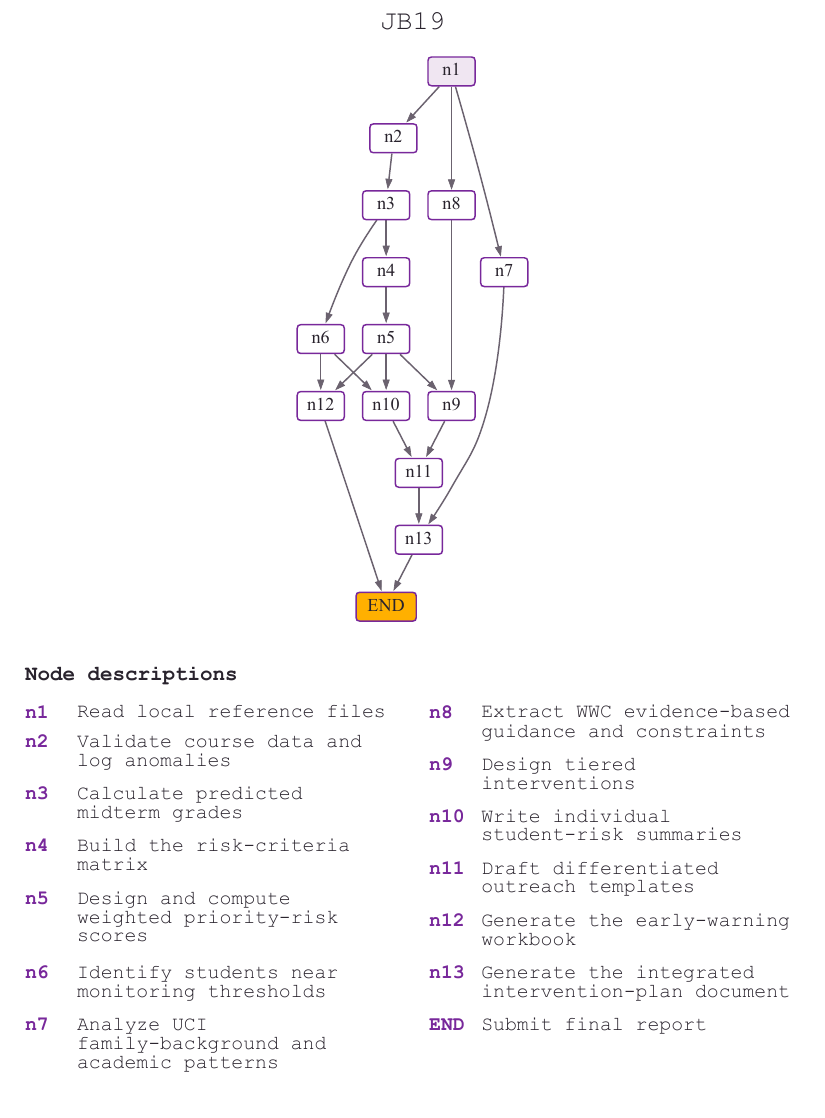}
\caption{DAG for \texttt{JB19}.}
\label{fig:dag-JB19}

\end{figure}

\clearpage

\begin{figure}[!ht]\centering

\includegraphics[width=\linewidth,height=.90\textheight,keepaspectratio]{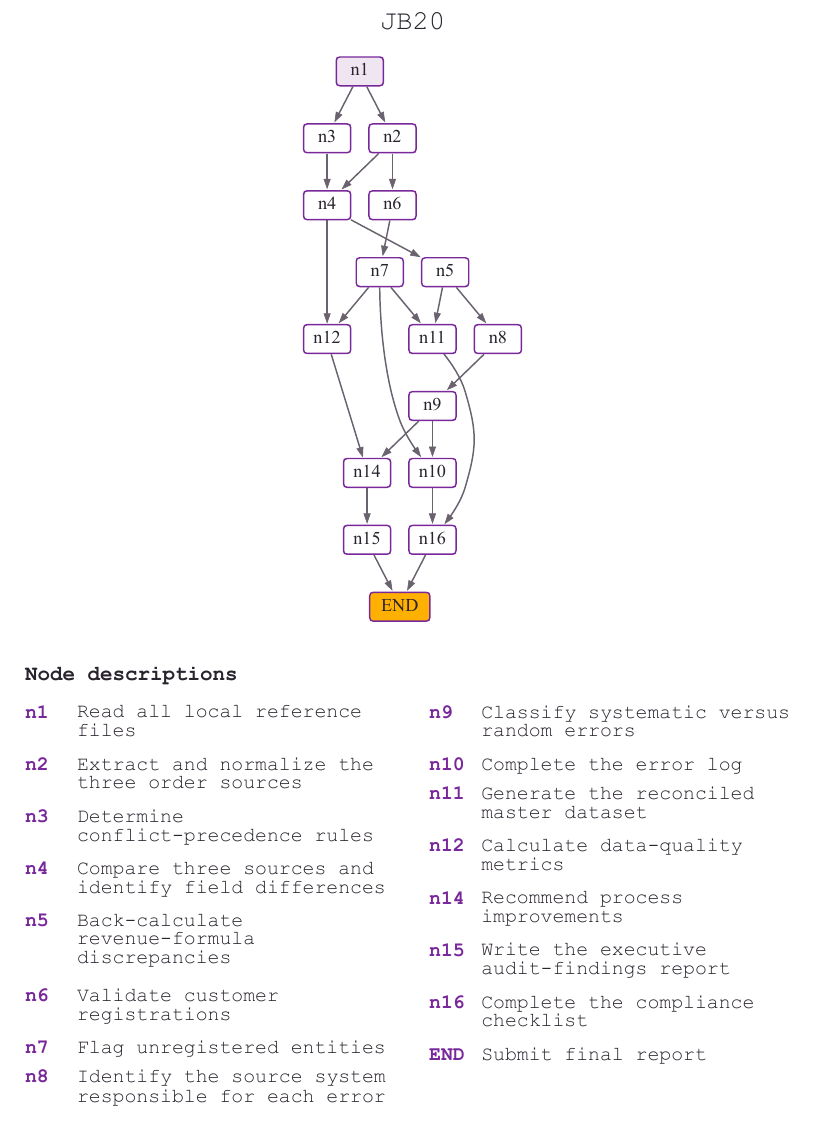}
\caption{DAG for \texttt{JB20}.}
\label{fig:dag-JB20}

\end{figure}

\clearpage

\begin{figure}[!ht]\centering

\includegraphics[width=\linewidth,height=.90\textheight,keepaspectratio]{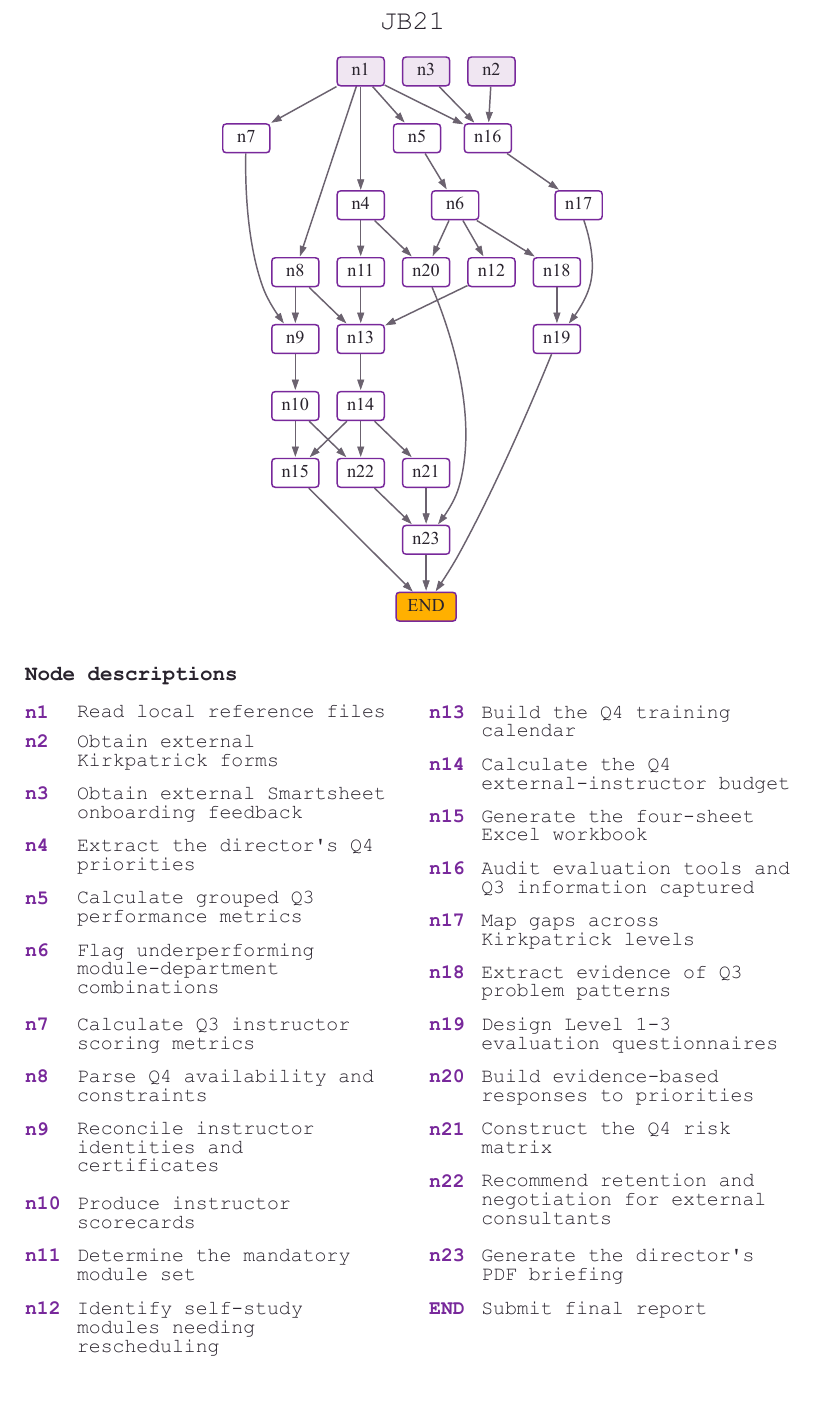}
\caption{DAG for \texttt{JB21}.}
\label{fig:dag-JB21}

\end{figure}

\clearpage

\begin{figure}[!ht]\centering

\includegraphics[width=\linewidth,height=.90\textheight,keepaspectratio]{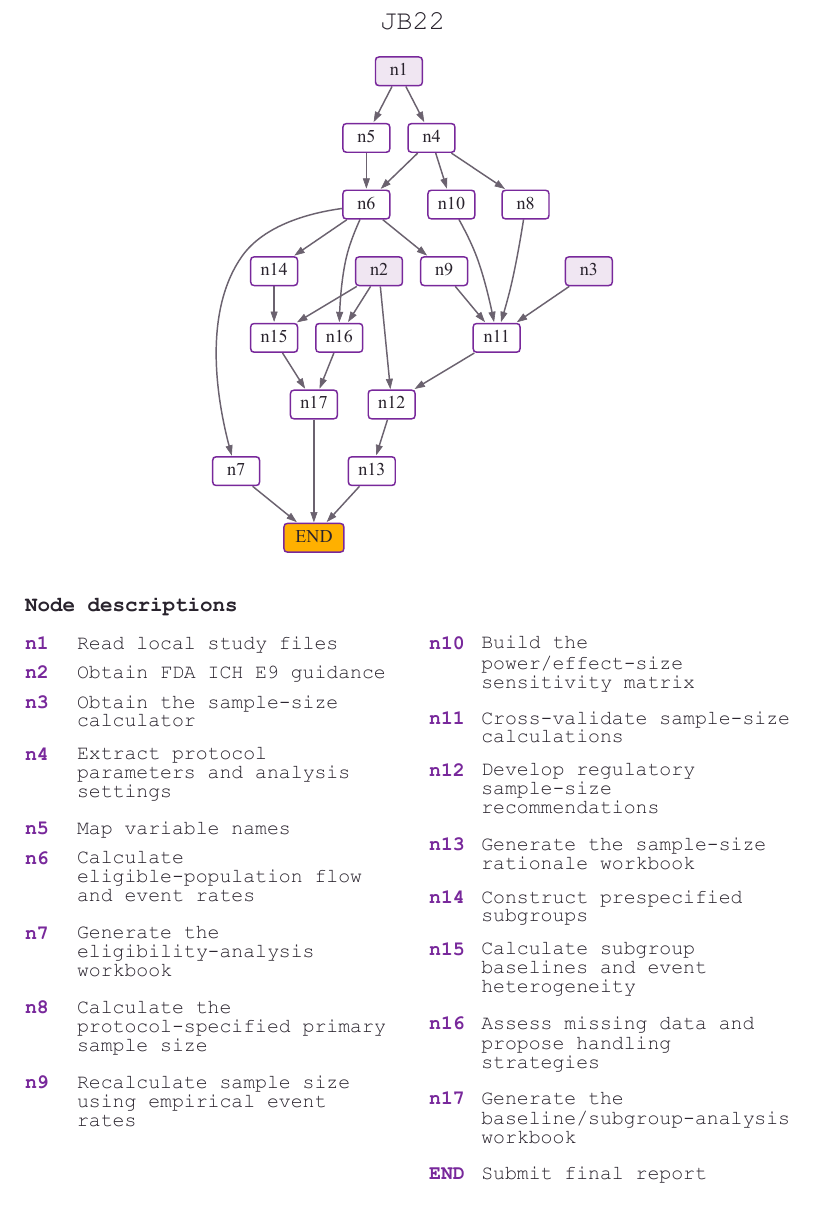}
\caption{DAG for \texttt{JB22}.}
\label{fig:dag-JB22}

\end{figure}

\clearpage

\begin{figure}[!ht]\centering

\includegraphics[width=\linewidth,height=.90\textheight,keepaspectratio]{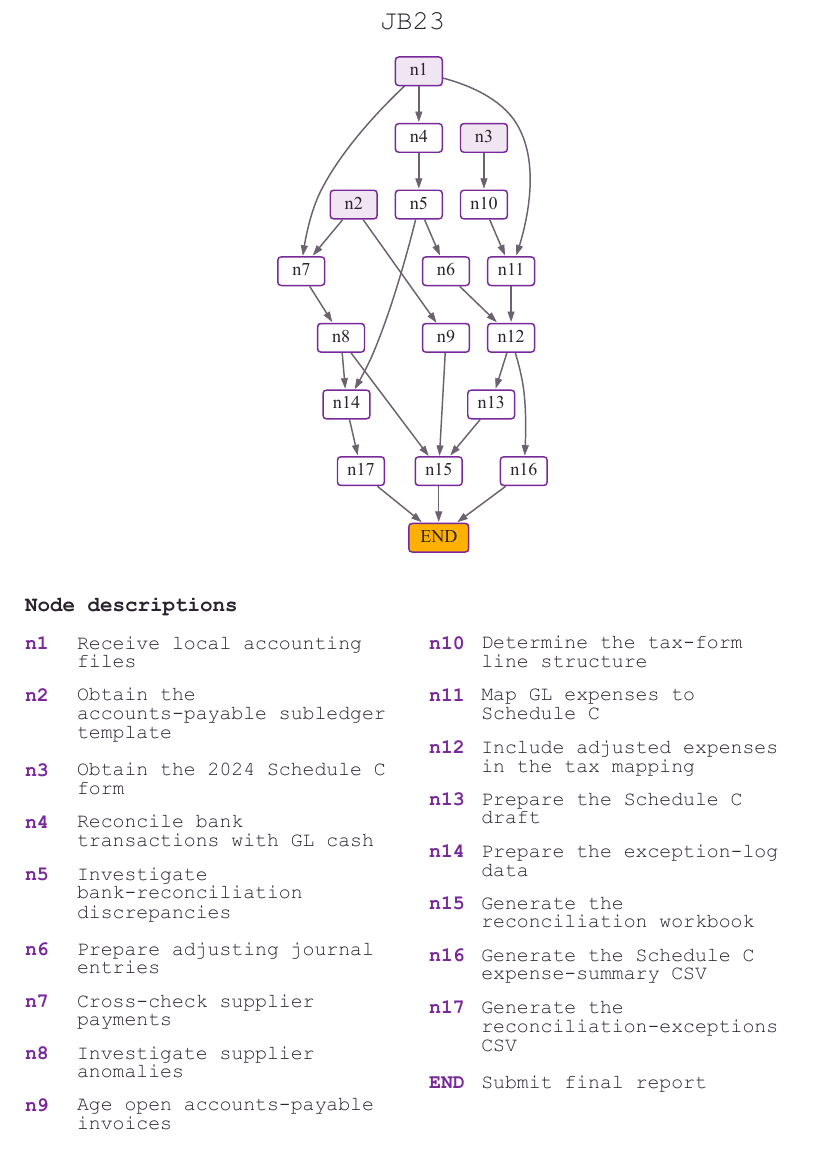}
\caption{DAG for \texttt{JB23}.}
\label{fig:dag-JB23}

\end{figure}

\clearpage

\begin{figure}[!ht]\centering

\includegraphics[width=\linewidth,height=.90\textheight,keepaspectratio]{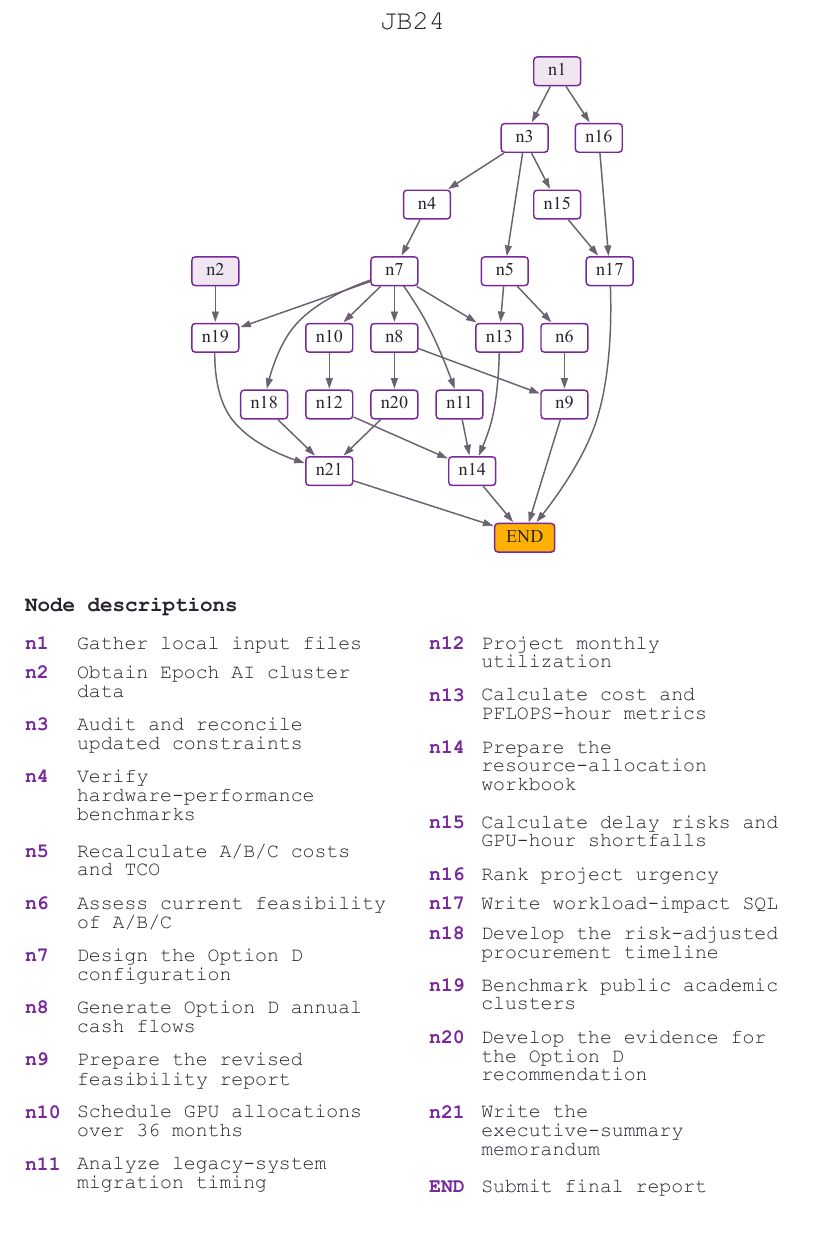}
\caption{DAG for \texttt{JB24}.}
\label{fig:dag-JB24}

\end{figure}

\clearpage

\begin{figure}[!ht]\centering

\includegraphics[width=\linewidth,height=.90\textheight,keepaspectratio]{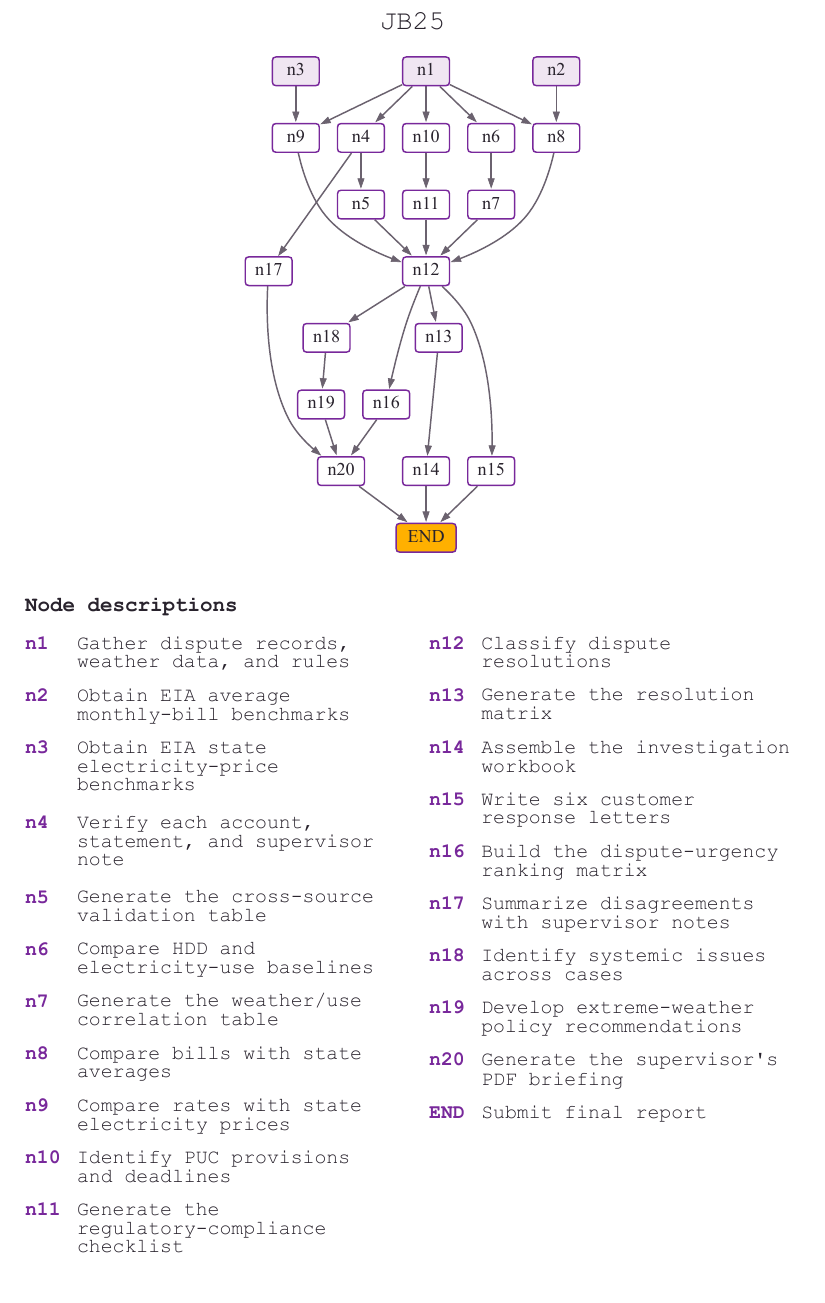}
\caption{DAG for \texttt{JB25}.}
\label{fig:dag-JB25}

\end{figure}

\clearpage

\begin{figure}[!ht]\centering

\includegraphics[width=\linewidth,height=.90\textheight,keepaspectratio]{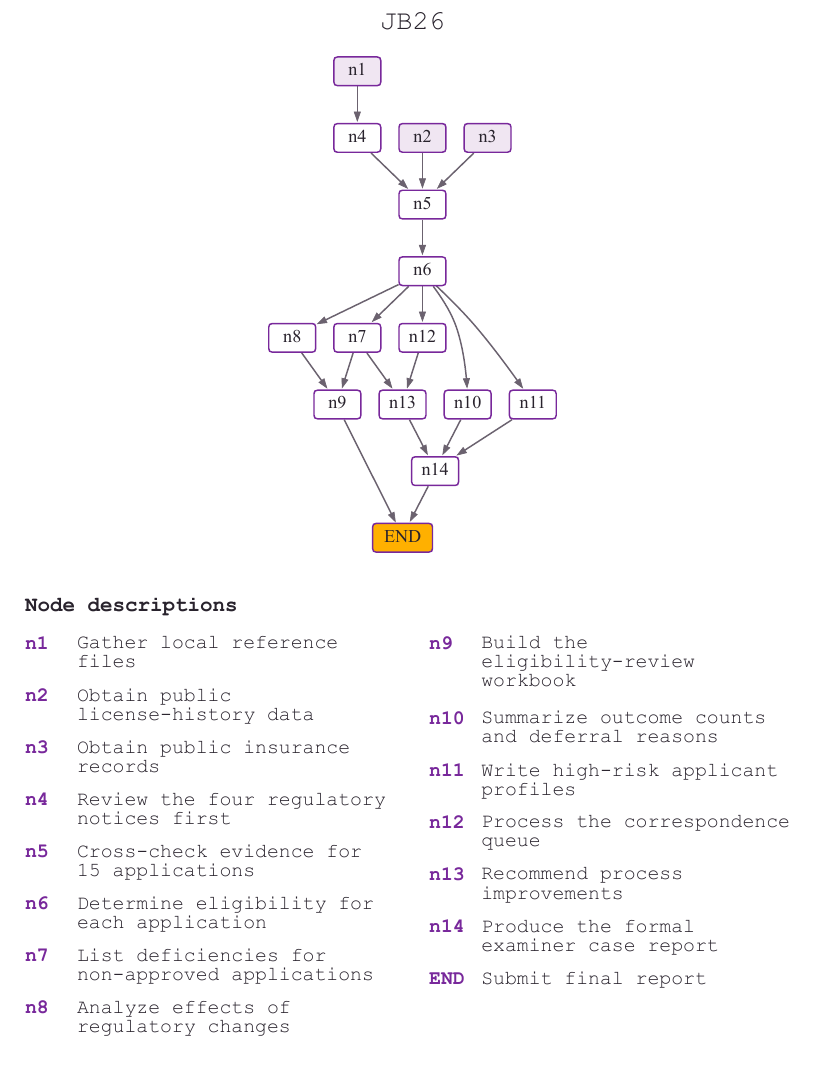}
\caption{DAG for \texttt{JB26}.}
\label{fig:dag-JB26}

\end{figure}

\clearpage

\begin{figure}[!ht]\centering

\includegraphics[width=\linewidth,height=.90\textheight,keepaspectratio]{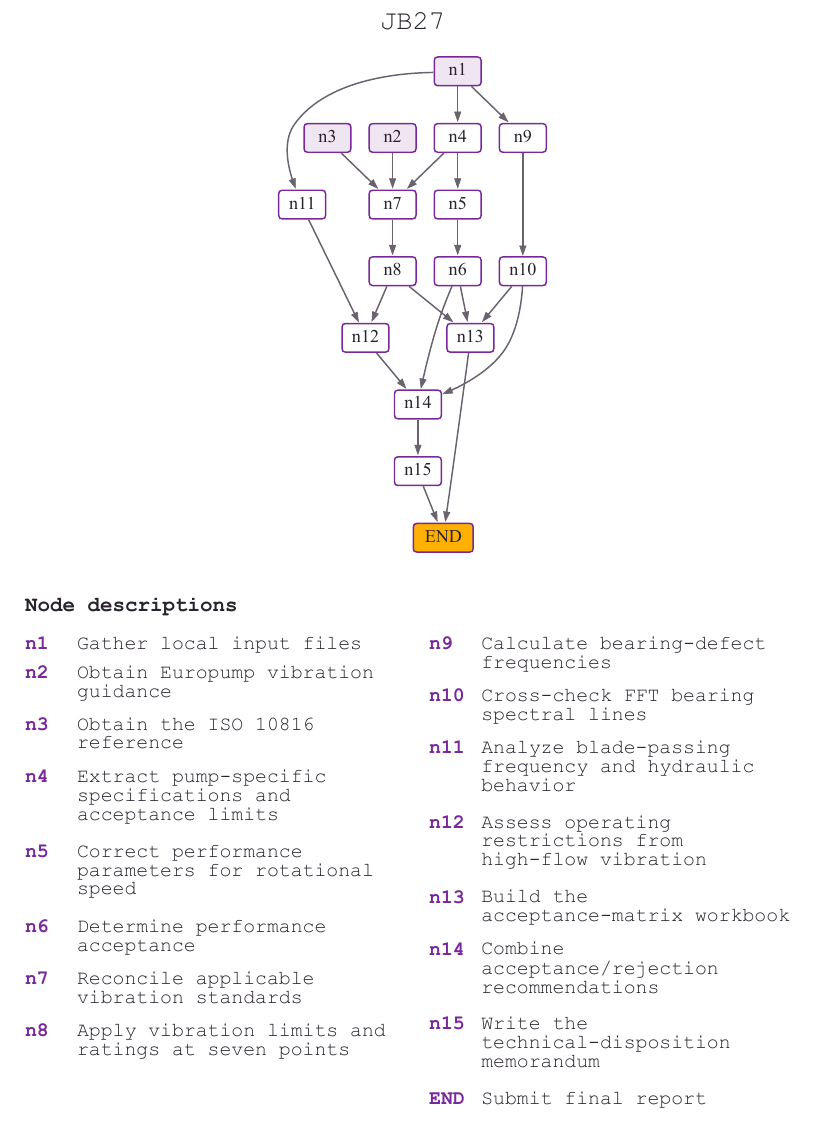}
\caption{DAG for \texttt{JB27}.}
\label{fig:dag-JB27}

\end{figure}

\clearpage

\begin{figure}[!ht]\centering

\includegraphics[width=\linewidth,height=.90\textheight,keepaspectratio]{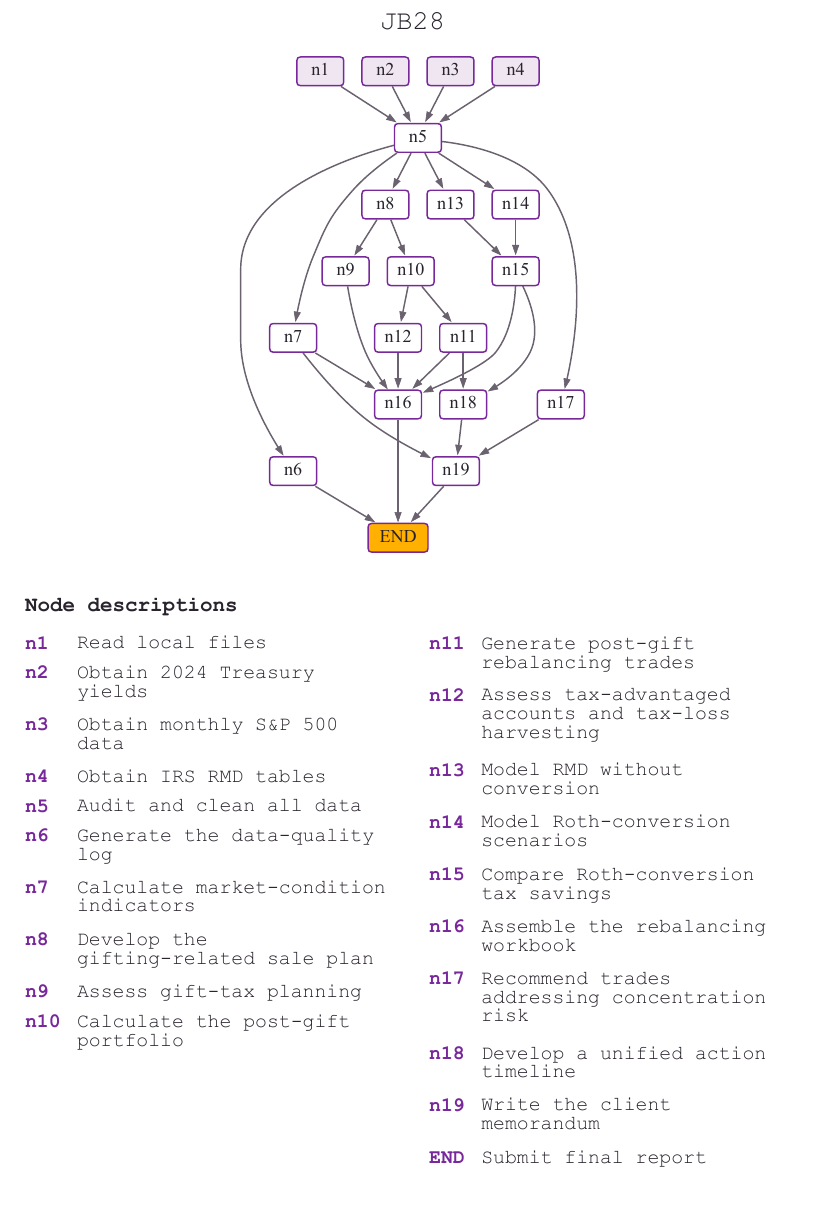}
\caption{DAG for \texttt{JB28}.}
\label{fig:dag-JB28}

\end{figure}

\clearpage

\begin{figure}[!ht]\centering

\includegraphics[width=\linewidth,height=.90\textheight,keepaspectratio]{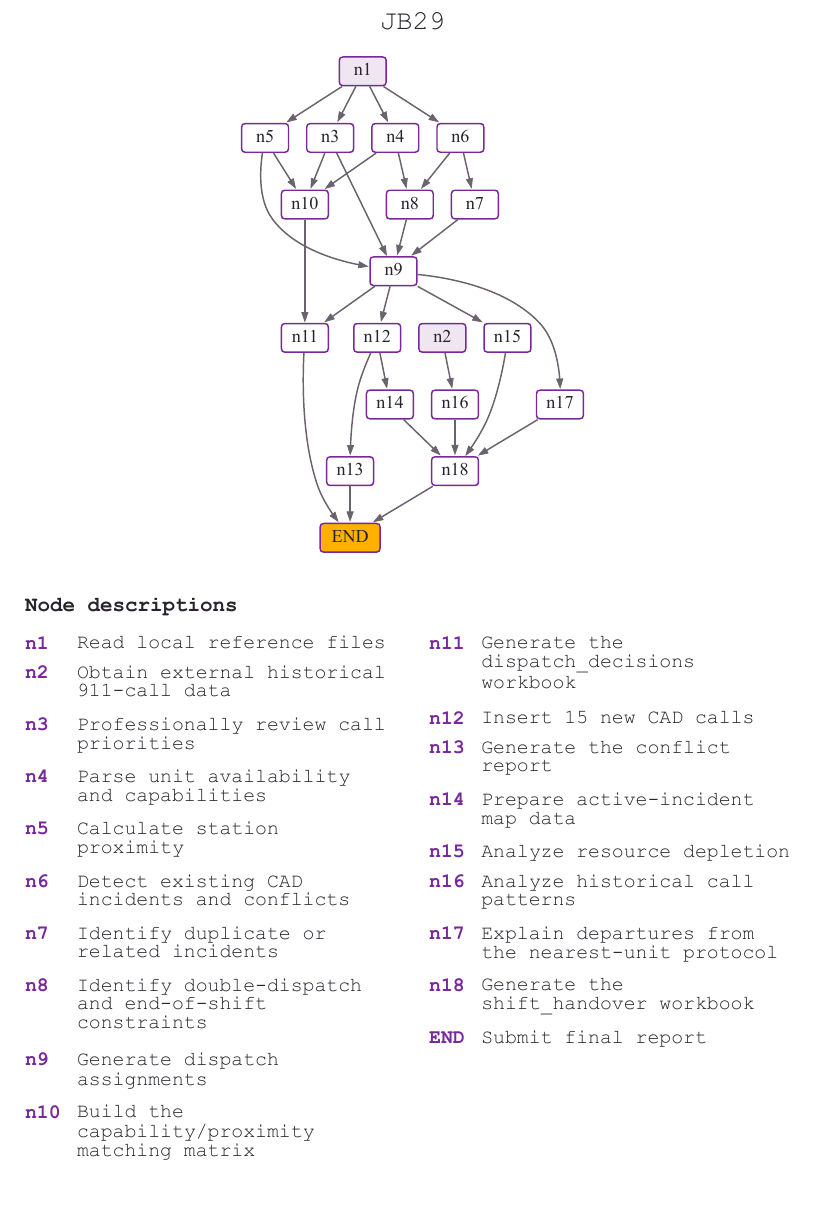}
\caption{DAG for \texttt{JB29}.}
\label{fig:dag-JB29}

\end{figure}

\clearpage

\begin{figure}[!ht]\centering

\includegraphics[width=\linewidth,height=.90\textheight,keepaspectratio]{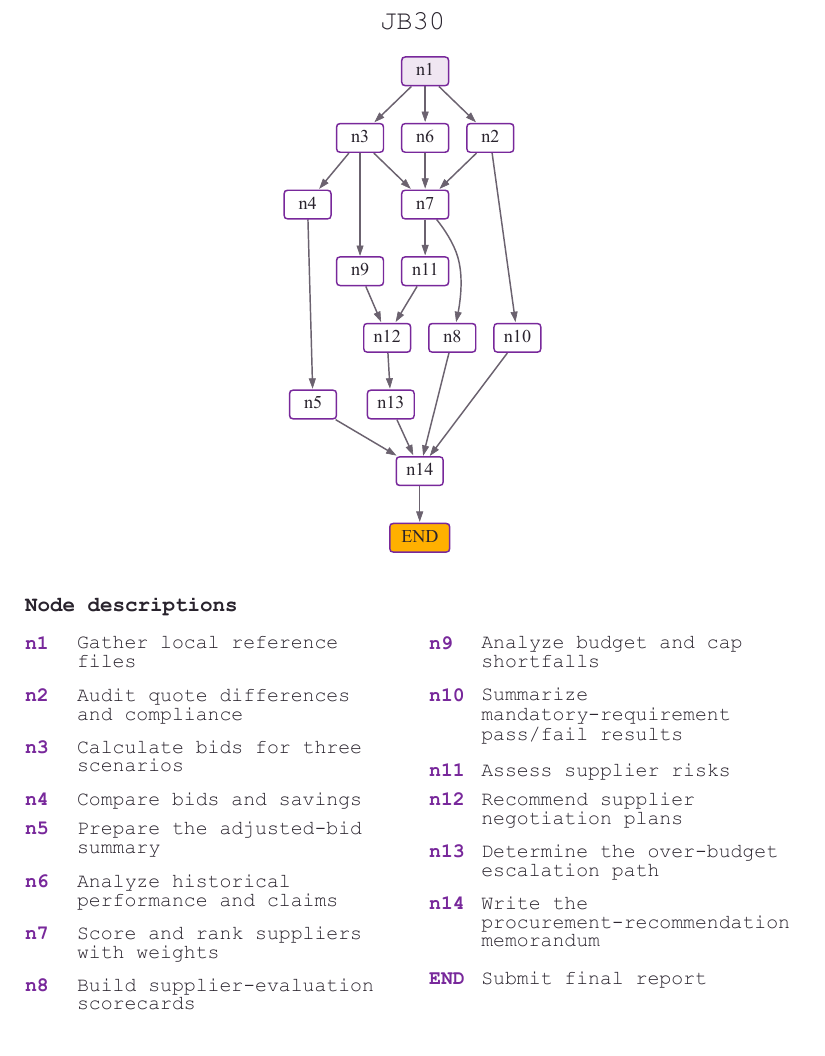}
\caption{DAG for \texttt{JB30}.}
\label{fig:dag-JB30}

\end{figure}


\begin{thebibliography}{25}
\providecommand{\natexlab}[1]{#1}
\providecommand{\url}[1]{\texttt{#1}}
\expandafter\ifx\csname urlstyle\endcsname\relax
  \providecommand{\doi}[1]{doi: #1}\else
  \providecommand{\doi}{doi: \begingroup \urlstyle{rm}\Url}\fi

\bibitem[Abhyankar et~al.(2026)Abhyankar, Qi, and
  Zhang]{abhyankar2025osworldhuman}
Reyna Abhyankar, Qi~Qi, and Yiying Zhang.
\newblock {OSWorld-Human}: Benchmarking the efficiency of computer-use agents.
\newblock \emph{Proceedings of Machine Learning and Systems}, 8:\penalty0
  482--494, 2026.

\bibitem[{Anthropic}(2026)]{anthropic2026opus55}
{Anthropic}.
\newblock {Claude Opus 5.5 System Card}.
\newblock System card, Anthropic, September 2026.
\newblock URL \url{https://www.anthropic.com/claude-opus-5-5-system-card}.
\newblock Published September 22, 2026.

\bibitem[Barres et~al.(2025)Barres, Dong, Ray, Si, and Narasimhan]{tau-bench-2}
Victor Barres, Honghua Dong, Soham Ray, Xujie Si, and Karthik Narasimhan.
\newblock {$\tau^2$-Bench}: Evaluating conversational agents in a dual-control
  environment.
\newblock \emph{arXiv preprint arXiv:2506.07982}, 2025.

\bibitem[Chen et~al.(2026)Chen, Trivedi, Pan, Zhang, Srinivasan,
  Balasubramanian, and Sabharwal]{chen2026appworldul}
Junzhi Chen, Harsh Trivedi, Jane Pan, Michael~JQ Zhang, Tejas Srinivasan,
  Niranjan Balasubramanian, and Ashish Sabharwal.
\newblock {AppWorld-UL}: Benchmarking diverse agent-user interactions for
  tool-use.
\newblock \emph{arXiv preprint arXiv:2607.20536}, 2026.

\bibitem[Dou et~al.(2026)Dou, Zhang, Yin, Huang, Shen, Wang, Chen, Ni, Ye,
  Zhang, et~al.]{dou2026clbench}
Shihan Dou, Ming Zhang, Zhangyue Yin, Chenhao Huang, Yujiong Shen, Junzhe Wang,
  Jiayi Chen, Yuchen Ni, Junjie Ye, Cheng Zhang, et~al.
\newblock {CL-bench}: A benchmark for context learning.
\newblock \emph{arXiv preprint arXiv:2602.03587}, 2026.

\bibitem[Hsieh et~al.(2024)Hsieh, Sun, Kriman, Acharya, Rekesh, Jia, Zhang, and
  Ginsburg]{hsieh2024ruler}
Cheng-Ping Hsieh, Simeng Sun, Samuel Kriman, Shantanu Acharya, Dima Rekesh, Fei
  Jia, Yang Zhang, and Boris Ginsburg.
\newblock {RULER}: What's the real context size of your long-context language
  models?
\newblock \emph{arXiv preprint arXiv:2404.06654}, 2024.

\bibitem[Huang et~al.(2024)Huang, Qing, Shang, Cui, and
  Zhang]{huang2024effibench}
Dong Huang, Yuhao Qing, Weiyi Shang, Heming Cui, and Jie Zhang.
\newblock {EffiBench}: Benchmarking the efficiency of automatically generated
  code.
\newblock \emph{Advances in Neural Information Processing Systems},
  37:\penalty0 11506--11544, 2024.

\bibitem[Jiang et~al.(2024)Jiang, Wang, Zeng, Zhong, Li, Mi, Shang, Jiang, Liu,
  and Wang]{jiang2024followbench}
Yuxin Jiang, Yufei Wang, Xingshan Zeng, Wanjun Zhong, Liangyou Li, Fei Mi,
  Lifeng Shang, Xin Jiang, Qun Liu, and Wei Wang.
\newblock {FollowBench}: A multi-level fine-grained constraints following
  benchmark for large language models.
\newblock In \emph{Proceedings of the 62nd Annual Meeting of the Association
  for Computational Linguistics (Volume 1: Long Papers)}, pp.\  4667--4688,
  2024.

\bibitem[Jimenez et~al.(2024)Jimenez, Yang, Wettig, Yao, Pei, Press, and
  Narasimhan]{swebench}
Carlos~E Jimenez, John Yang, Alexander Wettig, Shunyu Yao, Kexin Pei, Ofir
  Press, and Karthik Narasimhan.
\newblock {SWE-bench}: Can language models resolve real-world {GitHub} issues?
\newblock In \emph{International Conference on Learning Representations},
  volume 2024, pp.\  54107--54157, 2024.

\bibitem[Kapoor et~al.(2024)Kapoor, Stroebl, Siegel, Nadgir, and
  Narayanan]{kapoor2024agents}
Sayash Kapoor, Benedikt Stroebl, Zachary~S Siegel, Nitya Nadgir, and Arvind
  Narayanan.
\newblock {AI} agents that matter.
\newblock \emph{arXiv preprint arXiv:2407.01502}, 2024.

\bibitem[Li et~al.(2026)Li, Feng, Xu, Ma, Zheng, Jiang, Sun, Shao, Chen, Huang,
  et~al.]{jobbench}
Yuetai Li, Yichen Feng, Zhangchen Xu, Zixian Ma, Kaiyuan Zheng, Fengqing Jiang,
  Xinghua Sun, Rulin Shao, Zichen Chen, Yue Huang, et~al.
\newblock {JobBench}: Aligning agent work with human will.
\newblock \emph{arXiv preprint arXiv:2605.26329}, 2026.
\newblock URL \url{https://arxiv.org/abs/2605.26329}.

\bibitem[Lu et~al.(2025)Lu, Holleis, Zhang, Aumayer, Nan, Bai, Ma, Ma, Li, Yin,
  et~al.]{toolsandbox}
Jiarui Lu, Thomas Holleis, Yizhe Zhang, Bernhard Aumayer, Feng Nan, Haoping
  Bai, Shuang Ma, Shen Ma, Mengyu Li, Guoli Yin, et~al.
\newblock {ToolSandbox}: A stateful, conversational, interactive evaluation
  benchmark for {LLM} tool use capabilities.
\newblock In \emph{Findings of the Association for Computational Linguistics:
  NAACL 2025}, pp.\  1160--1183, 2025.

\bibitem[Ma et~al.(2024)Ma, Zhang, Zhu, Yang, Yang, Jin, Lan, Kong, and
  He]{ma2024agentboard}
Chang Ma, Junlei Zhang, Zhihao Zhu, Cheng Yang, Yujiu Yang, Yaohui Jin,
  Zhenzhong Lan, Lingpeng Kong, and Junxian He.
\newblock {AgentBoard}: An analytical evaluation board of multi-turn {LLM}
  agents.
\newblock \emph{Advances in Neural Information Processing Systems},
  37:\penalty0 74325--74362, 2024.

\bibitem[{OpenAI}(2026)]{openai2026astra}
{OpenAI}.
\newblock {GPT-6 Astra System Card}.
\newblock System card, OpenAI, September 2026.
\newblock URL \url{https://deploymentsafety.openai.com/gpt-6-astra}.
\newblock Published September 3, 2026. Updated September 22, 2026.

\bibitem[Patil et~al.(2025)Patil, Mao, Yan, Ji, Suresh, Stoica, and
  Gonzalez]{patil2025the}
Shishir~G Patil, Huanzhi Mao, Fanjia Yan, Charlie Cheng-Jie Ji, Vishnu Suresh,
  Ion Stoica, and Joseph~E Gonzalez.
\newblock The {Berkeley Function Calling Leaderboard (BFCL)}: From tool use to
  agentic evaluation of large language models.
\newblock In \emph{Forty-second International Conference on Machine Learning},
  2025.

\bibitem[Qian et~al.(2025)Qian, Liu, Prabhakar, Liu, Zhang, Chen, Ji, Yao,
  Heinecke, Savarese, et~al.]{userbench}
Cheng Qian, Zuxin Liu, Akshara Prabhakar, Zhiwei Liu, Jianguo Zhang, Haolin
  Chen, Heng Ji, Weiran Yao, Shelby Heinecke, Silvio Savarese, et~al.
\newblock {UserBench}: An interactive gym environment for user-centric agents.
\newblock \emph{arXiv preprint arXiv:2507.22034}, 2025.

\bibitem[Seshadri et~al.(2026)Seshadri, Cahyawijaya, Odumakinde, Singh, and
  Goldfarb-Tarrant]{lost-in-simulation}
Preethi Seshadri, Samuel Cahyawijaya, Ayomide Odumakinde, Sameer Singh, and
  Seraphina Goldfarb-Tarrant.
\newblock Lost in simulation: {LLM}-simulated users are unreliable proxies for
  human users in agentic evaluations.
\newblock In \emph{Proceedings of the 64th Annual Meeting of the Association
  for Computational Linguistics (Volume 1: Long Papers)}, pp.\  47423--47439,
  2026.

\bibitem[Styles et~al.(2024)Styles, Miller, Cerda-Mardini, Guha, Sanchez, and
  Vidgen]{workbench}
Olly Styles, Sam Miller, Patricio Cerda-Mardini, Tanaya Guha, Victor Sanchez,
  and Bertie Vidgen.
\newblock {WorkBench}: A benchmark dataset for agents in a realistic workplace
  setting.
\newblock \emph{arXiv preprint arXiv:2405.00823}, 2024.
\newblock URL \url{https://arxiv.org/abs/2405.00823}.

\bibitem[Subramonyam et~al.(2024)Subramonyam, Pea, Pondoc, Agrawala, and
  Seifert]{subramonyam2024}
Hari Subramonyam, Roy Pea, Christopher Pondoc, Maneesh Agrawala, and Colleen
  Seifert.
\newblock Bridging the gulf of envisioning: Cognitive challenges in prompt
  based interactions with {LLM}s.
\newblock In \emph{Proceedings of the 2024 CHI Conference on Human Factors in
  Computing Systems}, pp.\  1--19, 2024.

\bibitem[Trivedi et~al.(2024)Trivedi, Khot, Hartmann, Manku, Dong, Li, Gupta,
  Sabharwal, and Balasubramanian]{trivedi2024appworld}
Harsh Trivedi, Tushar Khot, Mareike Hartmann, Ruskin Manku, Vinty Dong, Edward
  Li, Shashank Gupta, Ashish Sabharwal, and Niranjan Balasubramanian.
\newblock {AppWorld}: A controllable world of apps and people for benchmarking
  interactive coding agents.
\newblock In \emph{Proceedings of the 62nd Annual Meeting of the Association
  for Computational Linguistics (Volume 1: Long Papers)}, pp.\  16022--16076,
  2024.

\bibitem[Yao et~al.(2022)Yao, Zhao, Yu, Du, Shafran, Narasimhan, and
  Cao]{yao2023react}
Shunyu Yao, Jeffrey Zhao, Dian Yu, Nan Du, Izhak Shafran, Karthik Narasimhan,
  and Yuan Cao.
\newblock {ReAct}: Synergizing reasoning and acting in language models.
\newblock \emph{arXiv preprint arXiv:2210.03629}, 2022.

\bibitem[Yao et~al.(2024)Yao, Shinn, Razavi, and Narasimhan]{yao2024tau}
Shunyu Yao, Noah Shinn, Pedram Razavi, and Karthik Narasimhan.
\newblock {$\tau$-bench}: A benchmark for tool-agent-user interaction in
  real-world domains.
\newblock \emph{arXiv preprint arXiv:2406.12045}, 2024.

\bibitem[Yen et~al.(2025)Yen, Gao, Hou, Ding, Fleischer, Izsak, Wasserblat, and
  Chen]{yen2024helmet}
Howard Yen, Tianyu Gao, Minmin Hou, Ke~Ding, Daniel Fleischer, Peter Izsak,
  Moshe Wasserblat, and Danqi Chen.
\newblock {HELMET}: How to evaluate long-context language models effectively
  and thoroughly.
\newblock \emph{arXiv preprint arXiv:2410.02694}, 2025.

\bibitem[Zhou et~al.(2023)Zhou, Lu, Mishra, Brahma, Basu, Luan, Zhou, and
  Hou]{zhou2023instruction}
Jeffrey Zhou, Tianjian Lu, Swaroop Mishra, Siddhartha Brahma, Sujoy Basu,
  Yi~Luan, Denny Zhou, and Le~Hou.
\newblock Instruction-following evaluation for large language models.
\newblock \emph{arXiv preprint arXiv:2311.07911}, 2023.

\bibitem[Zhou et~al.(2024)Zhou, Xu, Zhu, Zhou, Lo, Sridhar, Cheng, Ou, Bisk,
  Fried, et~al.]{webarena}
Shuyan Zhou, Frank~F Xu, Hao Zhu, Xuhui Zhou, Robert Lo, Abishek Sridhar,
  Xianyi Cheng, Tianyue Ou, Yonatan Bisk, Daniel Fried, et~al.
\newblock {WebArena}: A realistic web environment for building autonomous
  agents.
\newblock In \emph{International Conference on Learning Representations},
  volume 2024, pp.\  15585--15606, 2024.

\end{thebibliography}
\end{document}